\documentclass{article}

    \usepackage[final, nonatbib]{neurips_2022}

\usepackage[utf8]{inputenc} % allow utf-8 input
\usepackage[T1]{fontenc}    % use 8-bit T1 fonts
\usepackage{hyperref}       % hyperlinks
\usepackage{url}            % simple URL typesetting
\usepackage{booktabs}       % professional-quality tables
\usepackage{amsfonts}       % blackboard math symbols
\usepackage{nicefrac}       % compact symbols for 1/2, etc.
\usepackage{microtype}      % microtypography
\usepackage{amsmath}
\usepackage{graphicx}
\usepackage{colortbl}
\usepackage{xcolor}
\usepackage{wrapfig}
\usepackage{subcaption}
\usepackage{multirow}
\usepackage{makecell}
\usepackage{mathrsfs}
\usepackage{algorithm}
\usepackage{algpseudocode}

\usepackage{pifont}% http://ctan.org/pkg/pifont
\newcommand{\cmark}{\ding{51}}%
\newcommand{\xmark}{\ding{55}}%

\title{\raggedright Attention-Based Synthetic Data Generation\\
\raggedright for Calibration-Enhanced Survival Analysis:\\
\raggedright A Case Study for Chronic Kidney Disease Using\\
\raggedright Electronic Health Records}

\author{Nicholas I-Hsien Kuo, Blanca Gallego, Louisa Jorm\\
Centre for Big Data Research in Health (CBDRH)\\
The University of New South Wales, Sydney, Australia\\
\footnotesize{\textcolor{white}{*}}\\
Corresponding author: Nicholas I-Hsien Kuo (\texttt{n.kuo@unsw.edu.au})
}

\begin{document}

\maketitle

%###===>>>%###===>>>%###===>>>
%###===>>>%###===>>>%###===>>>
%###===>>>%###===>>>%###===>>>
\begin{abstract}
Access to real-world healthcare data is limited by stringent privacy regulations and data imbalances, hindering advancements in research and clinical applications. Synthetic data presents a promising solution, yet existing methods often fail to ensure the realism, utility, and calibration essential for robust survival analysis. Here, we introduce \textbf{Masked Clinical Modelling (MCM)}, an attention-based framework capable of generating high-fidelity synthetic datasets that preserve critical clinical insights, such as hazard ratios, while enhancing survival model calibration. Unlike traditional statistical methods like SMOTE and machine learning models such as VAEs, MCM supports both standalone dataset synthesis for reproducibility and conditional simulation for targeted augmentation, addressing diverse research needs. Validated on a chronic kidney disease electronic health records dataset, MCM reduced the general calibration loss over the entire dataset by 15\%; and MCM reduced a mean calibration loss by 9\% across 10 clinically stratified subgroups, outperforming 15 alternative methods. By bridging data accessibility with translational utility, MCM advances the precision of healthcare models, promoting more efficient use of scarce healthcare resources\footnote{\label{Ref:OpenSource}All codes will be made publicly available after paper acceptance.}.
\end{abstract}

%%%===%%%%%%===%%%%%%===%%%
%%%===%%%%%%===%%%%%%===%%%
%%%===%%%%%%===%%%%%%===%%%
\section{Introduction}
Access to real-world clinical data is vital for advancing healthcare research and improving patient outcomes. However, stringent privacy regulations (\textit{e.g.,} HIPAA in the USA~\cite{nosowsky2006health} and the Privacy Act 1988 in Australia~\cite{okeefe2010privacy}) as well as ethical considerations often restrict data availability, creating barriers to collaboration, reproducibility, and innovation. Synthetic data, which replicates the statistical properties of real-world datasets while removing identifiable information, offers a transformative solution to these challenges~\cite{kuo2022health}. By enabling secure data sharing and research collaboration, synthetic data holds promise for applications ranging from educational training~\cite{nicholas2024enriching} to the development of robust machine learning (ML) models~\cite{ingabire2024canonical, kowsar2024attention}. Yet, despite its potential, synthetic data is often critiqued for its lack of practicality when realism or utility is compromised for the sake of privacy~\cite{goncalves2020generation, appenzeller2022privacy, hermsen2023privacy, sarmin2024synthetic}.

In this study, we argue that synthetic data must demonstrate practicality -- its capacity to enhance model performance and yield translational insights in real-world healthcare scenarios. Survival analysis~\cite{clark2003survival}, a statistical approach focused on modelling time-to-event outcomes for disease progression and patient mortality, is particularly dependent on practicality, as it underpins tasks like policy-making~\cite{kim2024determinants}, clinical trials~\cite{emmerson2021understanding}, and personalised treatment planning~\cite{smith2016personalized}. A key metric in survival analysis is the hazard ratio (HR), which quantifies the relative effect of covariates, such as age or comorbidities, on the likelihood of an event occurring at a given time. To be practical, synthetic data must also ensure calibration -- the agreement between predicted probabilities and observed outcomes across risk groups~\cite{alba2017discrimination, van2019calibration} -- which is critical for producing reliable and actionable clinical insights, such as identifying individuals who are at high risk for adverse outcomes and may benefit from early intervention, and estimating the probability of resource-intensive care (\textit{e.g.,} ICU admissions).

Synthetic data for survival analysis must capture the intricate dependencies between temporal dynamics, covariates, and outcomes, a challenge unmet by many existing approaches. Methods like variational autoencoders (VAEs)~\cite{kingma2014auto}, generative adversarial networks (GANs)~\cite{goodfellow2014generative}, and traditional statistical techniques (\textit{e.g.,} SMOTE~\cite{chawla2002smote}) often fall short in preserving the nuanced relationships critical for survival models~\cite{kuo2024ck4gen}, particularly for Cox Proportional Hazards (CoxPH) models~\cite{cox1972regression}. Furthermore, few studies focus on synthetic data applications in survival analysis, and existing work rarely addresses practical considerations in healthcare, such as ensuring stratified calibration for high-risk subgroups.

To address these challenges, we propose \textbf{Masked Clinical Modelling} (MCM), an attention-based machine learning framework inspired by masked language modelling~\cite{bahdanau2014neural, vaswani2017attention, devlin2018bert}. MCM generates synthetic data that balances realism, utility, and practicality by masking portions of clinical data and reconstructing them using contextual feature dependencies. This dual capability -- standalone synthesis and conditional augmentation -- enables MCM to preserve key metrics like hazard ratios, as well as addressing data imbalance to enhance calibration in downstream survival models.

Our experiments evaluate MCM on a real-world chronic kidney disease (CKD) electronic health records (EHR) dataset~\cite{al2018chronic}, demonstrating its ability to augment existing datasets and enhance downstream CoxPH model performance in identifying the onset of CKD stages 3–5~\cite{hill2016global}. To validate these claims, we benchmark MCM against 17 competitive baseline methods, including deep learning generative models and traditional statistical approaches for rebalancing imbalanced clinical features. The results showcase MCM’s favourable performance in generating clinically meaningful synthetic data that improves model robustness and calibration.

The CKD EHR dataset utilised in this study is openly available and de-identified. This work prioritises the \textbf{realism}, \textbf{utility}, and \textbf{practicality} of the generated data rather than privacy considerations, given the dataset's pre-existing anonymisation. Importantly, synthetic data retains considerable value even within highly secure environments, such as those regulated by the Australian Institute of Health and Welfare’s (AIHW) \textit{Secure On-Site Access} principle~\cite{aihw_five_safes}, established under the Data Availability and Transparency Act 2020~\cite{australia_data_availability_bill}. These frameworks impose stringent safeguards, including supervised data access, mandatory national background checks, and physical presence requirements, ensuring the security of the raw datasets. In such \textit{walled-garden} contexts~\cite{starren2023privacy}, synthetic data serves as a critical tool for augmenting training datasets and enhancing the calibration of national risk score calculators (\textit{e.g.,} for cardiovascular disease prognosis~\cite{nelson20242023}). By enabling improved representation across diverse populations, synthetic data supports the equitable and robust development of predictive healthcare models without compromising the security or integrity of the original data.

%%%===%%%%%%===%%%%%%===%%%
%%%===%%%%%%===%%%%%%===%%%
%%%===%%%%%%===%%%%%%===%%%
%\newpage
\section{Preliminaries}
\subsection{Chronic Kidney Disease}\label{Sec:CKD}
CKD is a global health concern, often progressing silently until advanced stages, requiring timely identification and intervention~\cite{hill2016global}. The dataset used in this study, introduced by Al-Shamsi \textit{et al.}~\cite{al2018chronic}, comprises longitudinal EHR from 491 adult UAE nationals treated at Tawam Hospital between 2008 and 2017. It includes key clinical measurements such as estimated glomerular filtration rate (eGFR), collected at three-month intervals, enabling tracking of CKD progression to stages 3 - 5. Demographic, clinical, and medication data, alongside lifestyle factors like smoking and obesity, provide a comprehensive resource for studying CKD trajectories (see Table \ref{Tab:CKDTab1}). The dataset is publicly available under the Creative Commons Attribution 4.0 International (CC BY 4.0) license~\cite{chicco2021chronic}.

This dataset replicates walled-garden research environments~\cite{starren2023privacy}, where privacy concerns are mitigated through prior de-identification. A prior study~\cite{chicco2021machine} demonstrated that binary classification of event outcomes on this dataset achieved an accuracy of 0.843 using random forest techniques, yet

%###---###%###---###%###---###%###---###%###---###
%###---###%###---###%###---###%###---###%###---###
%###---###%###---###%###---###%###---###%###---###
\newpage
\begin{table}[ht]
\centering
\caption{\label{Tab:CKDTab1}Descriptive Statistics}
\begin{tabular}{p{7.5cm}p{1.25cm}p{3.75cm}}
\hline
\textbf{Category} & 
\textbf{Attribute} & 
\textbf{Median [IQR] \& \% share} \\
\hline
\hline
%%%===%%%%%%===%%%%%%===%%%
\multicolumn{3}{l}{\textbf{Demographic and Lifestyle Variables}} \\

\hspace{10pt} Age (years) & 
Numeric& 
54.00 [44.00 - 64.00] \\

\hspace{10pt} Sex (\% share of females) & 
Binary & 
50.92\% \\

\hspace{10pt} Smoking history (\%) & 
Binary & 
15.27\% \\

\hspace{10pt} Obesity history (\%) & 
Binary & 
50.51\% \\

%%%===%%%%%%===%%%%%%===%%%
\hline
\multicolumn{3}{l}{\textbf{Clinical Baseline Measurements}} \\
\hspace{10pt} Total cholesterol (mmol/L) & 
Numeric& 
5.00 [4.20 - 5.77] \\

\hspace{10pt} Serum creatinine ($\mu$mol/L) & 
Numeric& 
66.00 [55.00 - 78.50] \\

\hspace{10pt} eGFR (mL/min/1.73m\textsuperscript{2}) & 
Numeric& 
98.10 [86.40 - 109.50] \\

\hspace{10pt} Systolic blood pressure (mmHg) & 
Numeric& 
131.00 [121.00 - 141.00] \\

\hspace{10pt} Diastolic blood pressure (mmHg) & 
Numeric& 
77.00 [69.00 - 83.00] \\

\hspace{10pt} Body mass index (kg/m\textsuperscript{2}) & 
Numeric& 
30.00 [26.00 - 33.00] \\

%%%===%%%%%%===%%%%%%===%%%
\hline
\multicolumn{3}{l}{\textbf{Medical History}} \\
\hspace{10pt} History of diabetes (\%) & 
Binary & 
43.79\% \\

\hspace{10pt} History of coronary heart disease (\%) &
Binary & 
9.16\% \\

\hspace{10pt} History of vascular disease (\%) & 
Binary & 
5.91\% \\

\hspace{10pt} History of hypertension (\%) & 
Binary & 
68.23\% \\

\hspace{10pt} History of dyslipidaemia (\%) & 
Binary & 
64.56\% \\

%%%===%%%%%%===%%%%%%===%%%
\hline
\multicolumn{3}{l}{\textbf{Medication Use}} \\
\hspace{10pt} Lipid-lowering medication (\%) & 
Binary & 
55.19\% \\

\hspace{10pt} Diabetes medication (\%) & 
Binary & 
32.79\% \\

\hspace{10pt} Blood pressure-lowering medication (\%) &
Binary & 
61.71\% \\

\hspace{10pt} ACE inhibitors/angiotensin receptor blockers (\%) & 
Binary & 
44.60\% \\

%%%===%%%%%%===%%%%%%===%%%
\hline
\multicolumn{3}{l}{\textbf{Follow-up Summary}} \\
\hspace{10pt} Follow-up duration (years)\newline
\hspace*{15pt} Overall\newline
\hspace*{15pt} Experienced CKD stage 3-5\newline
\hspace*{15pt} Not experienced CKD stage 3-5& 
Numeric& 
\textcolor{white}{.}\newline
8.00 [6.00 - 8.00]\newline
4.00 [2.00 - 7.00]\newline
8.00 [7.00 - 8.00]
\\
\hspace{10pt} Event rates (\%) & 
Binary & 
11.41\% \\
\hline
\end{tabular}
\end{table}

%###---###%###---###%###---###%###---###%###---###
%###---###%###---###%###---###%###---###%###---###
%###---###%###---###%###---###%###---###%###---###
yielded a modest F1-score of 0.550, highlighting challenges in balanced predictive performance.

While diagnostic tasks have been explored, the dataset remains underutilised\footnote{The CKD EHR dataset, originally collected by Al-Shamsi \textit{et al.}~\cite{al2018chronic}, was designed to develop sex-specific risk score equations for assessing risk factors. As of 2024-11-21, the study has been cited 54 times on Google Scholar, with most research focusing on binary classification tasks (\textit{e.g.,} diagnosis)~\cite{chicco2021machine, alloghani2020performance}. Additionally, this dataset has also been utilised in other forms of machine learning applications, such as employing explainable artificial intelligence (XAI) techniques to enhance interpretability in modelling decisions~\cite{nguycharoen2024explainable}, while its prognostic potential remains largely unexplored.} in prognostic applications, offering a unique opportunity to enhance downstream survival analysis models, such as CoxPH, with a focus on calibration -- ensuring precise identification of high-risk patients.

This dataset is well-suited for evaluating our MCM framework, which generates synthetic data for standalone use and augments existing datasets to address feature imbalances and enhance stratified calibration. By applying MCM, we aim to improve model reliability and equity in CKD prognosis, aligning with public health priorities for accurate and equitable predictions, particularly for underserved populations~\cite{van2022harm}.

\subsection{Cox Proportional Hazard Models and Model Calibration}\label{Sec:CoxPHCali}
\paragraph{Cox Proportional Hazard Model (CoxPH)}
CoxPH~\cite{cox1972regression} is a popular modelling approach in survival analysis for investigating the relationship between covariates and time-to-event outcomes. Let \( T \) denote the time-to-event duration, and \( \mathbf{X} = (X_1, X_2, \dots, X_d)^\top \) represent a vector of \( d \) covariates for an individual. The CoxPH model specifies the hazard function \( h(t | \mathbf{X}) \) as:
\begin{equation}\label{Eq:CoxPHHR}
h(t | \mathbf{X}) = h_0(t) \exp(\mathbf{X}^\top \boldsymbol{\beta}).
\end{equation}

%###---###%###---###%###---###%###---###%###---###
%###---###%###---###%###---###%###---###%###---###
%###---###%###---###%###---###%###---###%###---###
\newpage
\begin{algorithm}[h]
\caption{5x2 Cross-Validation for Individual Risks}
\label{alg:Risks}
\begin{algorithmic}[1]
\Require Clinical dataset \( \mathcal{D} \) with covariates \( \mathbf{X} \) and event information, and set of time points \( \mathcal{T} \)

\Procedure{5x2 Cross-Validation and LPH Computation}{}
    \For{each fold \( f = 1, \dots, 5 \)}
        \For{each split \( s = 0, 1 \)}
            \State \textbf{Split Dataset}: 
            Partition \( \mathcal{D} \) into training set \( \mathcal{D}_{\text{train}} \) and test set \( \mathcal{D}_{\text{test}} \) based on \( f \) and \( s \)
            \State \textbf{Fit CoxPH Model}: 
            \State \textbf{Compute LPH for Test Set}:\\
            \hspace*{25mm}For each patient \( i \in \mathcal{D}_{\text{test}} \), calculate log partial hazard (LPH):\\
            \hspace*{35mm}\(\text{LPH}_i = (\mathbf{X}_i - \boldsymbol{\mu})^\top \boldsymbol{\beta}
            \)\\
            \hspace*{25mm}with mean of the covariates \( \boldsymbol{\mu} \) in \( \mathcal{D}_{\text{train}} \)\\
            \hspace*{25mm}and the CoxPH coefficients \( \boldsymbol{\beta} \) (see Equation \eqref{Eq:CoxPHHR}).
        \EndFor
    \EndFor
    \State \textbf{Aggregate LPH}: 
    Compute the mean LPH for each patient across all folds for each \( t \in \mathcal{T} \).
\EndProcedure
\end{algorithmic}
\end{algorithm}

%%%###===###
\begin{algorithm}[h]
\caption{Calibration Slope Computation}
\label{alg:Calibration}
\begin{algorithmic}[1]
\Require Input dataset \( \mathcal{D}_\text{input} \)

\Procedure{Calibration Analysis}{}
    \State \textbf{Compute Metrics}:\\
    \hspace*{10mm}Apply Algrithm \ref{alg:Risks} to acquire \( \mathscr{R}_{\text{perc}}(\mathcal{D}_\text{input})\) and \( \mathscr{E}_{\text{perc}} (\mathcal{D}_\text{input})\)
    \State \textbf{Fit Calibration Slope}:\\
    \hspace*{10mm}Fit a linear regression model for
    \hspace*{3mm}\(
    \mathscr{R}_{\text{perc}}(\mathcal{D}_\text{input}) = \mathscr{S} \cdot \mathscr{E}_{\text{perc}}(\mathcal{D}_\text{input})
    \)
    \State \textbf{Quantify Calibration Error}: 
    Compute
    \(
    |1 - \mathscr{S}|
    \)
\EndProcedure
\end{algorithmic}
\end{algorithm}

%%%###===###
\begin{algorithm}[h]
\caption{Stratified Calibration}
\label{alg:Stratified}
\begin{algorithmic}[1]
\Require Stratification dataset \( \mathcal{S} \)
\Procedure{Stratified Calibration by Subgroup}{}
    \State Apply Algorithm \ref{alg:Calibration} with sub-dataset \( S \)
\EndProcedure
\end{algorithmic}
\end{algorithm}

\( h_0(t) \) is the baseline hazard function, representing the hazard when all covariates are zero; and \(\boldsymbol{\beta} = (\beta_1, \beta_2, \dots, \beta_p)^\top \) is a vector of coefficients estimating the effect of covariates on the hazard.

The model assumes proportional hazards, meaning that the ratio of hazard functions for two individuals is constant over time:
\begin{equation}
\frac{h(t | \mathbf{X}_i)}{h(t | \mathbf{X}_j)} = \exp\left( (\mathbf{X}_i - \mathbf{X}_j)^\top \boldsymbol{\beta} \right).
\end{equation}
The coefficients \( \beta_k \) are interpreted as log-hazard ratios. For a one-unit increase in \( X_k \), the hazard changes by a factor of \( \exp(\beta_k) \), holding other variables constant\footnote{Our work does not delve into specific techniques such as centring variable values, using splines and knots, or sex-specific modelling. While these approaches hold significant epidemiological value~\cite{coxph_centering_uw}, our focus is on introducing a ML-based method to enhance calibration in predictive tasks. The interpretation of individual variables is more suited to causal inference studies, which are best conducted with unitary models rather than multivariate approaches. As highlighted in \cite{mainzer2023handling}, conflating prediction and causal inference can lead to the Table 2 fallacy~\cite{westreich2013table}, resulting in misinterpretations and potentially misleading conclusions in healthcare contexts.}.

\paragraph{Model Calibration in Survival Analysis}
Calibration assesses the agreement between predicted risks and observed outcomes, ensuring predictions are actionable and reliable~\cite{van2019calibration}. However, challenges such as imbalanced data and rare events can undermine calibration accuracy, particularly in high-risk subgroups.

This study employs a robust quantile-based calibration framework, inspired by the New Zealand cardiovascular disease PREDICT risk score methodology~\cite{barbieri2022predicting, varianz2012}, to evaluate model reliability and interpretability in survival analysis. The calibration process integrates a systematic 5x2 cross-validation approach~\cite{dietterich1998approximate}, utilising the Lifelines library in Python~\cite{Davidson-Pilon2019} for CoxPH modelling. Predicted risk scores, represented as log partial hazard (LPH) values, are computed for each fold and split. Additionally, risks at predefined time points are calculated using baseline survival probabilities. These results are then aggregated across all folds and splits. The pseudocode summarising this process is presented in Algorithm~\ref{alg:Risks}, with full implementation details provided in Appendix~\ref{App:Cali}.

\paragraph{Slope in the Calibration Plot} 
The calibration framework evaluates the alignment between predicted risk (\( \mathscr{R}_{\text{perc}} \)) and observed risks (\( \mathscr{E}_{\text{perc}} \)). By assuming the two risks have a linear relationship:
\begin{align}\label{Eq:CalSlope}
\mathscr{R}_{\text{perc}} = \mathscr{S} \cdot \mathscr{E}_{\text{perc}},
\end{align}
we can use the calibration slope \( \mathscr{S} \) to represent the relationship between predicted and observed risks. A slope of \( \mathscr{S} = 1 \) indicates perfect calibration, while deviations from 1 quantify calibration error:
\begin{align}\label{Eq:CalLoss}
\text{Calibration Error} = |1 - \mathscr{S}|.
\end{align}
The calibration slope quantifies the agreement between predicted and observed risks, and a visualisation of this relationship is typically presented through calibration plots. The pseudocode for this process is outlined in Algorithm~\ref{alg:Calibration}.

To ensure model reliability across diverse subgroups, calibration is further stratified by clinically relevant variables such as diabetes status or eGFR levels; see the pseudocode in Algorithm~\ref{alg:Stratified}.

%%%===%%%%%%===%%%%%%===%%%
%%%===%%%%%%===%%%%%%===%%%
%%%===%%%%%%===%%%%%%===%%%
%\newpage
\section{Related Work}\label{Sec:RelatedWork}

\textbf{Advances in Synthetic Data Generation}\\
Synthetic data generation has long been a core component of ML development, progressing from autoencoders~\cite{hinton2011transforming} to sophisticated models like variational autoencoders (VAEs)~\cite{kingma2014auto}, generative adversarial networks (GANs)~\cite{goodfellow2014generative}, and denoising diffusion probabilistic models (DDPMs)~\cite{ho2020denoising}. These methods have achieved exceptional success in domains such as image generation~\cite{zhu2017unpaired}, natural language processing in text~\cite{reed2016generative}, and text-to-video generation~\cite{singer2022make}.

\textbf{Synthetic Data in Healthcare}\\
Stringent privacy regulations in healthcare often restrict data sharing, making generative ML a valuable alternative~\cite{kuo2022health, nicholas2023generating, kuo2023synthetic}. Global initiatives highlight the growing importance of synthetic data in this domain. The U.S. Department of Homeland Security, for instance, has partnered with Betterdata AI (Singapore), DataCebo (Massachusetts), MOSTLY AI (Austria), and Rockfish Data (California) to advance synthetic data technologies~\cite{dhs_synthetic_data_contracts}. Within Australia, the Synthetic Data Community of Practice (SynD), led by the Digital Health CRC, fosters interstate collaboration among universities in New South Wales, Queensland, and Victoria, alongside health ministries in Western Australia and the Northern Territory, to promote privacy-compliant, accessible synthetic data research~\cite{synthetic_data_practice_synd}. Similarly, South Australia's health ministry has collaborated with Gretel AI (California) to develop innovative synthetic data solutions~\cite{microsoft_sa_health, australian_fake_patients}. These efforts exemplify the global momentum toward leveraging synthetic data for secure and impactful research.

\textbf{Challenges in Survival Analysis \& Limitations of Existing Methods}\\
Despite its promise, generating synthetic data for survival analysis remains challenging due to the temporal and conditional nature of the data. Patients with similar baseline characteristics may experience vastly different event outcomes or durations, requiring generative models to capture these complex dependencies~\cite{nagpal2021deep}. Unlike tabular datasets, survival datasets demand nuanced modelling of covariates, outcomes, and their interactions over time. For instance, a model must accurately simulate patients progressing to CKD at different time horizons, such as one year versus eight years.

Well-known methods like SMOTE~\cite{chawla2002smote}, designed to address dataset imbalances by interpolating between data points, are ill-suited for survival analysis~\cite{kuo2024ck4gen}. By blending high-risk and low-risk groups, SMOTE risks generating synthetic patients with unrealistic risk profiles, which distort survival curves and hazard ratios, undermining their clinical utility. Similar limitations are observed with na\"ively implemented VAEs and GANs~\cite{kuo2024ck4gen}. VAEs force diverse patient profiles into a shared latent space, potentially obscuring distinctions between risk groups, while GANs generate data from random vectors without explicitly modelling the differences in latent distributions across risk categories. Both methods are prone to structural issues: VAEs suffer from posterior collapse~\cite{van2017neural}, and GANs from mode collapse~\cite{che2016mode}, resulting in reduced dataset diversity and completeness. These deficiencies can degrade the performance of downstream healthcare models, potentially leading to suboptimal clinical decisions and adverse patient outcomes~\cite{kuo2022health}.

Recent work has shown that generation conditioned on outcomes is crucial to the success of generating high-utility survival analysis. This can either be done by modelling the marginal distribution of the time and event pair~\cite{norcliffe2023survivalgan}, or by discovering the distinct patterns utilising an internal clustering algorithm~\cite{kuo2024ck4gen}. However, these approaches often prioritise realism and utility, overlooking the critical need for model calibration -- ensuring predicted risks align with observed outcomes.

\textbf{Stratified Calibration and Conditional Generation}\\
This paper emphasises calibration as a core objective in synthetic data generation, highlighting its role in improving clinical predictions, particularly in walled-garden scenarios with restricted data sharing. By targeting calibration, we address an often-overlooked aspect of synthetic data utility, offering benefits for equitable and reliable healthcare models.

Synthetic data generation differs from data augmentation in its approach and purpose. While synthetic data generation replicates the statistical properties of original datasets to create entirely new datasets, data augmentation focuses on addressing gaps or imbalances within datasets~\cite{dempster1977maximum}. Traditional synthetic methods often faithfully reflect and perpetuate existing imbalances, such as skewed sex distributions, limiting their utility for augmentation. To overcome this, conditional data generation~\cite{micheletti2023generative, kuo2024masked} allows the creation of synthetic patients with predefined covariate subsets, enabling targeted augmentation to generated data of specific characteristics for stratified calibration.

%###===>>>%###===>>>%###===>>>
%###===>>>%###===>>>%###===>>>
%###===>>>%###===>>>%###===>>>
%\newpage
\section{Methods}
Our \textbf{Masked Clinical Modelling (MCM)} framework adapts principles from masked language modelling, widely used in natural language processing (e.g., BERT~\cite{devlin2018bert}), to clinical data. As illustrated in Figure \ref{fig:Pipeline}, MCM masks selected clinical features (\textit{e.g.,} age, sex, event outcomes) in a dataset and predicts the missing values using the contextual relationships among unmasked features.

The workflow begins with preprocessing, where raw clinical data is scaled into a standardised range. During training, the model receives masked input features and reconstructs the missing information, leveraging patterns in the remaining data. After reconstruction, postprocessing is applied to revert the features to their original scales. MCM is optimised using mean square error loss to ensure accurate reconstruction of variables.

The internal components of the MCM model, including attention layers, residual connections, and masking mechanisms, are detailed in Section \ref{Sec:ModArc}. Section \ref{Sec:Training} elaborates on the training procedure. Section \ref{Sec:CKDStrat} discusses stratification of the CKD EHR dataset for modelling. Section \ref{Sec:MCMGen} explores MCM’s application in standalone synthetic data generation and conditional data augmentation. Last, we provide a comparative analysis between the baseline methods in the Results Section to our proposed MCM framework in Section \ref{Sec:ComparingWithBaselines}. Further details on preprocessing and postprocessing are provided in Appendix~\ref{App:Processing}.

%###===>>>%###===>>>%###===>>>
%###===>>>%###===>>>%###===>>>
%###===>>>%###===>>>%###===>>>
%\newpage
\subsection{Model Architecture and Mechanism}\label{Sec:ModArc}

Our MCM framework integrates attention mechanisms and multilayer perceptrons (MLPs) within a two-block architecture. For clarity, we occasionally use the subscript \( _{\{B\}} \) to denote block membership, but it is omitted where unnecessary. Unless explicitly stated otherwise, all input features referenced in this section correspond to preprocessed data. Note that the codes are publicly available$^{\ref{Ref:OpenSource}}$.

\subsubsection{Attention Layers}

\begin{figure}[h]
    \centering
    \includegraphics[width=\linewidth]{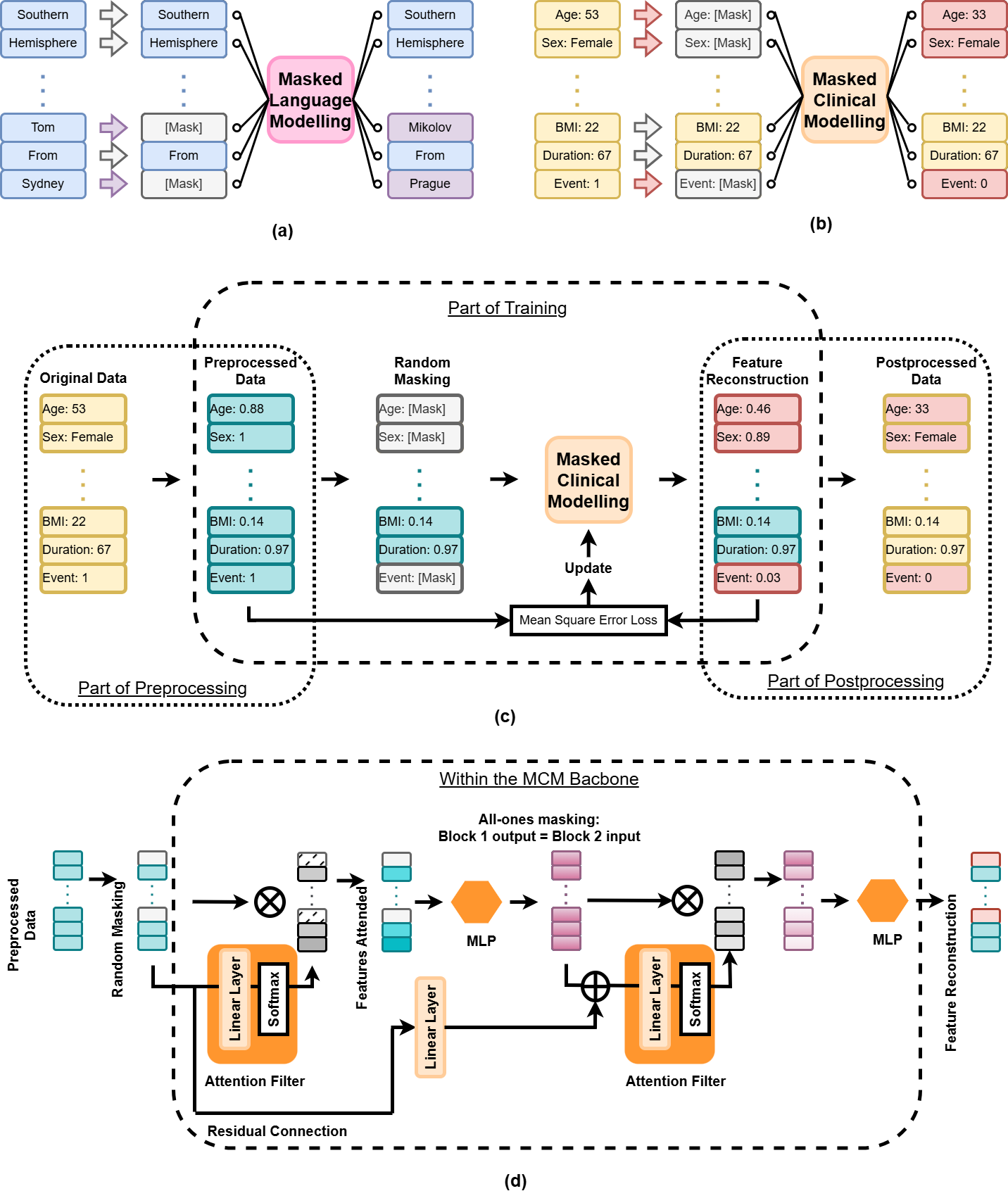}
    \caption{An overview of our masked clinical modelling framework.\\
    Subfigure (a) illustrates the principles of masked language modelling, where random words in a sentence are masked and predicted. Subfigure (b) adapts this to MCM, masking random features in clinical data and reconstructing them. Subfigure (c) outlines the engineering pipeline: (1) scaling the raw clinical data to a standardised range; (2) applying random masking; (3) predicting missing values using the remaining features as context; and (4) rescaling features to their original ranges. Subfigure (d) presents the internal model architecture, comprising attention layers, residual connections, and linear layers, which enable the framework to capture dependencies between clinical covariates.}
    \label{fig:Pipeline}
\end{figure}

Given a batch of input features \( \mathbf{X} \in \mathbb{R}^{N \times D} \), where \( N \) is the batch size and \( D \) is the feature dimension, an attention layer computes feature-wise weights through a learnable linear transformation. Specifically, the attention scores are calculated as \( \mathbf{A} = \mathbf{X} \mathbf{W}_{\text{att}} \), where \( \mathbf{W}_{\text{att}} \in \mathbb{R}^{D \times D} \) is a learnable weight matrix. In the first block, a binary mask \( \mathbf{M} \in \{0, 1\}^{N \times D} \) is applied to exclude features withheld for masked clinical modelling, as depicted in Figure~\ref{fig:Pipeline}(b). This masking mechanism ensures that attention scores for withheld features are set to \( -\infty \), preventing their influence on

\newpage
subsequent computations. The masked attention scores are represented as:
\begin{align}\label{Eq:Masked}
\mathbf{A}'_{ij} =
\begin{cases}
\mathbf{A}_{ij}, & \text{if } \mathbf{M}_{ij} = 1, \\
-\infty, & \text{if } \mathbf{M}_{ij} = 0.
\end{cases}
\end{align}

%###===>>>%###===>>>%###===>>>
%###===>>>%###===>>>%###===>>>
%###===>>>%###===>>>%###===>>>
Attention weights are then computed using the \texttt{Softmax} function~\cite{bishop2006pattern}, defined as:
\begin{align}\label{Eq:Attention}
\mathscr{A}_{ij} = \texttt{Softmax}(\mathbf{A}') = \frac{\texttt{exp}(\mathbf{A}'_{ij})}{\sum_{k=1}^D \texttt{exp}(\mathbf{A}'_{ik})},
\end{align}
where \( \texttt{exp} \) represents the exponential function. The attention-weighted features are subsequently computed via element-wise multiplication:
\begin{align}\label{Eq:Attneded}
\mathbf{Y} = \mathscr{A} \odot \mathbf{X},
\end{align}
where \( \odot \) denotes the Hadamard product.

In the second block, residual attention is employed to retain original feature information while refining it further. The input to this block is computed as 
\begin{align}\label{Eq:MLPed}
\mathbf{Z} = \mathbf{H} + \texttt{ReLU}(\mathbf{X} \mathbf{W}_{\text{res}}) \text{ , where } \mathbf{H} = \text{MLP}_{\{1\}}(\mathbf{Y}_{\{1\}}),
\end{align} 
\( \mathbf{W}_{\text{res}} \in \mathbb{R}^{D \times H} \) is a learnable weight matrix, and \( H \) is the hidden dimension. The rectified linear unit (\texttt{ReLU}) activation~\cite{nair2010rectified} ensures non-linear transformations. The second block does not apply \( -\infty \) masking, as feature withholding is restricted to the first block. This design parallels the encoder mechanism in the initial layer of Transformers for text processing in NLP~\cite{vaswani2017attention}.

\subsubsection{Multilayer Perceptrons}\label{Sec:MLP}

Each block incorporates an MLP to refine the attention-weighted features. The MLP operation, denoted as \(\Gamma(\mathscr{X})\), is defined as follows:
\begin{align}\label{Eq:Gamma}
\Gamma(\mathscr{X}) = \mathbf{W}_2 \left( \texttt{LayerNorm}\left( \texttt{ReLU}\left( \mathbf{W}_1 \mathscr{X} + \mathbf{b}_1 \right) \right) \right) + \mathbf{b}_2,
\end{align}
where \( \mathbf{W}_1 \in \mathbb{R}^{D \times H} \), \( \mathbf{W}_2 \in \mathbb{R}^{H \times H} \), and \( \mathbf{b}_1, \mathbf{b}_2 \in \mathbb{R}^H \) represent the weights and biases in Block 1. In Block 2, all weight matrices are \( \in \mathbb{R}^{H \times H} \). Layer Normalisation (\texttt{LayerNorm})~\cite{lei2016layer} is applied after the first \texttt{ReLU} activation in both blocks to stablise training.

The output activations of the MLPs are derived directly from \(\Gamma\), with their functional role tailored to the specific block. In Block 1, the final output uses a \texttt{ReLU} activation followed by \texttt{LayerNorm} for representation learning. In Block 2, a \texttt{Sigmoid} activation~\cite{han1995influence} is applied to \(\Gamma(\mathscr{X})\), ensuring that the reconstructed outputs, denoted as \( \mathbf{V} \), are constrained within the range \([0, 1]\), consistent with the preprocessed data range described in Appendix~\ref{App:PreProcessing}.

\subsubsection{Design Decisions}\label{Sec:DesignChoice}

The architecture deliberately avoids unnecessary complexity. Advanced features such as multi-head attention~\cite{vaswani2017attention}, GELU activations~\cite{hendrycks2016gaussian}, and alternative normalisation techniques, such as instance normalisation~\cite{ulyanov2016instance}, were tested but showed no significant performance improvements. Similarly, increasing the number of blocks or employing specialised initialisation strategies, such as Kaiming initialisation~\cite{he2015delving}, yielded negligible gains. 

\subsection{Training Process}\label{Sec:Training}

The training procedure is shown in Algorithm \ref{alg:training_mcm}, and it is designed to reconstruct missing data with varying levels of sparsity. To achieve this, the framework employs a dynamic masking strategy, which randomly conceals between 10\% and 95\% of the features for each instance in the batch during each training iteration. The training process explicitly optimises the reconstruction of masked features.

Our MCM backbone network is composed of two primary components, an attention filter (AF) and a multilayer perceptron (MLP), and can be expressed in the following (blockless) shorthand notation:
\begin{align}
F_{\text{MCM}} = F_{\text{AF}} \circ F_{\text{MLP}}.
\end{align}

%###===>>>%###===>>>%###===>>>
%###===>>>%###===>>>%###===>>>
%###===>>>%###===>>>%###===>>>
\newpage

%###===>>>%###===>>>%###===>>>
%###===>>>%###===>>>%###===>>>
%###===>>>%###===>>>%###===>>>
\newpage
\begin{algorithm}[h]
\caption{Training the MCM Framework with Dynamic Masking}
\label{alg:training_mcm}
\begin{algorithmic}[1]
\Require Preprocessed dataset \( \mathbf{X} \), masking range \([r_{\text{min}} = 10\%, r_{\text{max}}= 95\%]\), and Model \( F_{\text{MCM}} \)

\Procedure{Dynamic Masking and Training}{}
    \For{each epoch}
        \For{each batch}
            \State \textbf{Generate Random Masks \( \mathbf{M} \)}: Masking proportion sampled from \( U(r_{\text{min}}, r_{\text{max}}) \).
            \State \textbf{Apply Mask to Data}: 
            Conceal features in \( \mathbf{X} \) where masked features are set to 0.
            \State \textbf{Forward Pass}:
            \( \mathbf{V} = F_{\text{MCM}} (\mathbf{X}, \mathbf{M}) \) 
            \State \textbf{Compute Loss}: 
            $L = \frac{1}{N} \sum_{i=1}^N \sum_{j=1}^D \mathbf{M}_{ij} \left( \mathbf{V}_{ij} - \mathbf{X}_{ij} \right)^2$
            \State \textbf{Update the MCM backbone weights}
        \EndFor
    \EndFor
\EndProcedure
\end{algorithmic}
\end{algorithm}

Simplifying from Equations \eqref{Eq:Masked}-\eqref{Eq:Attention},  the attention filter computes the attention scores
\begin{align}
\mathscr{A} = F_{\text{AF}}(\mathbf{X}, \mathbf{M})
\end{align}
with the input features \( \mathbf{X} \) and the binary mask \( \mathbf{M} \) that indicates which features are withheld. Likewise, we simplify Equations \eqref{Eq:Attneded}-\eqref{Eq:MLPed} to represent the processing of the attention-weighted features as
\begin{align}
\mathbf{V} = F_{\text{MLP}}(\mathscr{A} \odot \mathbf{X}).
\end{align}

We use \( \mathbf{V} \) to denote the reconstructed features. Training is focused explicitly on the masked features, with the mean squared error loss defined as:
\begin{align}\label{Eq:MSELoss}
L = \frac{1}{N} \sum_{i=1}^N \sum_{j=1}^D \mathbf{M}_{ij} \left( \mathbf{V}_{ij} - \mathbf{X}_{ij} \right)^2,
\end{align}
where \( \mathbf{M}_{ij} \) ensures that only the masked features contribute to the loss.

In our experiments, we found that the MCM framework eliminates the need for auxiliary losses commonly used in other synthetic data generation frameworks, such as the Pearson correlation loss in GAN-based approaches~\cite{kuo2022health} or the random projection loss employed in diffusion models~\cite{salimans2016improved, kuo2023synthetic}. We attribute this to the inherent design of MCM, which leverages the attention mechanism to effectively capture data dependencies, making it naturally suited for discovering underlying correlations in healthcare data. Additionally, while the variable masking strength ranges from 10\% to 95\%, we observed that the framework does not require curriculum learning schedules~\cite{bengio2009curriculum}. The model’s backbone trains smoothly and exhibits stable convergence.

We employ the Adam optimiser~\cite{kingma2014adam} with a learning rate of 0.001, over a total of 500 epochs. The model employs a hidden dimension of \( H = 64 \) within its MLP layers. Following training, the framework imputes missing data by applying its learned attention and reconstruction mechanisms. To ensure consistency with the original feature distributions, postprocessing steps are performed, including reversing Box-Cox transformations and rescaling, as detailed in Appendix~\ref{App:PostProcessing}.

%###===>>>%###===>>>%###===>>>
%###===>>>%###===>>>%###===>>>
%###===>>>%###===>>>%###===>>>
%\newpage
\subsection{Stratification of Variables in Chronic Kidney Disease Analysis}\label{Sec:CKDStrat}

In our analysis of CKD, we employed binary stratification variables to evaluate their impact on disease progression and outcomes. These variables were selected based on their clinical significance and methodological considerations. Given the limited sample size of 491 individuals in the Al-Shamsi \textit{et al.} EHR's (see Section~\ref{Sec:CKD}), we adopted binary stratifications to ensure robust convergence in downstream CoxPH modelling. Below, we provide a detailed rationale for each stratification.

\subsubsection{Estimated Glomerular Filtration Rate (eGFR) Stratification}

eGFR is a fundamental biomarker for assessing renal function and diagnosing CKD~\cite{levey2005definition}. It reflects the kidneys' ability to filter waste and fluids; reductions in eGFR are associated with increased risks of cardiovascular events, progression to end-stage renal disease, and mortality~\cite{chronic2010association}. According to the Kidney Disease: Improving Global Outcomes (KDIGO) guidelines, eGFR is pivotal for CKD staging and management~\cite{levey2005definition}.

\newpage
For this study, we stratified eGFR into two categories: \textit{Normal} ($\geq 90$ mL/min/1.73~m$^2$) and \textit{Non-Ideal} ($<90$ mL/min/1.73m$^2$). An eGFR below 90mL/min/1.73m$^2$ may indicate kidney damage even in the absence of other clinical signs, highlighting its importance in identifying at-risk individuals. This binary categorisation simplifies the analysis while maintaining clinical relevance, allowing us to focus on the presence of reduced renal function as a significant risk factor.

\subsubsection{Diabetes Stratification}

Diabetes mellitus is a leading cause of CKD worldwide~\cite{who_diabetes_factsheet}, making its stratification essential for understanding CKD progression. Chronic hyperglycemia leads to glomerular hyperfiltration and diabetic nephropathy, exacerbating renal damage and increasing the risk of complications.

To enhance the accuracy of diabetes classification, we employed a triangulation approach~\cite{chan2010electronic}, integrating multiple data points from EHRs to ensure maximal coverage. Patients were categorised into three groups: \textit{No Diabetes}, with no history of diabetes; \textit{Diabetes Without Medication}, diagnosed with diabetes but not undergoing pharmacological treatment; and \textit{Diabetes With Medication}, actively managing diabetes with medications. Recognising that the presence of diabetes itself is the predominant risk factor for CKD, we consolidated these categories into a binary stratification, distinguishing between patients with diabetes (regardless of treatment status) and those without.

\subsubsection{Hypertension Stratification}

Hypertension is a significant risk factor for CKD, contributing through mechanisms such as increased intraglomerular pressure leading to renal scarring and loss of function~\cite{kohagura2023public}. Additionally, CKD can exacerbate hypertension via fluid overload and activation of the renin-angiotensin-aldosterone system, creating a worsening cycle that accelerates disease progression~\cite{ku2019hypertension}.

We applied a similar triangulation approach for hypertension classification: \textit{No Hypertension}, no history of hypertension; \textit{Hypertension Without Medication}, diagnosed with hypertension but not receiving antihypertensive treatment; and \textit{Hypertension With Medication}, managing hypertension with medications such as ACE inhibitors or ARBs. For binary analysis, these groups were consolidated into a single indicator of hypertension presence versus absence.

\subsubsection{Age Stratification}

Age is a well-established, non-modifiable risk factor for CKD~\cite{ndumele2023cardiovascular}. Renal function declines with age due to structural and functional changes, such as nephrosclerosis and decreased nephron number~\cite{denic2016structural}. Older adults are also more susceptible to comorbidities like hypertension and diabetes, compounding the risk of renal dysfunction.

We employed a binary stratification of age into \textit{Younger} ($<65$ years) and \textit{Older} ($\geq 65$ years). This threshold aligns with demographic classifications and reflects shifts in physiological function and comorbidity burden observed in older populations~\cite{o2006mortality}. By using this binary categorisation, we can directly assess the impact of older age on CKD outcomes while maintaining sufficient statistical power in each group.

\subsubsection{Cardiovascular Disease Stratification}

Cardiovascular disease (CVD) and CKD share common risk factors and pathophysiological mechanisms, including hypertension, diabetes, dyslipidemia, chronic inflammation, and endothelial dysfunction~\cite{weiner2006cardiovascular}. The presence of CVD significantly heightens morbidity and mortality among CKD patients, emphasising its role as a critical comorbid condition.

To assess the impact of CVD on CKD outcomes, we employed a binary stratification. Patients were classified as having \textit{CVD} if they had a documented history of coronary heart disease, heart failure, stroke, or peripheral vascular disease. Those without such history were categorised as \textit{No CVD}.

\subsubsection{Stratification Statistics and Their Role in Data Augmentation}

Table \ref{tab:ckd_stratification} summarises the key stratifications used in our analysis, which are crucial for assessing the effectiveness of conditional data augmentation and stratified calibration in downstream CoxPH modelling. After applying our stratification stratigy, 32.79\% of patients were found to have non-ideal eGFR values, 43.79\% had diabetes, 68.23\% were classified as hypertensive, 22.81\% belonged to the older cohort, and 13.85\% had a history of CVD.

These distributions reveal a spectrum of balance, ranging from near-equilibrium (\textit{e.g.,} 43.79\% with diabetes) to substantial imbalance (\textit{e.g.,} 13.85\% with CVD). This variability offers an ideal testbed for evaluating MCM’s ability to perform conditional data augmentation, generating synthetic patients with specific characteristics to enhance stratified calibration. The inherent imbalance across subgroups 

\begin{table}[h!]
    \centering
    \caption{Stratification Variables and Proportions in CKD Analysis}
    \label{tab:ckd_stratification}
    \begin{tabular}{p{2cm}p{3.75cm}p{6.25cm}}
        \hline
        \textbf{Stratification} & \textbf{Proportion} & \textbf{Notes on Stratification} \\
        \hline
        \hline
        
        \textbf{eGFR} & 
        Normal (67.21\%)\newline 
        Non-Ideal (32.79\%) & eGFR values were stratified into\newline
        ``Normal'' ($\ge$ 90 mL/min/1.73 m²) and\newline
        ``Non-Ideal'' ($<$ 90 mL/min/1.73 m²).\\
        \hline
        
        \textbf{Diabetes} & 
        No Diabetes (56.21\%)\newline
        Diabetes (43.79\%) & 
        Determined through triangulation: patients were first categorised as having\newline
        ``No Diabetes'', ``Diabetes Without Medication'', or ``Diabetes With Medication'';\newline
        and then turned into a binary stratification. \\
        \hline
        
        \textbf{Hypertension} & 
        No Hypertension (31.77\%)\newline 
        Hypertension (68.23\%) & 
        Hypertension status first incorporated medication use: ``No Hypertension'', ``Hypertension Without Medication'', or `Hypertension With Medication'', using ACE inhibitors or ARBs as a marker; and then turned into a binary stratification. \\
        \hline
        
        \textbf{Age} & 
        Younger (77.19\%)\newline
        Older (22.81\%) & 
        Age was stratified into\newline
        ``Younger'' ($<$65 years) and\newline
        ``Older'' ($\ge$65 years). \\
        \hline
        
        \textbf{CVD} & 
        No CVD (86.15\%)\newline
        CVD (13.85\%) & 
        CVD status was based on a history of coronary heart disease or vascular disease/ \\
        \hline
    \end{tabular}
\end{table}

also presents a challenge for MCM to generate standalone datasets that accurately reflect the real-world distributions, ensuring the synthetic data captures the nuanced imbalances.

%###===>>>%###===>>>%###===>>>
%###===>>>%###===>>>%###===>>>
%###===>>>%###===>>>%###===>>>
%\newpage
\subsection{Generating Data with MCM}\label{Sec:MCMGen}

Below we elaborate on the differences between data synthesis and conditional data generation.

\subsubsection{Data Synthesis}
\begin{algorithm}[h]
\caption{Data Synthesis with the Trained MCM Framework}
\label{alg:data_synthesis}
\begin{algorithmic}[1]
\Require Preprocessed dataset \( \mathbf{X} \), trained model \( F_{\text{MCM}} \), masking ratio \( r = 50\% \)

\Procedure{Synthetic Data Generation}{}
    \State \textbf{Initialise Synthetic Dataset}: Create a copy of the preprocessed dataset \( \mathbf{X}_{\text{synth}} \gets \mathbf{X} \)

    \For{each instance in \( \mathbf{X}_{\text{synth}} \)}
    \State \textbf{Generate Mask}: Mask \( k = \lfloor r \cdot D \rfloor \) features
        
    \State \textbf{Reconstruct Missing Features}:\( \mathbf{V} = F_{\text{MCM}}(\mathbf{X}_{\text{synth}}, \mathbf{M}) \)
        
    \State \textbf{Update Dataset}: \( \mathbf{X}_{\text{synth}} \gets \mathbf{M} \odot \mathbf{X}_{\text{synth}} + (1 - \mathbf{M}) \odot \mathbf{V} \)
    \EndFor
    
    \State \textbf{Postprocess Synthetic Dataset}
\EndProcedure
\end{algorithmic}
\end{algorithm}

Once the MCM framework has been trained, it is employed to generate synthetic datasets. The pseudocode is provided in Algorithm \ref{alg:data_synthesis}. The process begins with a copy of the entirety of the preprocessed original dataset, which serves as the foundation for synthesis.

To introduce variability and enforce feature reconstruction, a binary mask is dynamically applied to the dataset. This procedure mirrors the principles of error-bounded lossy compression~\cite{jin2022improving, underwood2024understanding}, wherein controlled distortions are introduced to the data while preserving critical information. For

\newpage
each instance, a fixed proportion of features is randomly selected for masking. The masking ratio, which we default to 50\% in this application\footnote{In our experiments, MCM demonstrated the ability to generate novel data points even with up to 80\% of an individual’s features masked. However, this paper focuses on showcasing MCM’s capability for generating standalone synthetic datasets and performing conditional data augmentation to enhance stratified calibration in downstream CoxPH models. Exploring the impact of varying masking setups is beyond the scope of this work and will be addressed in future studies. Further note that we can generate an unlimited amount of synthetic data using Algorithm \ref{alg:data_synthesis}; however, we default our study towards generating 1x the amount of real patient data (\textit{i.e.,} to generate 491 synthetic patient data, when applied for data synthesis.}, ensures that approximately half of the features for each instance are concealed. Crucially, each patient’s features are masked differently.

The description below follows notations in Equations \eqref{Eq:Masked} - \eqref{Eq:MSELoss}. The reconstruction is formalised as:
\begin{align}
\mathbf{V} = F_{\text{MCM}}(\mathbf{X}, \mathbf{M}).
\end{align}
The reconstructed features in \( \mathbf{V} \) are then integrated back into the dataset, replacing the masked entries, while unmasked features remain unchanged:
\begin{align}
\mathbf{\hat{X}} \gets \mathbf{M} \odot \mathbf{X} + (1 - \mathbf{M}) \odot \mathbf{V}.
\end{align}
Afterwards, the postprocessing steps outlined in Appendix~\ref{App:PostProcessing} are applied to rescale the variables to their original distribution range.

%###===###
%\newpage
\subsubsection{Enhanced Calibration for Targeted Stratified Groups}\label{Sec:EnhancedCalibration}
\begin{algorithm}[h]
\caption{5x2 Cross-Validation with Enhanced Stratified Risk Calculation Using Simulated Data}
\label{alg:StratifiedCalibration}
\begin{algorithmic}[1]
\Require Clinical dataset \( \mathcal{D} \) with covariates \( \mathbf{X} \) and event information, and set of time points \( \mathcal{T} \),\\
\hspace*{9mm}stratification target variable $Z^\dagger$

\Procedure{5x2 Cross-Validation and Enhanced Risk Calculation}{}
    \For{\colorbox{cyan!25}{each iteration \( i = 1, \dots, 5 \)}}
    
        \For{each fold \( f = 1, \dots, 5 \)}
        
            \For{each split \( s = 0, 1 \)}
            \State \textbf{Split Dataset}: 
                Create training set \( \mathcal{D}_{\text{train}} \) and test set \( \mathcal{D}_{\text{test}} \) based on \( f \) and \( s \)

                \State \colorbox{orange!25}{\textbf{Generate simulated dataset}:}\\
                \hspace*{25mm}\colorbox{orange!25}{1) Filter all data in the training dataset that suffice condition $Z^\dagger$}\\
                \hspace*{29mm}\colorbox{orange!25}{(\textit{e.g.,} condition age is young ($<65$ years, see Table \ref{tab:ckd_stratification}))}\\
                \hspace*{25mm}\colorbox{orange!25}{2) Apply a simulated data approach on the filtered dataset,}\\
                \hspace*{29mm}\colorbox{orange!25}{such as MCM with Algorithm \ref{alg:data_synthesis},}\\
                \hspace*{29mm}\colorbox{orange!25}{to create} \colorbox{cyan!25}{the current iteration} \colorbox{orange!25}{of conditional dataset \( \mathcal{D}_{\text{sim}, Z^\dagger, a, i} \)}
                        
                \State \colorbox{orange!25}{\textbf{Update Training Data}: 
                Append \( \mathcal{D}_{\text{train}} \gets \mathcal{D}_{\text{train}} \cup \mathcal{D}_{\text{sim}, Z^\dagger, a, i} \)}
                
                \State \textbf{Fit CoxPH Model}
                
                \State \textbf{Compute LPH for Test Set}
            
            \EndFor
        \EndFor

        \State \textbf{Aggregate LPH}: 
        \colorbox{cyan!25}{For each ith iteration}, compute the mean LPH across folds and splits,\\
        \hspace*{36mm}for each \( t \in \mathcal{T} \).
    \EndFor \colorbox{cyan!25}{- - -}
\EndProcedure
\end{algorithmic}
\end{algorithm}

Algorithm~\ref{alg:StratifiedCalibration} outlines the enhanced calibration process for targeted stratified groups, building on the risk calculation in Algorithm~\ref{alg:Risks}. This enhancement arises from the need to fit the CoxPH model to compute LPH values, which are essential for deriving predicted risks \( \mathscr{R}_{\text{perc}} \) (see Algorithm~\ref{alg:Calibration}).

The process begins by pre-training the MCM framework on the entire dataset to learn the statistical relationships and dependencies in the data. Simulated data is not generated upfront as a standalone dataset; instead, it is created dynamically during each fold and split of the cross-validation (line 6, Algorithm~\ref{alg:StratifiedCalibration}). This ensures that the simulated data aligns with the characteristics of the current training set and the stratification condition \( Z^\dagger \).

We refer to this process as \textit{simulation} rather than \textit{synthesis} to emphasise that the generated data augments the current training dataset rather than forming a standalone dataset. Simulated data is derived only from the training dataset of the current fold and split, avoiding any exposure to test data and thus eliminating potential data leakage (line 7). This safeguards the integrity of the evaluation and the robustness of the downstream CoxPH model.

The simulated data for the stratified condition \( Z^\dagger \) is appended to the original training data (line 13), forming an augmented dataset used to fit the CoxPH model (line 14). This step ensures that the model is trained on a dataset that better represents the stratified group of interest. To increase robustness and account for variability in the simulation process, the procedure is repeated for five iterations (\( i = 1, \dots, 5 \)), as shown in the iterative loop (lines 3 and 19). This approach enables the computation of error bars\footnote{Each of the 5 iterations builds on the 5x2 cross-validation framework, resulting in a total of 50 variations. In each iteration, for the current fold, the process involves training on split 0 and testing on split 1, followed by training on split 1 and testing on split 0.}, improving the reliability of the calibration results.

The pseudocode is colour coded for clarity: \colorbox{cyan!25}{blue} represents the iterative loop for robustness, while \colorbox{orange!25}{orange} highlights the steps for conditional data simulation. This design is adaptable to various simulation methods. Line 11 in the pseudocode can seamlessly accommodate MCM as well as alternatives like SMOTE or VAEs, allowing for fair comparisons across data simulation methods.

As outlined in line 11 of Algorithm~\ref{alg:StratifiedCalibration}, our default approach for simulating conditional data employs MCM following the methodology detailed in Algorithm~\ref{alg:data_synthesis}. However, alternative configurations can be explored to further optimise the outcomes of conditional stratification. These alternatives and their impact will be discussed in detail in the Results section, with additional implementation variations provided in Appendix \ref{App:McmMice} for readers seeking further insights.

%###===>>>%###===>>>%###===>>>
%###===>>>%###===>>>%###===>>>
%###===>>>%###===>>>%###===>>>
%\newpage
\subsection{Comparison of Synthetic Data Generation Methods and Frameworks}\label{Sec:ComparingWithBaselines}

\begin{table}[h]
\tiny
    \centering
    \caption{Comparison of Synthetic Data Generation Methods}
    \label{tab:methods_comparison}
    \begin{tabular}{|p{2cm}||c|c|c|p{3.5cm}|}
        \hline
        \textbf{Method} & \textbf{Data Augmentation} & \textbf{Dataset Synthesis} & \textbf{Conditional Generation} & \textbf{Training Notes} \\
        \hline
        \hline
        MCM (Ours) & \cmark & \cmark & \cmark & \\
        \hline
        \multicolumn{5}{l}{\tiny{\textit{Data Augmentation Techniques}}}\\
        \hline
        Random Oversampling & \cmark & \xmark & \xmark & \\
        SMOTE & \cmark & \xmark & \xmark & \\
        SMOTENC & \cmark & \xmark & \xmark & \\
        ADASYN & \cmark & \xmark & \xmark & \\
        BorderlineSMOTE & \cmark & \xmark & \xmark & \\
        SVMSMOTE & \cmark & \xmark & \xmark & \\
        \hline
        \multicolumn{5}{l}{\tiny{\textit{Deep Learning Generative Models}}}\\
        \hline
        VAE & \cmark & \cmark & \xmark & suffers from posterior collapse\\
        &&&&\\
        WGAN & \cmark & \cmark & \xmark & suffers from convergence instability\newline suffers from model collapse \\
        CK4Gen & \cmark & \cmark & \xmark & multi-phased non end-to-end training\newline requires knowledge distillation\\
        \hline
    \end{tabular}
\end{table}

Random oversampling addresses data imbalance by replicating underrepresented instances, offering simplicity and ease of use. SMOTE~\cite{chawla2002smote} and its derivatives (SMOTENC~\cite{chawla2002smote}, ADASYN~\cite{he2008adasyn}, BorderlineSMOTE~\cite{han2005borderline}, SVMSMOTE~\cite{nguyen2011borderline}) interpolate between data points for augmentation but fail to synthesise realistic standalone datasets as they cannot capture complex data distributions.

VAEs~\cite{kingma2014auto}, GANs~\cite{goodfellow2014generative} (and Wasserstein GAN [WGAN]~\cite{arjovsky2017wasserstein}), and CK4Gen~\cite{kuo2024ck4gen} excel at standalone synthesis but lack conditional data generation. Without this capability, these models require complete retraining for each new subgroup, making them inefficient and unsuitable for robust assessments such as 5x2 cross-validation (line 10 of Algorithm \ref{alg:StratifiedCalibration}). CK4Gen, specifically designed for survival data, employs an autoencoder with clustering to identify latent risk profiles. However, its multi-step training pipeline, reliance on a pre-trained CoxPH model for knowledge distillation~\cite{hinton2015distilling}, and lack of end-to-end training make it complex and inflexible. Like VAEs and GANs, it cannot efficiently generate subgroup-specific data without retraining.

The MCM framework that we propose provides an elegant and scalable alternative. It supports both conditional data augmentation and standalone synthesis without requiring retraining or multi-phase pipelines. This adaptability makes MCM a simple yet powerful framework for generating synthetic data across diverse clinical settings and stratification scenarios. The comparative advantages are outlined in Table~\ref{tab:methods_comparison}.

%###===>>>%###===>>>%###===>>>
%###===>>>%###===>>>%###===>>>
%###===>>>%###===>>>%###===>>>
%\newpage
\section{Results}

The results section presents a streamlined, three-tiered technical validation framework designed to evaluate the effectiveness and practicality of the MCM framework. This includes a realism check, utility verification, and practicality analysis.

In the \textbf{realism check}, we employ Algorithm \ref{alg:data_synthesis} to generate a standalone synthetic dataset that mirrors the statistical and structural properties of the ground truth CKD EHR (Table \ref{Tab:CKDTab1}). The distributions of individual variables and their correlations are assessed to ensure fidelity to the original data.

The \textbf{utility verification} examines whether the synthetic dataset can replace the real dataset for survival analysis. Kaplan-Meier survival curves (KM curves)~\cite{kaplan1958nonparametric} and CoxPH models are constructed from both datasets. Hazard ratios (HRs, Equation \eqref{Eq:CoxPHHR}) are analysed to confirm the synthetic data’s ability to capture relative risks for progressing to CKD stages 3--5.

The \textbf{practicality analysis} evaluates the role of simulated datasets in improving stratified calibration of survival models under imbalanced scenarios (Table \ref{tab:ckd_stratification}). A large-scale comparison of 17 methods is conducted to determine their effectiveness in enhancing model precision across various stratifications.

%###===>>>%###===>>>%###===>>>
%###===>>>%###===>>>%###===>>>
%###===>>>%###===>>>%###===>>>
%\newpage
\subsection{Realism Check}\label{Sec:Check01}
This phase evaluates the static properties of the synthetic data by comparing its distributions and statistical moments, specifically the mean and variance, with those of the real data. Probability density functions of the numeric variables are overlaid using kernel density estimations (KDEs)~\cite{davis2011remarks} to provide a visual comparison. For binary variables, side-by-side histograms are employed to illustrate their distributions. Additionally, the correlations between variable pairs are calculated and compared across the synthetic and real datasets to ensure that inter-variable relationships are accurately preserved.

\begin{figure}[h]
    \centering
    \includegraphics[width=0.72\linewidth]{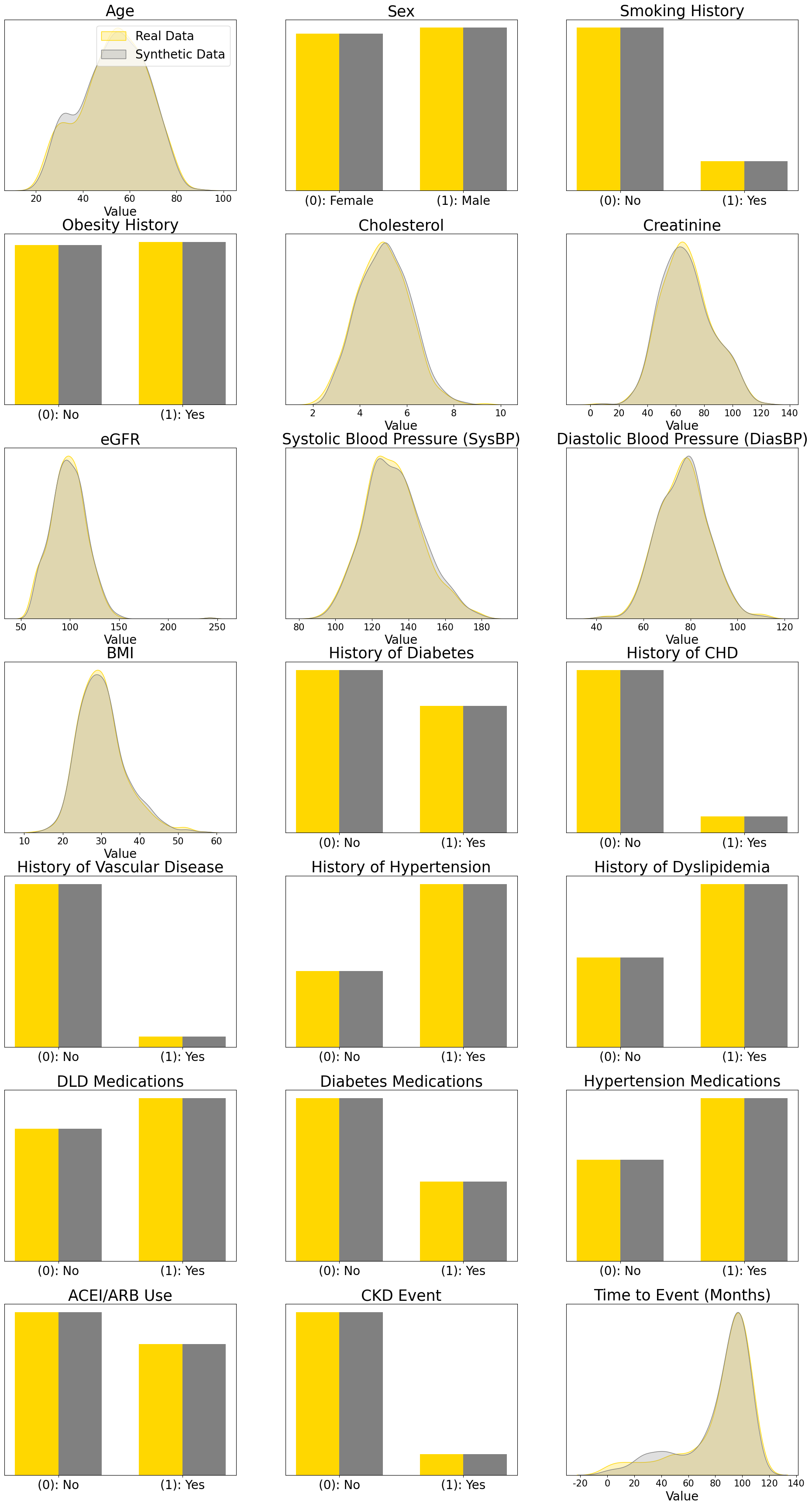}
    \caption{Comparison of real variables (gold) and synthetic counterparts (grey) from the CKD EHR dataset. Binary variables are visualised with histograms, while numeric variables are compared using kernel density estimations (KDEs).}
    \label{Fig:CkdDist_MCM}
\end{figure}

%#-----------------------------------------------
The distribution comparisons in Figure \ref{Fig:CkdDist_MCM} show gold representing the ground truth data and grey representing the synthetic data generated by MCM. No significant visual differences are observed between the two datasets, irrespective of the variable type or its statistical properties. This consistency holds across balanced binary variables like obesity history, imbalanced binary variables such as smoking status, the bowl-shaped distribution of cholesterol levels, and the long-tailed distribution\footnote{Notably, for the strictly positive time-to-event duration variable, some negative values appear due to the KDE algorithm, which employs Gaussian kernels with tails extending beyond the true data range. This is an artefact of the KDE method and does not reflect the data itself. The ground truth data set is publicly accessible~\cite{al2018chronic}, and the synthetic dataset will be made available for further examination after the paper is accepted. See more about the KDE issue at: \cite{stats_kde_negative}.} of time-to-event durations. The MCM framework effectively captures these diverse characteristics.
 
\begin{figure}[h]
    \centering
    \includegraphics[width=0.8\linewidth]{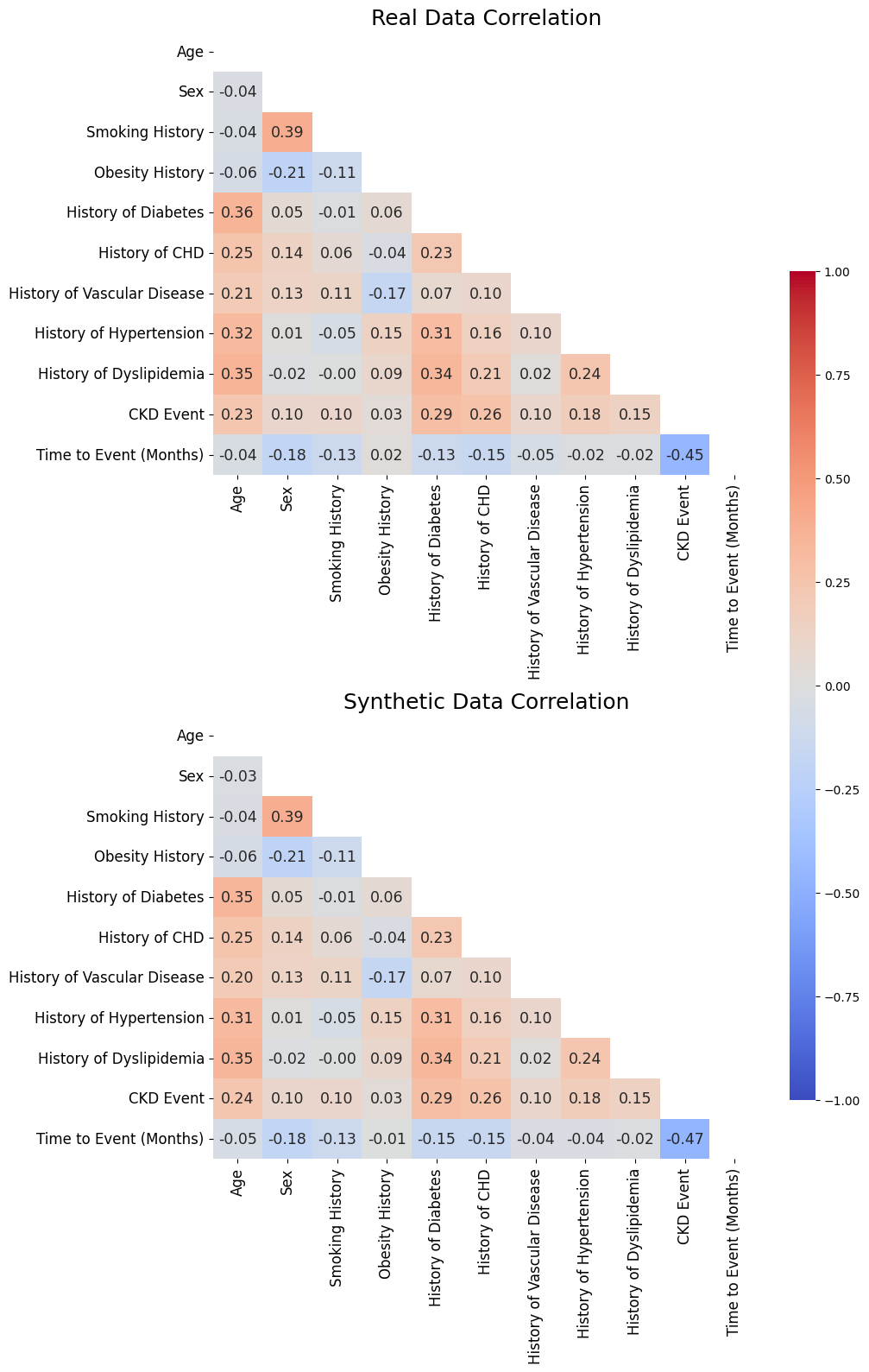}
    \caption{Comparison of correlations among selected variables from the CKD EHR dataset. The top panel shows real data, and the bottom panel displays synthetic data. Blue represents negative correlations, and red represents positive correlations.}
    \label{Fig:CorrPartial_MCM}
\end{figure}

\begin{figure}[h]
    \centering
    \includegraphics[width=0.88\linewidth]{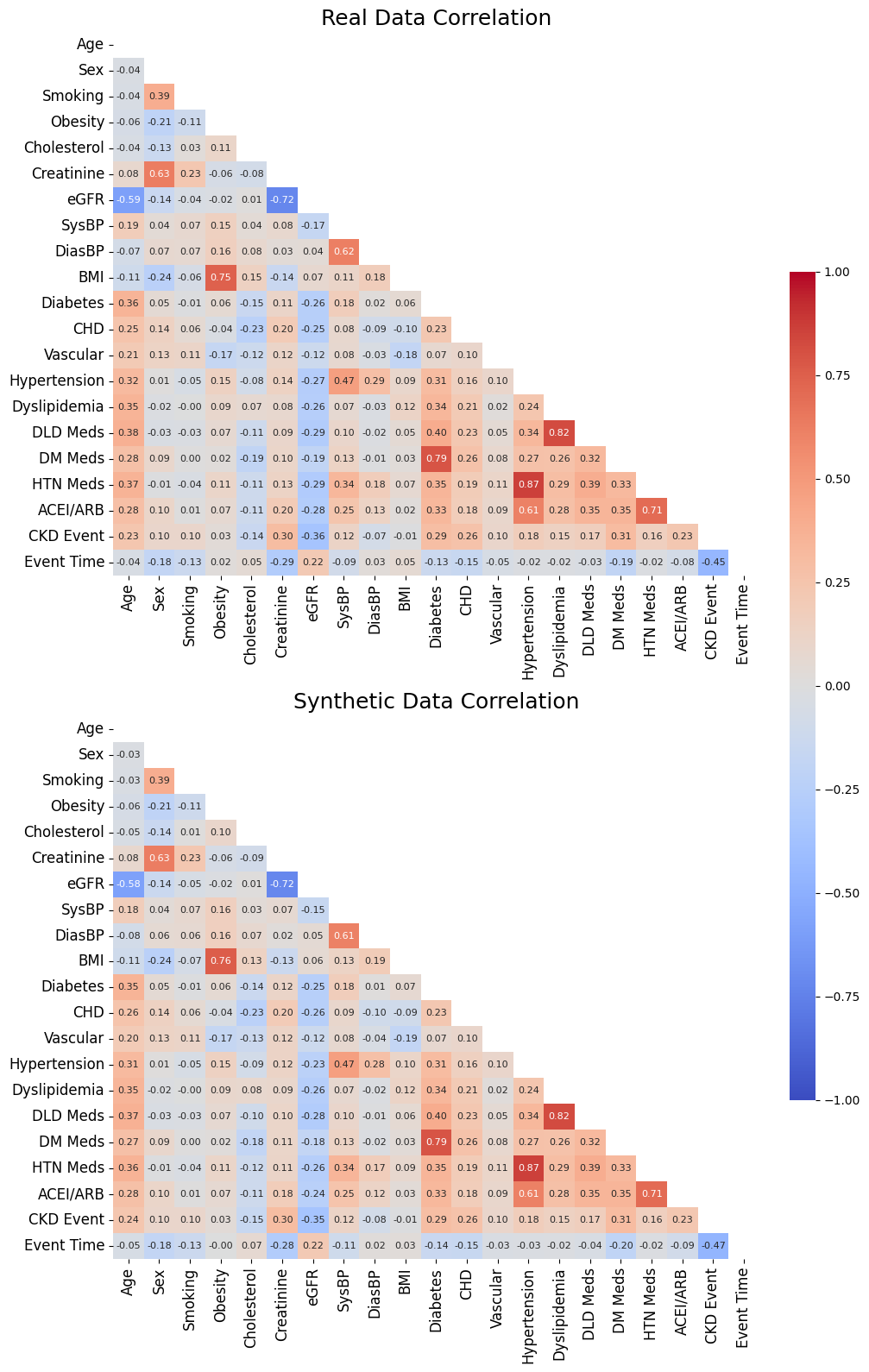}
    \caption{Complete Correlation.}
    \label{Fig:CorrFull_MCM}
\end{figure}

We next evaluate variable correlations to ensure inter-variable relationships are faithfully reproduced. Due to the extensive number of features in the CKD EHR, correlations are presented in two figures. Figure \ref{Fig:CorrPartial_MCM} focuses on demographics and lifestyle variables, as well as medical history and outcome; while Figure \ref{Fig:CorrFull_MCM} includes the complete set of clinical measurements. The synthetic data replicates key relationships observed in the ground truth, including strong negative correlations (\textit{e.g.,} eGFR and creatinine), strong positive correlations (\textit{e.g.,} hypertension history and hypertension medication use), and variables with minimal correlation (\textit{e.g.,} dyslipidaemia history and smoking status).

The observed alignment of single-variable distributions and inter-variable correlations demonstrates that MCM produces high-fidelity synthetic datasets, effectively preserving both independent and relational characteristics of the original data. Further results comparing distributions and correlations for VAEs, WGANs, and CK4Gen are provided in Appendix \ref{App:ExtraResults1}.

%#-----------------------------------------------
\newpage

%#-----------------------------------------------
\newpage

%#-----------------------------------------------
\newpage

%###===>>>%###===>>>%###===>>>
%###===>>>%###===>>>%###===>>>
%###===>>>%###===>>>%###===>>>
\newpage
%#-----------------------------------------------
%\newpage
\begin{figure}[h]
    \centering
    \includegraphics[width=\linewidth]{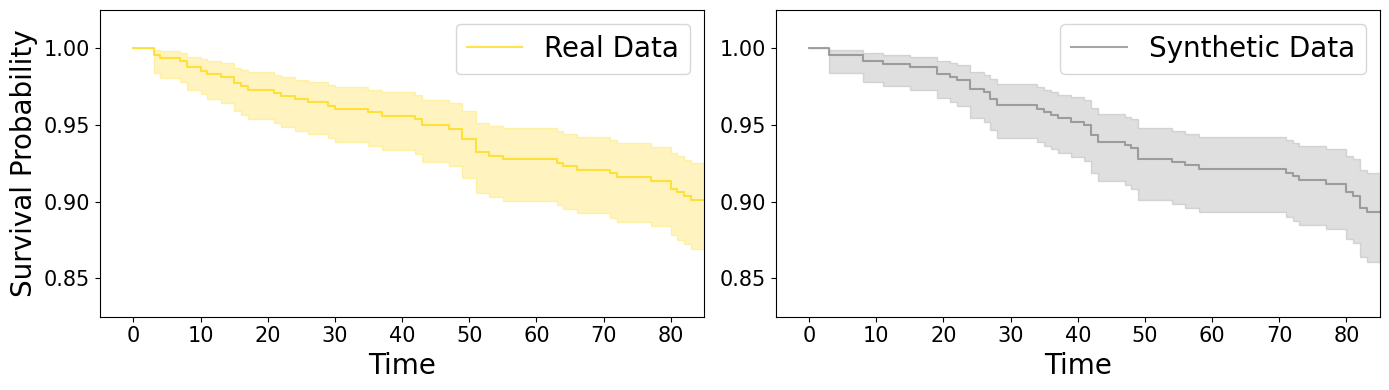}
    \caption{Comparison of Kaplan-Meier (KM) curves between real and synthetic data.}
    \label{Fig:KmCurve_MCM}
\end{figure}

%#-----------------------------------------------
\begin{figure}[h]
    \centering
    \includegraphics[width=\linewidth]{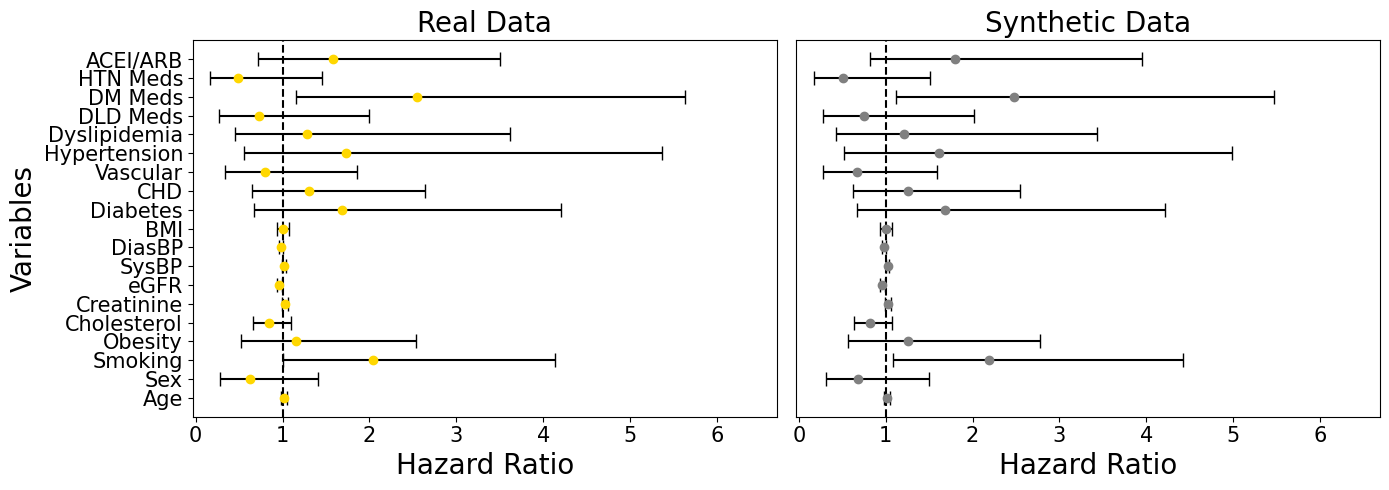}
    \caption{Comparison of hazard ratio (HR) consistencies between real and synthetic data.}
    \label{Fig:HrConsistency_MCM}
\end{figure}

%###===>>>%###===>>>%###===>>>
%###===>>>%###===>>>%###===>>>
%###===>>>%###===>>>%###===>>>
\subsection{Utility Verification}\label{Sec:Check02}
We evaluate the utility of MCM’s synthetic datasets using KM curves and HRs, two essential metrics in survival analysis. KM curves estimate survival probabilities over time, comparing synthetic and real data on final survival rate, slope, and variability. The final survival rate indicates the proportion of survivors at the end of observation, reflecting long-term risks. The slope reveals event rates, with steeper slopes indicating higher frequencies, while variability captures dynamic changes in risk. These features test the synthetic data’s ability to replicate temporal survival patterns.

HRs from CoxPH models validate the association between predictors and event risks. Point estimates assess the strength of associations, while confidence interval (CI) spreads indicate uncertainty. CIs including 1 suggest no significant effect, while those entirely above or below 1 signal increased or protective risks. Alignment of HRs and CIs between real and synthetic data confirms MCM’s fidelity in capturing associations and uncertainties.

The results in Figure \ref{Fig:KmCurve_MCM} demonstrate close alignment between the KM curves of real and synthetic data. The synthetic data accurately captures gradients, variability, and final survival probabilities, replicating the temporal survival patterns of the original dataset. The comparison of HRs in Figure \ref{Fig:HrConsistency_MCM} further validates the reliability of MCM. The point estimates align closely, preserving the strength of associations between predictors and survival outcomes. Additionally, the confidence intervals remain consistent, reflecting the uncertainty inherent in the EHR. All HRs maintain their correct positioning relative to the baseline hazard of 1.

The alignment of KM curves and HR consistency confirms that MCM generates synthetic datasets with significant utility. The synthetic data can effectively substitutes real data for downstream clinical research, preserving critical clinical characteristics when applied to survival models. Additional comparisons of KM curves and HR consistency for VAEs, WGANs, and CK4Gen are available in Appendix \ref{App:ExtraResults2}.

%###===>>>%###===>>>%###===>>>
%###===>>>%###===>>>%###===>>>
%###===>>>%###===>>>%###===>>>
\newpage
\subsection{Practicality Analysis}\label{Sec:Check03}

\begin{figure}[h]
    \centering
    \includegraphics[width=0.5\linewidth]{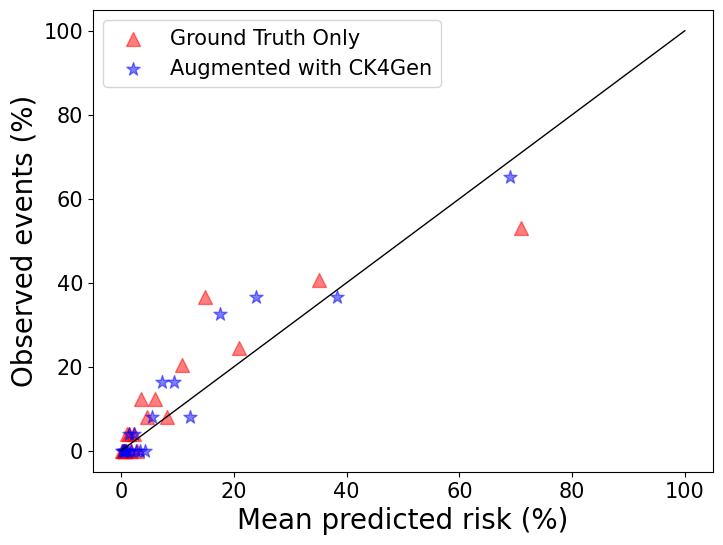}
    \caption{Calibration curves for the CKD EHR dataset comparing CoxPH model performance trained on ground truth data (red triangles) and models augmented with MCM synthetic data (blue stars) at the median follow-up time.}
    \label{Fig:GeneralCalibration}
\end{figure}

Calibration, as discussed in Section \ref{Sec:CoxPHCali}, measures how well predicted event probabilities align with observed outcomes. Calibration curves, plotting predicted versus observed risk at specific time points, ideally follow a 45-degree line. Figure \ref{Fig:GeneralCalibration} illustrates the general calibration at the median follow-up time, comparing CoxPH models trained solely on ground truth data with those augmented by MCM-simulated data. Models augmented with MCM data demonstrate better alignment with the ideal calibration line, indicating improved performance.

To quantitatively assess calibration, the slope of the calibration curve (closer to 1 is better) and the error (lower is better) were evaluated, as defined in Equations \eqref{Eq:CalSlope} and \eqref{Eq:CalLoss}. Calibration was further assessed at the 25th, 50th, and 75th percentiles of follow-up, reflecting the evolving nature of disease progression and ensuring robust risk predictions over time. Using the original CKD EHR data, the CoxPH model achieved calibration slopes of 0.9880, 0.8255, and 0.7710, corresponding to calibration losses of 0.0120, 0.1745, and 0.2290 at these time points, respectively. In contrast, the MCM-augmented data yielded slopes of 1.1161, 0.9164, and 0.8489, with calibration losses of 0.1161, 0.0836, and 0.1511, respectively. This indicates improved precision at mid and long follow-up durations with MCM augmentation, though performance at shorter durations slightly worsened.

As one of the first studies to evaluate simulated data for calibration, this work focuses primarily on the \textit{overall} calibration by summing losses across three time points. The overall calibration loss without augmentation was 0.4155, compared to 0.3508 with MCM augmentation (around 15\% loss reduction\footnote{$1 - 0.3508/0.4155 \approx 0.15$.\label{FN:01}}), demonstrating the practicality of employing generative models to enhance healthcare decision-making systems. Further exploration of calibration across different time-to-event durations will be addressed in future studies.

Table \ref{tab:meta_calibration_loss} provides a detailed comparison of calibration losses across the stratification subgroups outlined in Table \ref{tab:ckd_stratification} for CoxPH models trained using the data augmentation methods listed in Table \ref{tab:methods_comparison}. The table summarises the meta-calibration loss, which aggregates the overall calibration losses across all subgroups, offering a comprehensive measure of a method's ability to generalise across diverse patient populations. Methods are ranked by their total meta-calibration loss, lower loss is ranked higher. Rows are colour-coded for clarity: the baseline (no augmentation) is highlighted in \colorbox{green!20}{green}, MCM is marked in \colorbox{blue!20}{blue}, and the top-performing method is highlighted in \colorbox{purple!20}{purple}. 

CK4Gen achieves the lowest meta-calibration loss of 11.78, showcasing exceptional robustness and accuracy across all stratifications. This performance stems from its design, which is optimised for survival data synthesis, delivering consistently low calibration losses across both individual subgroups and their aggregate measures. MCM ranks second with a meta-calibration loss of 13.54. While slightly higher than CK4Gen, MCM’s scalability and support for conditional data synthesis make it a more versatile option, particularly in real-world scenarios where retraining is impractical. 

\newpage
%###===###%###===###%###===###
\begin{table}[h]
    \tiny
    \centering
    \caption{Meta-Calibration Loss ($\mathbf{\sum}$) for Stratified Subgroups with Various Augmentation Methods.}
    \label{tab:meta_calibration_loss}
    \begin{tabular}{|p{1.8cm}||p{0.5cm}|p{0.9cm}|p{0.5cm}|p{0.5cm}|p{0.5cm}|p{0.5cm}|p{0.5cm}|p{0.5cm}|p{0.5cm}|p{0.5cm}|p{0.5cm}|p{0.5cm}|}
        \hline
        \multirow{2}{*}{} & 
        \multicolumn{2}{c|}{\textbf{Renal Function}} & 
        \multicolumn{2}{c|}{\textbf{Diabetes}} & 
        \multicolumn{2}{c|}{\textbf{Hypertension}} & 
        \multicolumn{2}{c|}{\textbf{Younger Age}} & 
        \multicolumn{2}{c|}{\textbf{CVD}} & 
        \multirow{2}{*}{$\mathbf{\sum}$} & 
        \multirow{2}{*}{\textbf{Rank}} \\ \cline{2-11}
        & Normal & Non-Ideal 
        & False & True 
        & False & True 
        & False & True 
        & False & True & & \\ 
        \hline \hline
        
        \cellcolor{green!20}\textbf{No Augmentation} & 
        \cellcolor{green!20}1.29 & \cellcolor{green!20}2.19 & \cellcolor{green!20}0.64 & \cellcolor{green!20}1.90 & 
        \cellcolor{green!20}1.46 & \cellcolor{green!20}1.27 & \cellcolor{green!20}0.62 & \cellcolor{green!20}2.50 & 
        \cellcolor{green!20}0.42 & \cellcolor{green!20}2.65 & \cellcolor{green!20}14.93 & 
        \cellcolor{green!20}3 \\
        \hline \hline
        
        \cellcolor{blue!20}\textbf{MCM (Ours)} & 
        \cellcolor{blue!20}0.59 & \cellcolor{blue!20}2.10 & \cellcolor{blue!20}0.69 & \cellcolor{blue!20}1.77 & 
        \cellcolor{blue!20}1.48 & \cellcolor{blue!20}1.03 & \cellcolor{blue!20}0.41 & \cellcolor{blue!20}2.40 & 
        \cellcolor{blue!20}0.51 & \cellcolor{blue!20}2.57 & \cellcolor{blue!20}13.54 & 
        \cellcolor{blue!20}2 \\
        \hline \hline
        
        \textbf{RandomOverSampler} & 1.83 & 2.13 & 1.96 & 1.83 & 2.19 & 1.14 & 0.69 & 2.50 & 0.83 & 2.63 & 17.73 & 8 \\ \hline
        
        \textbf{SMOTE} & 1.83 & 2.13 & 1.93 & 1.81 & 2.57 & 1.15 & 0.77 & 2.45 & 0.77 & 2.61 & 18.00 & 10 \\ \hline
        
        \textbf{SMOTENC} & 1.94 & 2.16 & 2.04 & 1.89 & 2.42 & 1.23 & 0.92 & 2.48 & 0.89 & 2.63 & 18.61 & 11 \\ \hline
        
        \textbf{ADASYN} & 1.84 & 2.09 & 1.96 & 1.77 & 2.60 & 1.03 & 0.73 & 2.43 & 0.74 & 2.61 & 17.80 & 9 \\ \hline
        
        \textbf{BorderlineSMOTE} & 1.77 & 2.08 & 1.32 & 1.78 & 1.54 & 1.06 & 0.84 & 2.45 & 0.82 & 2.61 & 16.27 & 6 \\ \hline 
        
        \textbf{SVMSMOTE} & 1.58 & 2.12 & 1.76 & 1.82 & 2.50 & 1.17 & 0.81 & 2.46 & 0.83 & 2.62 & 17.66 & 7 \\ \hline
        
        \textbf{VAE} & 1.59 & 2.07 & 1.05 & 1.77 & 1.86 & 1.02 & 0.41 & 2.50 & 0.45 & 2.63 & 15.34 & 4 \\ \hline
        
        \textbf{WGAN} & 1.01 & 2.21 & 1.01 & 1.91 & 1.89 & 1.39 & 0.66 & 2.48 & 0.38 & 2.65 & 15.59 & 5 \\ \hline
        
        \cellcolor{purple!20}\textbf{CK4Gen} & 
        \cellcolor{purple!20}0.59 & \cellcolor{purple!20}0.59 & \cellcolor{purple!20}0.59 & \cellcolor{purple!20}1.75 & 
        \cellcolor{purple!20}1.46 & \cellcolor{purple!20}1.02 & \cellcolor{purple!20}0.40 & \cellcolor{purple!20}2.39 & 
        \cellcolor{purple!20}0.42 & \cellcolor{purple!20}2.57 & \cellcolor{purple!20}11.78 & 
        \cellcolor{purple!20}1 \\ 
        
        \hline
    \end{tabular}
\end{table}

The non-augmented baseline also performs well, with a meta-calibration loss of 14.93, surpassing traditional augmentation methods like SMOTE and generative approaches such as VAE.

Although methods like WGAN excel in specific subgroups, such as achieving an overall calibration loss of 1.01 for the normal renal function group, their inconsistency across all stratifications is reflected in a higher meta-calibration loss of 15.59. This highlights the critical importance of robustness, as demonstrated by the stable performance of CK4Gen and MCM across all subgroups, underscoring their suitability for diverse and complex healthcare scenarios.

A detailed breakdown of the column entries in Table \ref{tab:meta_calibration_loss} is provided in Tables \ref{tab:calibration_loss_normal} - \ref{tab:calibration_loss_cvd_0}, corresponding to the \textit{Sum of Calibration Loss} columns. These tables report calibration losses across three time points -- 25th, 50th, and 75th percentiles of the time-to-event duration -- for stratified groups such as normal renal function (eGFR $\geq 90$ mL/min/1.73m²) and non-ideal renal function (eGFR $< 90$ mL/min/1.73m²). Calibration losses are presented as mean (standard deviation) for all augmentation methods, except for the baseline, which reflects a single CoxPH model without variability from cross-validation. All augmentation methods were evaluated using the setup detailed in Algorithm \ref{alg:StratifiedCalibration}, utilising 5x2 cross-validation with five iterations of simulated data per fold to ensure robust assessment across training-test splits. Despite the inherent reduction in training data for augmented methods due to cross-validation, MCM consistently outperformed the baseline, delivering superior calibration for downstream CoxPH models. Additionally, Appendix \ref{App:ExtraResults3} presents secondary results for seven methods that combine data simulation with MICE. While some combinations showed promise for specific stratified calibrations, they were found to be highly unstable and less reliable overall.

%###===>>>%###===>>>%###===>>>
%###===>>>%###===>>>%###===>>>
%###===>>>%###===>>>%###===>>>
%\newpage
\begin{table}[ht]
\tiny
\centering
\caption{Calibration Target: eGFR}
\label{tab:calibration_loss_normal}
\begin{tabular}{|p{2.5cm}||p{1.75cm}|p{1.75cm}|p{1.75cm}|p{1.75cm}|p{1cm}|}

\multicolumn{6}{c}{\footnotesize{Stratification On: Normal Renal Function (eGFR $\geq$ 90 mL/min/1.73m²)}}\\
 
\hline
\textbf{Augmentation Method} & 
\textbf{Duration Target:\newline 25th Percentile} &
\textbf{Duration Target:\newline 50th Percentile} &
\textbf{Duration Target:\newline 75th Percentile} &
\textbf{Sum of\newline Calibration Loss} & 
\textbf{Rank} \\ 
\hline\hline

\cellcolor{green!20}\textbf{No Augmentation}
& \cellcolor{green!20}0.3183
& \cellcolor{green!20}0.4603                
& \cellcolor{green!20}0.5086                      
& \cellcolor{green!20}1.2872              
& \cellcolor{green!20}4 \\ 
\hline\hline

\cellcolor{blue!20}\textbf{MCM} (Ours)
& \cellcolor{blue!20}0.0645 (0.0103)
& \cellcolor{blue!20}0.2172 (0.0073)
& \cellcolor{blue!20}0.3114 (0.0069)
& \cellcolor{blue!20}0.5931
& \cellcolor{blue!20}2 \\ 
\hline\hline

\textbf{RandomOverSampler}   
& 0.5318 (0.0582)
& 0.6315 (0.0403)
& 0.6668 (0.0337)
& 1.8301
& 9 \\ 
\hline

\textbf{SMOTE}
& 0.5196 (0.0629)
& 0.6327 (0.0457)
& 0.6744 (0.0424)
& 1.8267
& 8 \\ 
\hline

\textbf{SMOTENC}             
& 0.5731 (0.0630)
& 0.6669 (0.0414)
& 0.6979 (0.0355)
& 1.9379
& 11 \\ 
\hline

\textbf{ADASYN}
& 0.5246 (0.0817)
& 0.6351 (0.0610)
& 0.6756 (0.0536)
& 1.8353
& 10 \\ 
\hline

\textbf{BorderlineSMOTE}     
& 0.4965 (0.0685)
& 0.6117 (0.0515)
& 0.6596 (0.0465)
& 1.7678
& 7 \\ 
\hline

\textbf{SVMSMOTE}
& 0.4153 (0.0880)
& 0.5549 (0.0644)
& 0.6081 (0.0591)
& 1.5783
& 5 \\ 
\hline

\textbf{VAE}
& 0.4188 (0.0890)
& 0.5602 (0.0649)
& 0.6081 (0.0562)
& 1.5871
& 6 \\ 
\hline

\textbf{WGAN}
& 0.1592 (0.1138)
& 0.3878 (0.0850)
& 0.4599 (0.0770)
& 1.0069
& 3 \\ 
\hline

\cellcolor{purple!20}\textbf{CK4Gen}
& \cellcolor{purple!20}0.0970 (0.0000)
& \cellcolor{purple!20}0.1948 (0.0000)
& \cellcolor{purple!20}0.2932 (0.0000)
& \cellcolor{purple!20}0.5850
& \cellcolor{purple!20}1 \\ 
\hline
%###===>>>%###===>>>%###===>>>
%###===>>>%###===>>>%###===>>>
%###===>>>%###===>>>%###===>>>
\multicolumn{6}{c}{}\\
\multicolumn{6}{c}{\footnotesize{Stratification On: Non-Ideal Renal Function (eGFR $<$ 90 mL/min/1.73m²)}}\\
\hline
\textbf{Augmentation Method} & 
\textbf{Duration Target:\newline 25th Percentile} &
\textbf{Duration Target:\newline 50th Percentile} &
\textbf{Duration Target:\newline 75th Percentile} &
\textbf{Sum of\newline Calibration Loss} & 
\textbf{Rank} \\ 
\hline\hline

\cellcolor{green!20}\textbf{No Augmentation}
& \cellcolor{green!20}0.6959
& \cellcolor{green!20}0.7375                
& \cellcolor{green!20}0.752                      
& \cellcolor{green!20}2.1854              
& \cellcolor{green!20}10 \\ 
\hline\hline

\cellcolor{blue!20}\textbf{MCM} (Ours)
& \cellcolor{blue!20}0.6561 (0.0029)
& \cellcolor{blue!20}0.7111 (0.0018)
& \cellcolor{blue!20}0.7308 (0.0015)
& \cellcolor{blue!20}2.0980
& \cellcolor{blue!20}5 \\ 
\hline\hline

\textbf{RandomOverSampler}   
& 0.6708 (0.0108)
& 0.7211 (0.0069)
& 0.7391 (0.0057)
& 2.1310
& 8 \\ 
\hline

\textbf{SMOTE}
& 0.6679 (0.0138)
& 0.7198 (0.0115)
& 0.7393 (0.0097)
& 2.1270
& 7 \\ 
\hline

\textbf{SMOTENC}             
& 0.6849 (0.0112)
& 0.7296 (0.0080)
& 0.7450 (0.0071)
& 2.1595
& 9 \\ 
\hline

\textbf{ADASYN}
& 0.6521 (0.0098)
& 0.7103 (0.0060)
& 0.7316 (0.0053)
& 2.0940
& 4 \\ 
\hline

\textbf{BorderlineSMOTE}     
& 0.6457 (0.0150)
& 0.7052 (0.0092)
& 0.7273 (0.0069)
& 2.0782
& 2 \\ 
\hline

\textbf{SVMSMOTE}
& 0.6642 (0.0149)
& 0.7173 (0.0101)
& 0.7369 (0.0083)
& 2.1184
& 6 \\ 
\hline

\cellcolor{purple!20}\textbf{VAE}
& \cellcolor{purple!20}0.6515 (0.0054)
& \cellcolor{purple!20}0.7013 (0.0054)
& \cellcolor{purple!20}0.7189 (0.0054)
& \cellcolor{purple!20}2.0717
& \cellcolor{purple!20}1 \\ 
\hline

\textbf{WGAN}
& 0.7133 (0.0018)
& 0.7448 (0.0013)
& 0.7554 (0.0012)
& 2.2135
& 11 \\ 
\hline

\textbf{CK4Gen}
& 0.6479 (0.0000)
& 0.7064 (0.0000)
& 0.7271 (0.0000)
& 2.0814
& 3 \\ 
\hline
\end{tabular}
\end{table}

%###===>>>%###===>>>%###===>>>
%###===>>>%###===>>>%###===>>>
%###===>>>%###===>>>%###===>>>
\newpage

\begin{table}[ht]
\tiny
\centering
\caption{Calibration Target: Diabetes Status}
\label{tab:calibration_loss_diabetes}
\begin{tabular}{|p{2.5cm}||p{1.75cm}|p{1.75cm}|p{1.75cm}|p{1.75cm}|p{1cm}|}

\multicolumn{6}{c}{\footnotesize{Stratification On: No Diabetes}}\\
\hline
\textbf{Augmentation Method} & 
\textbf{Duration Target:\newline 25th Percentile} &
\textbf{Duration Target:\newline 50th Percentile} &
\textbf{Duration Target:\newline 75th Percentile} &
\textbf{Sum of\newline Calibration Loss} & 
\textbf{Rank} \\ 
\hline\hline

\cellcolor{green!20}\textbf{No Augmentation}
& \cellcolor{green!20}0.0268
& \cellcolor{green!20}0.2583                
& \cellcolor{green!20}0.3536                      
& \cellcolor{green!20}0.6387              
& \cellcolor{green!20}2 \\ 
\hline\hline

\cellcolor{blue!20}\textbf{MCM} (Ours)
& \cellcolor{blue!20}0.0907 (0.0361)
& \cellcolor{blue!20}0.2524 (0.0693)
& \cellcolor{blue!20}0.3466 (0.0608)
& \cellcolor{blue!20}0.6897
& \cellcolor{blue!20}3 \\ 
\hline\hline

\textbf{RandomOverSampler}   
& 0.5942 (0.0538)
& 0.6689 (0.0406)
& 0.6973 (0.0332)
& 1.9604
& 10 \\ 
\hline

\textbf{SMOTE}
& 0.5750 (0.0499)
& 0.6605 (0.0408)
& 0.6951 (0.0401)
& 1.9306
& 8 \\ 
\hline

\textbf{SMOTENC}             
& 0.6251 (0.0515)
& 0.6955 (0.0461)
& 0.7213 (0.0461)
& 2.0419
& 11 \\ 
\hline

\textbf{ADASYN}
& 0.5879 (0.0335)
& 0.6682 (0.0325)
& 0.7007 (0.0348)
& 1.9568
& 9 \\ 
\hline

\textbf{BorderlineSMOTE}     
& 0.3901 (0.1862)
& 0.4360 (0.2311)
& 0.4981 (0.1931)
& 1.3242
& 6 \\ 
\hline

\textbf{SVMSMOTE}
& 0.5040 (0.0544)
& 0.6110 (0.0410)
& 0.6499 (0.0392)
& 1.7649
& 7 \\ 
\hline

\textbf{VAE}
& 0.2325 (0.1129)
& 0.3714 (0.1497)
& 0.4458 (0.1284)
& 1.0497
& 5 \\ 
\hline

\textbf{WGAN}
& 0.1604 (0.0556)
& 0.3892 (0.0409)
& 0.4598 (0.0370)
& 1.0094
& 4 \\ 
\hline

\cellcolor{purple!20}\textbf{CK4Gen}
& \cellcolor{purple!20}0.1159 (0.0000)
& \cellcolor{purple!20}0.1879 (0.0000)
& \cellcolor{purple!20}0.2908 (0.0000)
& \cellcolor{purple!20}0.5946
& \cellcolor{purple!20}1 \\ 
\hline
%###===>>>%###===>>>%###===>>>
%###===>>>%###===>>>%###===>>>
%###===>>>%###===>>>%###===>>>
\multicolumn{6}{c}{}\\
\multicolumn{6}{c}{\footnotesize{Stratification On: Diabetes}}\\
\hline
\textbf{Augmentation Method} & 
\textbf{Duration Target:\newline 25th Percentile} &
\textbf{Duration Target:\newline 50th Percentile} &
\textbf{Duration Target:\newline 75th Percentile} &
\textbf{Sum of\newline Calibration Loss} & 
\textbf{Rank} \\ 
\hline\hline

\cellcolor{green!20}\textbf{No Augmentation}
& \cellcolor{green!20}0.5898
& \cellcolor{green!20}0.6458                
& \cellcolor{green!20}0.6655                      
& \cellcolor{green!20}1.9011              
& \cellcolor{green!20}10 \\ 
\hline\hline

\cellcolor{blue!20}\textbf{MCM} (Ours)
& \cellcolor{blue!20}0.5276 (0.0024)
& \cellcolor{blue!20}0.6069 (0.0014)
& \cellcolor{blue!20}0.6354 (0.0011)
& \cellcolor{blue!20}1.7699
& \cellcolor{blue!20}3 \\ 
\hline\hline

\textbf{RandomOverSampler}   
& 0.5539 (0.0110)
& 0.6257 (0.0042)
& 0.6515 (0.0038)
& 1.8311
& 8 \\ 
\hline

\textbf{SMOTE}
& 0.5416 (0.0162)
& 0.6177 (0.0144)
& 0.6464 (0.0133)
& 1.8057
& 6 \\ 
\hline

\textbf{SMOTENC}             
& 0.5816 (0.0032)
& 0.6442 (0.0090)
& 0.6655 (0.0107)
& 1.8913
& 9 \\ 
\hline

\textbf{ADASYN}
& 0.5208 (0.0121)
& 0.6070 (0.0069)
& 0.6383 (0.0065)
& 1.7661
& 2 \\ 
\hline

\textbf{BorderlineSMOTE}     
& 0.5251 (0.0139)
& 0.6115 (0.0102)
& 0.6440 (0.0090)
& 1.7806
& 5 \\ 
\hline

\textbf{SVMSMOTE}
& 0.5475 (0.0205)
& 0.6204 (0.0173)
& 0.6482 (0.0150)
& 1.8161
& 7 \\ 
\hline

\textbf{VAE}
& 0.5333 (0.0161)
& 0.6061 (0.0124)
& 0.6314 (0.0110)
& 1.7708
& 4 \\ 
\hline

\textbf{WGAN}
& 0.6041 (0.0030)
& 0.6479 (0.0015)
& 0.6629 (0.0013)
& 1.9149
& 11 \\ 
\hline

\cellcolor{purple!20}\textbf{CK4Gen}
& \cellcolor{purple!20}0.5204 (0.0000)
& \cellcolor{purple!20}0.6021 (0.0000)
& \cellcolor{purple!20}0.6316 (0.0000)
& \cellcolor{purple!20}1.7541
& \cellcolor{purple!20}1 \\ 
\hline
\end{tabular}
\end{table}

%###===>>>%###===>>>%###===>>>
%###===>>>%###===>>>%###===>>>
%###===>>>%###===>>>%###===>>>
\begin{table}[ht]
\tiny
\centering
\caption{Calibration Target: Hypertension Status}
\label{tab:calibration_loss_hypertension_0}
\begin{tabular}{|p{2.5cm}||p{1.75cm}|p{1.75cm}|p{1.75cm}|p{1.75cm}|p{1cm}|}

\multicolumn{6}{c}{\footnotesize{Stratification On: No Hypertension}}\\
\hline
\textbf{Augmentation Method} & 
\textbf{Duration Target:\newline 25th Percentile} &
\textbf{Duration Target:\newline 50th Percentile} &
\textbf{Duration Target:\newline 75th Percentile} &
\textbf{Sum of\newline Calibration Loss} & 
\textbf{Rank} \\ 
\hline\hline

\cellcolor{green!20}\textbf{No Augmentation}
& \cellcolor{green!20}0.3590
& \cellcolor{green!20}0.5224                
& \cellcolor{green!20}0.5778                      
& \cellcolor{green!20}1.4592              
& \cellcolor{green!20}2 \\ 
\hline\hline

\cellcolor{blue!20}\textbf{MCM} (Ours)
& \cellcolor{blue!20}0.3681 (0.0114)
& \cellcolor{blue!20}0.5297 (0.0078)
& \cellcolor{blue!20}0.5831 (0.0068)
& \cellcolor{blue!20}1.4809
& \cellcolor{blue!20}3 \\ 
\hline\hline

\textbf{RandomOverSampler}   
& 0.6787 (0.1683)
& 0.7439 (0.1257)
& 0.7677 (0.1106)
& 2.1903
& 7 \\ 
\hline

\textbf{SMOTE}
& 0.8351 (0.0469)
& 0.8601 (0.0375)
& 0.8720 (0.0313)
& 2.5672
& 10 \\ 
\hline

\textbf{SMOTENC}             
& 0.7748 (0.0362)
& 0.8165 (0.0323)
& 0.8316 (0.0295)
& 2.4229
& 8 \\ 
\hline

\textbf{ADASYN}
& 0.8478 (0.0315)
& 0.8703 (0.0240)
& 0.8807 (0.0201)
& 2.5988
& 11 \\ 
\hline

\textbf{BorderlineSMOTE}     
& 0.3945 (0.2388)
& 0.5487 (0.1679)
& 0.6016 (0.1443)
& 1.5448
& 4 \\ 
\hline

\textbf{SVMSMOTE}
& 0.8058 (0.0214)
& 0.8385 (0.0183)
& 0.8526 (0.0156)
& 2.4969
& 9 \\ 
\hline

\textbf{VAE}
& 0.5275 (0.0760)
& 0.6456 (0.0565)
& 0.6850 (0.0496)
& 1.8581
& 5 \\ 
\hline

\textbf{WGAN}
& 0.5387 (0.0176)
& 0.6542 (0.0110)
& 0.6928 (0.0101)
& 1.8857
& 6 \\ 
\hline

\cellcolor{purple!20}\textbf{CK4Gen}
& \cellcolor{purple!20}0.3577 (0.0000)
& \cellcolor{purple!20}0.5224 (0.0000)
& \cellcolor{purple!20}0.5768 (0.0000)
& \cellcolor{purple!20}1.4569
& \cellcolor{purple!20}1 \\ 
\hline
%###===>>>%###===>>>%###===>>>
%###===>>>%###===>>>%###===>>>
%###===>>>%###===>>>%###===>>>
\multicolumn{6}{c}{}\\
\multicolumn{6}{c}{\footnotesize{Stratification On: Hypertension}}\\
\hline
\textbf{Augmentation Method} & 
\textbf{Duration Target:\newline 25th Percentile} &
\textbf{Duration Target:\newline 50th Percentile} &
\textbf{Duration Target:\newline 75th Percentile} &
\textbf{Sum of\newline Calibration Loss} & 
\textbf{Rank} \\ 
\hline\hline

\cellcolor{green!20}\textbf{No Augmentation}
& \cellcolor{green!20}0.3503
& \cellcolor{green!20}0.4434                
& \cellcolor{green!20}0.4759                      
& \cellcolor{green!20}1.2696              
& \cellcolor{green!20}10 \\ 
\hline\hline

\cellcolor{blue!20}\textbf{MCM} (Ours)
& \cellcolor{blue!20}0.2379 (0.0054)
& \cellcolor{blue!20}0.3728 (0.0033)
& \cellcolor{blue!20}0.4212 (0.0026)
& \cellcolor{blue!20}1.0319
& \cellcolor{blue!20}4 \\ 
\hline\hline

\textbf{RandomOverSampler}   
& 0.2837 (0.0273)
& 0.4056 (0.0190)
& 0.4484 (0.0182)
& 1.1377
& 6 \\ 
\hline

\textbf{SMOTE}
& 0.2830 (0.0296)
& 0.4101 (0.0246)
& 0.4558 (0.0220)
& 1.1489
& 7 \\ 
\hline

\textbf{SMOTENC}             
& 0.3250 (0.0195)
& 0.4362 (0.0161)
& 0.4724 (0.0168)
& 1.2336
& 9 \\ 
\hline

\textbf{ADASYN}
& 0.2246 (0.0323)
& 0.3750 (0.0249)
& 0.4283 (0.0231)
& 1.0279
& 3 \\ 
\hline

\textbf{BorderlineSMOTE}     
& 0.2384 (0.0381)
& 0.3854 (0.0353)
& 0.4401 (0.0324)
& 1.0639
& 5 \\ 
\hline

\textbf{SVMSMOTE}
& 0.3014 (0.0529)
& 0.4117 (0.0419)
& 0.4537 (0.0366)
& 1.1668
& 8 \\ 
\hline

\cellcolor{purple!20}\textbf{VAE}
& \cellcolor{purple!20}0.2369 (0.0320)
& \cellcolor{purple!20}0.3692 (0.0238)
& \cellcolor{purple!20}0.4147 (0.0213)
& \cellcolor{purple!20}1.0208
& \cellcolor{purple!20}1 \\ 
\hline

\textbf{WGAN}
& 0.4138 (0.0066)
& 0.4751 (0.0065)
& 0.4961 (0.0071)
& 1.3850 
& 11 \\ 
\hline

\textbf{CK4Gen}
& 0.2344 (0.0000)
& 0.3695 (0.0000)
& 0.4181 (0.0000)
& 1.0220
& 2 \\ 
\hline
\end{tabular}
\end{table}

%###===>>>%###===>>>%###===>>>
%###===>>>%###===>>>%###===>>>
%###===>>>%###===>>>%###===>>>
\newpage
\begin{table}[h]
\tiny
\centering
\caption{Calibration Target: Age}
\label{tab:calibration_loss_age_0}
\begin{tabular}{|p{2.5cm}||p{1.75cm}|p{1.75cm}|p{1.75cm}|p{1.75cm}|p{1cm}|}

\multicolumn{6}{c}{\footnotesize{Stratification On: Younger (Age $<$ 65 years)}}\\
\hline
\textbf{Augmentation Method} & 
\textbf{Duration Target:\newline 25th Percentile} &
\textbf{Duration Target:\newline 50th Percentile} &
\textbf{Duration Target:\newline 75th Percentile} &
\textbf{Sum of\newline Calibration Loss} & 
\textbf{Rank} \\ 
\hline\hline

\cellcolor{green!20}\textbf{No Augmentation}
& \cellcolor{green!20}0.0885
& \cellcolor{green!20}0.2405                
& \cellcolor{green!20}0.2932                      
& \cellcolor{green!20}0.6222              
& \cellcolor{green!20}4 \\ 
\hline\hline

\cellcolor{blue!20}\textbf{MCM} (Ours)
& \cellcolor{blue!20}0.0123 (0.0062)
& \cellcolor{blue!20}0.1664 (0.0090)
& \cellcolor{blue!20}0.2298 (0.0090)
& \cellcolor{blue!20}0.4085
& \cellcolor{blue!20}2 \\ 
\hline\hline

\textbf{RandomOverSampler}   
& 0.1363 (0.0494)
& 0.2449 (0.0843)
& 0.3044 (0.0758)
& 0.6856
& 6 \\ 
\hline

\textbf{SMOTE}
& 0.1555 (0.0274)
& 0.2748 (0.0942)
& 0.3410 (0.0823)
& 0.7713
& 8 \\ 
\hline

\textbf{SMOTENC}             
& 0.2044 (0.0534)
& 0.3354 (0.0365)
& 0.3792 (0.0342)
& 0.919
& 11 \\ 
\hline

\textbf{ADASYN}
& 0.1243 (0.0751)
& 0.2710 (0.0884)
& 0.3392 (0.0782)
& 0.7345
& 7 \\ 
\hline

\textbf{BorderlineSMOTE}     
& 0.1475 (0.0896)
& 0.3134 (0.0688)
& 0.3747 (0.0620)
& 0.8356
& 10 \\ 
\hline

\textbf{SVMSMOTE}
& 0.1745 (0.0771)
& 0.2886 (0.0996)
& 0.3475 (0.0867)
& 0.8106
& 9 \\ 
\hline

\textbf{VAE}
& 0.2446 (0.0942)
& 0.0456 (0.0539)
& 0.1231 (0.0556)
& 0.4133
& 3 \\ 
\hline

\textbf{WGAN}
& 0.1206 (0.0245)
& 0.2495 (0.0169)
& 0.2933 (0.0145)
& 0.6634
& 5 \\ 
\hline

\cellcolor{purple!20}\textbf{CK4Gen}
& \cellcolor{purple!20}0.0138 (0.0000)
& \cellcolor{purple!20}0.1632 (0.0000)
& \cellcolor{purple!20}0.2269 (0.0000)
& \cellcolor{purple!20}0.4039
& \cellcolor{purple!20}1 \\ 
\hline
%###===>>>%###===>>>%###===>>>
%###===>>>%###===>>>%###===>>>
%###===>>>%###===>>>%###===>>>
\multicolumn{6}{c}{}\\
\multicolumn{6}{c}{\footnotesize{Stratification On: Older (Age $\ge$ 65 years)}}\\
\hline
\textbf{Augmentation Method} & 
\textbf{Duration Target:\newline 25th Percentile} &
\textbf{Duration Target:\newline 50th Percentile} &
\textbf{Duration Target:\newline 75th Percentile} &
\textbf{Sum of\newline Calibration Loss} & 
\textbf{Rank} \\ 
\hline\hline

\cellcolor{green!20}\textbf{No Augmentation}
& \cellcolor{green!20}0.8115
& \cellcolor{green!20}0.8387                
& \cellcolor{green!20}0.8483                      
& \cellcolor{green!20}2.4985              
& \cellcolor{green!20}10 \\ 
\hline\hline

\cellcolor{blue!20}\textbf{MCM} (Ours)
& \cellcolor{blue!20}0.7659 (0.0029)
& \cellcolor{blue!20}0.8076 (0.0020)
& \cellcolor{blue!20}0.8221 (0.0018)
& \cellcolor{blue!20}2.3956
& \cellcolor{blue!20}2 \\ 
\hline\hline

\textbf{RandomOverSampler}   
& 0.8138 (0.0100)
& 0.8381 (0.0045)
& 0.8468 (0.0033)
& 2.4987
& 11 \\ 
\hline

\textbf{SMOTE}
& 0.7897 (0.0136)
& 0.8222 (0.0090)
& 0.8339 (0.0075)
& 2.4458
& 4 \\ 
\hline

\textbf{SMOTENC}             
& 0.8039 (0.0092)
& 0.8321 (0.0042)
& 0.8417 (0.0029)
& 2.4777
& 7 \\ 
\hline

\textbf{ADASYN}
& 0.7838 (0.0101)
& 0.8185 (0.0073)
& 0.8311 (0.0067)
& 2.4334
& 3 \\ 
\hline

\textbf{BorderlineSMOTE}     
& 0.7907 (0.0179)
& 0.8230 (0.0166)
& 0.8346 (0.0093)
& 2.4483
& 5 \\ 
\hline

\textbf{SVMSMOTE}
& 0.7948 (0.0121)
& 0.8255 (0.0076)
& 0.8366 (0.0062)
& 2.4569
& 6 \\ 
\hline

\textbf{VAE}
& 0.8101 (0.0137)
& 0.8383 (0.0118)
& 0.8482 (0.0111)
& 2.4966
& 9 \\ 
\hline

\textbf{WGAN}
& 0.7999 (0.0105)
& 0.8340 (0.0049)
& 0.8453 (0.0032)
& 2.4792
& 8 \\ 
\hline

\cellcolor{purple!20}\textbf{CK4Gen}
& \cellcolor{purple!20}0.7642 (0.0000)
& \cellcolor{purple!20}0.8066 (0.0000)
& \cellcolor{purple!20}0.8213 (0.0000)
& \cellcolor{purple!20}2.3921
& \cellcolor{purple!20}1 \\ 
\hline
\end{tabular}
\end{table}

%###===>>>%###===>>>%###===>>>
%###===>>>%###===>>>%###===>>>
%###===>>>%###===>>>%###===>>>
%\newpage
\begin{table}[h]
\tiny
\centering
\caption{Calibration Target: Cardiovascular Disease Status}
\label{tab:calibration_loss_cvd_0}
\begin{tabular}{|p{2.5cm}||p{1.75cm}|p{1.75cm}|p{1.75cm}|p{1.75cm}|p{1cm}|}

\multicolumn{6}{c}{\footnotesize{Stratification On: No CVD}}\\
\hline
\textbf{Augmentation Method} & 
\textbf{Duration Target:\newline 25th Percentile} &
\textbf{Duration Target:\newline 50th Percentile} &
\textbf{Duration Target:\newline 75th Percentile} &
\textbf{Sum of\newline Calibration Loss} & 
\textbf{Rank} \\ 
\hline\hline

\cellcolor{green!20}\textbf{No Augmentation}
& \cellcolor{green!20}0.0966
& \cellcolor{green!20}0.1259                
& \cellcolor{green!20}0.2021                      
& \cellcolor{green!20}0.4246              
& \cellcolor{green!20}3 \\ 
\hline\hline

\cellcolor{blue!20}\textbf{MCM} (Ours)
& \cellcolor{blue!20}0.0607 (0.0160)
& \cellcolor{blue!20}0.1908 (0.0496)
& \cellcolor{blue!20}0.2576 (0.0470)
& \cellcolor{blue!20}0.5091
& \cellcolor{blue!20}5 \\ 
\hline\hline

\textbf{RandomOverSampler}   
& 0.1531 (0.0848)
& 0.3105 (0.0499)
& 0.3658 (0.0376)
& 0.8294
& 10 \\ 
\hline

\textbf{SMOTE}
& 0.1402 (0.0479)
& 0.2813 (0.0620)
& 0.3438 (0.0466)
& 0.7653
& 7 \\ 
\hline

\textbf{SMOTENC}             
& 0.1853 (0.0746)
& 0.3307 (0.0497)
& 0.3778 (0.0422)
& 0.8938
& 11 \\ 
\hline

\textbf{ADASYN}
& 0.1116 (0.0633)
& 0.2848 (0.0412)
& 0.3474 (0.0323)
& 0.7438
& 6 \\ 
\hline

\textbf{BorderlineSMOTE}     
& 0.1591 (0.0644)
& 0.3027 (0.0601)
& 0.3612 (0.0469)
& 0.823
& 8 \\ 
\hline

\textbf{SVMSMOTE}
& 0.1767 (0.0654)
& 0.2973 (0.0787)
& 0.3547 (0.0604)
& 0.8287
& 9 \\ 
\hline

\textbf{VAE}
& 0.1536 (0.0635)
& 0.1021 (0.0433)
& 0.1900 (0.0378)
& 0.4457
& 4 \\ 
\hline

\cellcolor{purple!20}\textbf{WGAN}
& \cellcolor{purple!20}0.2032 (0.1026)
& \cellcolor{purple!20}0.0537 (0.0638)
& \cellcolor{purple!20}0.1253 (0.0594)
& \cellcolor{purple!20}0.3822
& \cellcolor{purple!20}1 \\ 
\hline

\textbf{CK4Gen}
& 0.0769 (0.0000)
& 0.1354 (0.0000)
& 0.2069 (0.0000)
& 0.4192
& 2 \\ 
\hline
%###===>>>%###===>>>%###===>>>
%###===>>>%###===>>>%###===>>>
%###===>>>%###===>>>%###===>>>
\multicolumn{6}{c}{}\\
\multicolumn{6}{c}{\footnotesize{Stratification On: CVD}}\\
\hline
\textbf{Augmentation Method} & 
\textbf{Duration Target:\newline 25th Percentile} &
\textbf{Duration Target:\newline 50th Percentile} &
\textbf{Duration Target:\newline 75th Percentile} &
\textbf{Sum of\newline Calibration Loss} & 
\textbf{Rank} \\ 
\hline\hline

\cellcolor{green!20}\textbf{No Augmentation}
& \cellcolor{green!20}0.8686
& \cellcolor{green!20}0.886                
& \cellcolor{green!20}0.8921                      
& \cellcolor{green!20}2.6467              
& \cellcolor{green!20}10 \\ 
\hline\hline

\cellcolor{purple!20}\textbf{MCM} (Ours)
& \cellcolor{purple!20}0.8328 (0.0015)
& \cellcolor{purple!20}0.8619 (0.0010)
& \cellcolor{purple!20}0.8728 (0.0009)
& \cellcolor{purple!20}2.5675
& \cellcolor{purple!20}1 \\ 
\hline\hline

\textbf{RandomOverSampler}   
& 0.8619 (0.0098)
& 0.8817 (0.0061)
& 0.8891 (0.0048)
& 2.6327
& 8 \\ 
\hline

\textbf{SMOTE}
& 0.8546 (0.0140)
& 0.8760 (0.0090)
& 0.8838 (0.0072)
& 2.6144
& 5 \\ 
\hline

\textbf{SMOTENC}             
& 0.8623 (0.0111)
& 0.8819 (0.0069)
& 0.8892 (0.0053)
& 2.6334
& 9 \\ 
\hline

\textbf{ADASYN}
& 0.8526 (0.0125)
& 0.8750 (0.0080)
& 0.8834 (0.0065)
& 2.611
& 4 \\ 
\hline

\textbf{BorderlineSMOTE}     
& 0.8513 (0.0125)
& 0.8733 (0.0077)
& 0.8818 (0.0062)
& 2.6064
& 3 \\ 
\hline

\textbf{SVMSMOTE}
& 0.8570 (0.0098)
& 0.8777 (0.0059)
& 0.8857 (0.0046)
& 2.6204
& 6 \\ 
\hline

\textbf{VAE}
& 0.8599 (0.0042)
& 0.8793 (0.0021)
& 0.8860 (0.0018)
& 2.6252
& 7 \\ 
\hline

\textbf{WGAN}
& 0.8706 (0.0045)
& 0.8882 (0.0024)
& 0.8943 (0.0016)
& 2.6531
& 11 \\ 
\hline

\textbf{CK4Gen}
& 0.8332 (0.0000)
& 0.8622 (0.0000)
& 0.8730 (0.0000)
& 2.5684
& 2 \\ 
\hline
\end{tabular}
\end{table}

%###===>>>%###===>>>%###===>>>
%###===>>>%###===>>>%###===>>>
%###===>>>%###===>>>%###===>>>
\newpage
\section{Conclusion}
This study addresses key challenges in healthcare research, focusing on the need for synthetic data that balances realism, utility, and practicality while ensuring privacy and enabling robust downstream modelling. Existing methods, such as SMOTE and VAEs, often fall short in delivering the calibration, scalability, and stratification necessary for effective survival analysis. To overcome these limitations, we introduce \textbf{Masked Clinical Modelling (MCM)}, an attention-based framework inspired by masked language modelling. MCM uniquely combines standalone data synthesis for reproducibility with conditional simulation for targeted augmentation, enabling enhanced calibration and robust modelling without the need for retraining.

Through a comprehensive three-tiered evaluation framework -- realism checks (Section \ref{Sec:Check01}), utility verification (Section \ref{Sec:Check02}), and practicality analysis (Section \ref{Sec:Check03}) -- we demonstrate MCM’s ability to generate high-fidelity synthetic datasets that replicate the statistical and structural properties of real-world clinical data. Kaplan-Meier survival curves and Cox Proportional Hazards (CoxPH) models validate MCM’s utility, preserving critical covariate-event associations and temporal patterns. Notably, MCM reduced the overall calibration loss across the entire dataset by more than 15\%$^\text{\ref{FN:01}}$ and achieved a reduction of over 9\% in meta-calibration loss for stratified cohorts compared to models trained without augmentation\footnote{Refer to Table \ref{tab:meta_calibration_loss}: $1 - 13.54/14.93 \approx 0.09$.}. This performance surpasses the majority of baseline methods, highlighting MCM’s effectiveness in enhancing calibration across diverse scenarios. While CK4Gen achieves slightly lower meta-calibration loss, its lack of scalability and the need for retraining when presented with the stratified calibration condition limit its practical application. In contrast, MCM demonstrates greater stability and effectiveness, as evidenced by its superior calibration performance across 10 clinically stratified subgroups (Table \ref{tab:meta_calibration_loss}) and its flexibility for diverse scenarios (Table \ref{tab:methods_comparison}).

MCM’s dual capability of generating standalone datasets and performing conditional augmentation ensures adaptability and scalability in real-world healthcare applications. Unlike CK4Gen, MCM does not require retraining for on-the-spot enhancement of CoxPH model calibration (Section \ref{Sec:EnhancedCalibration}). By promoting equitable representation across diverse patient populations, MCM advances the precision and reliability of healthcare models, delivering translational benefits in restricted-access environments and supporting data-driven decision-making to optimise scarce healthcare resources.

%###===>>>%###===>>>%###===>>>
%###===>>>%###===>>>%###===>>>
%###===>>>%###===>>>%###===>>>
\section{Discussion}
\underline{Realism, Utility, and Privacy}\\
While MCM demonstrates superior capabilities in generating synthetic datasets, we prioritised improving practicality over immediate privacy considerations because poor-quality data compromises downstream utility, leading to flawed predictions and potential harm to patient outcomes. Synthetic datasets must first demonstrate robust utility in enhancing model calibration and predictive precision, as seen in this study’s use of a publicly accessible CKD EHR dataset. By focusing on practicality, we ensure that synthetic data delivers meaningful translational benefits in real-world healthcare applications, establishing a strong foundation for future research. With these advancements in place, subsequent efforts can integrate privacy-preserving mechanisms. Now that MCM has established exceptionally high realism in its synthetic datasets, addressing privacy concerns becomes increasingly critical, warranting the integration of privacy-preserving mechanisms in subsequent research.

\underline{Calibration Improvements and Data Imbalance}\\
See Table \ref{tab:meta_calibration_loss}, MCM demonstrated notable improvements in calibration for specific subgroups, such as reducing the overall calibration loss for normal renal function from 1.29 to 0.59. However, for other subgroups, like patients with CVD, the gains were marginal, indicating areas for further optimisation. The interplay between data imbalance and calibration performance warrants deeper investigation. Future studies could explore varying data mixture ratios in simulations and assess how imbalances influence the effectiveness of MCM, ensuring optimal performance across all stratifications.

\underline{Small Dataset Constraints and Scalability}\\
This study utilises the CKD EHR dataset with only 491 patients, a limitation that constrains its immediate generalisability to large-scale healthcare applications. While MCM has demonstrated strong performance on this dataset, its effectiveness on national risk equations developed using datasets containing millions of records -- such as QRISK4~\cite{hippisley2024development} in the UK, PREVENT~\cite{khan2024development} in the US, or PREDICT~\cite{barbieri2022predicting} in New Zealand -- remains to be validated. However, MCM’s capacity to excel in highly stratified scenarios, such as specific subgroups (\textit{e.g.,} females aged 20 to 40 with high progesterone but low oestrogen levels, a combination of characteristics likely presented in datasets such as GBSG2~\cite{schumacher1994randomized}), suggests its potential for application in niche areas where data availability is inherently limited. This scalability in specialised settings underscores MCM’s promise as a flexible tool for diverse healthcare challenges.

\underline{Redefining Metrics for Realism, Utility, and Practicality}\\
Future advancements in synthetic data methodologies should prioritise the redefinition of metrics for realism, utility, and practicality, particularly within the framework of causal inference. Incorporating concepts such as propensity scores~\cite{caliendo2008some} into synthetic data generation could enable the creation of closely matched patient pairs, aiding in the detection and analysis of underlying clinical confounders~\cite{greenland1999confounding}. By aligning synthetic datasets with principles used in emulated trial designs~\cite{hernan2016using}, researchers could simulate target trials that approximate randomised controlled trials using observational data. This refinement would significantly enhance the interpretability and causal validity of synthetic data, transforming it into a robust tool not only for healthcare research but also for improving evidence-based decision-making in clinical practice.

\underline{Synthetic Data in Restricted Access Healthcare Environments}\\
The growing adoption of synthetic data underscores its transformative potential in healthcare. As mentioned in Section \ref{Sec:RelatedWork}, governments, academic institutions, and private organisations are collaborating globally to harness synthetic data for secure and impactful research. Initiatives such as those by the U.S. Department of Homeland Security~\cite{dhs_synthetic_data_contracts} and Australia’s Digital Health CRC~\cite{synthetic_data_practice_synd} highlight its role in overcoming stringent privacy regulations and improving data accessibility. However, the idea that synthetic data is limited to walled-garden scenarios reflects a narrow, overly optimistic approach. Large institutions, despite operating in restricted environments, remain constant targets for malicious actors, and breaches can compromise vast amounts of sensitive information. By advancing highly realistic synthetic datasets, we not only add a proactive layer of protection but also foster collaborative innovation beyond enclosed systems. A future combining real and synthetic data offers a balanced solution, reducing risks while minimising barriers to research and education, thereby enabling meaningful advancements in healthcare optimisation.

%###===>>>%###===>>>%###===>>>
%###===>>>%###===>>>%###===>>>
%###===>>>%###===>>>%###===>>>
\newpage
\bibliographystyle{IEEEtran}
\bibliography{iclr2023_conference}

%###===>>>%###===>>>%###===>>>
%###===>>>%###===>>>%###===>>>
%###===>>>%###===>>>%###===>>>
\newpage
\appendix
\section{Appendix: Reproducibility}

\subsection*{Hardware and System Configuration}

To ensure reproducibility and transparency, we provide detailed information on the hardware and system configuration used in this study:

\begin{table}[h!]
\centering
\begin{tabular}{|p{5cm}|p{7cm}|}
\hline
\textbf{Category} & \textbf{Information} \\
\hline
\hline
\textbf{Hardware} & \\
Number of GPUs & 1 \\
GPU Name & Tesla T4 \\
CUDA Version & 12.2 \\
\hline
\hline
\textbf{System Details} & \\
Operating System & Linux-6.1.85 \\
Processor & x86\_64 \\
Python Version~\cite{vanRossum1995} & 3.10.12 \\
\hline
\hline
\textbf{Library Versions} & \\
NumPy Version~\cite{harris2020array} & 1.26.4 \\
Pandas Version~\cite{mckinney-proc-scipy-2010} & 2.2.2 \\
Matplotlib Version~\cite{Hunter2007} & 3.8.0 \\
Lifelines Version~\cite{Davidson-Pilon2019} & 0.30.0 \\
Scikit-Learn Version~\cite{sklearn_api} & 1.2.2 \\
Imbalanced-Learn Version~\cite{JMLR:v18:16-365} & 0.9.1 \\
PyTorch Version~\cite{paszke2019pytorch} & 2.5.1+cu121 \\
\hline
\end{tabular}
\caption{Hardware and system configuration details for reproducibility.}
\label{tab:hardware}
\end{table}

\subsection*{Ensuring Reproducibility}

To facilitate reproducibility, all functions involving stochastic operations, such as data splitting and model initialisation, were seeded consistently throughout the study. This ensures that results can be replicated under similar conditions. The library versions and system configurations listed in Table~\ref{tab:hardware} provide additional details for replicating the experiments on similar hardware.

All codes and scripts required to reproduce the results will be made publicly available upon acceptance of this paper$^{\ref{Ref:OpenSource}}$.

%###===>>>%###===>>>%###===>>>
%###===>>>%###===>>>%###===>>>
%###===>>>%###===>>>%###===>>>
\newpage
\section{Appendix: Calibration Computation Process\label{App:Cali}}

This appendix provides more descriptions on the implementations of the quantile-based calibration used in the NZ national cardiovascular disease PREDICT risk score equations. Specifically, we follow the instructions under the \texttt{calibration\_plot} function in line 27 in\\
\url{https://github.com/VIEW2020/Varianz2012/blob/master/code/VARIANZ/plot.py}.\\
More useful codes, on the predicted risk score can be found in\\
\url{https://github.com/CamDavidsonPilon/lifelines/blob/master/lifelines/fitters/cox_time_varying_fitter.py}\\
in the official Lifelines library show more details on how risks at predefined time points are calculated using baseline survival probabilities.

\textbf{Step 1: Assigning Quantiles}\\
Predicted risk scores are divided into quantiles to categorise individuals with similar risk predictions. In this study, deciles are used, grouping the population into 10 equally sized categories based on predicted risk percentages. For each individual, the quantile group is assigned as:
\[
\text{Quantile Group} = Q_q(\text{Predicted Risk}),
\]
where \( Q_q \) represents the quantile function, and \( q \) determines the number of quantiles (e.g., \( q=10 \) for deciles). This stratification facilitates a structured comparison between predicted and observed risks.

\textbf{Step 2: Aggregating Predictions and Events}\\
Within each quantile, two key statistics are computed:
    \[
    \text{Mean Predicted Risk} = \frac{\sum \text{Predicted Risk in Quantile}}{\text{Number of Individuals in Quantile}}.
    \]\\
    \[
    \text{Observed Events} = \sum \text{Events in Quantile}.
    \]
These metrics summarise the model’s predictions and the actual outcomes for each quantile, providing a clear basis for calibration assessment.

\textbf{Step 3: Computing Observed Event Rates}\\
Observed event rates are normalised for interpretability by scaling them to a rate per 1,000 individuals:
\[
\text{Event Rate per Quantile} = \frac{\text{Observed Events in Quantile}}{\text{Total Number of Individuals}} \times 1000.
\]
This scaling ensures consistency across quantiles, allowing a meaningful comparison between observed outcomes and predicted risks.

\textbf{Step 4: Visualising and Interpreting Calibration}\\
A calibration plot is generated to compare predicted and observed risks across quantiles. The x-axis represents the observed event rate (normalised percentage of events). The y-axis represents the mean predicted risk percentage. Deviations from this line highlight areas where the model overestimates (\( y > x \)) or underestimates (\( y < x \)) risk.

%%%===%%%%%%===%%%%%%===%%%
%%%===%%%%%%===%%%%%%===%%%
%%%===%%%%%%===%%%%%%===%%%
\newpage
\section{Appendix: Preprocessing and Postprocessing}\label{App:Processing}

This Appendix details the preprocessing and postprocessing procedures for our MCM framework.

\subsection{Preprocessing}\label{App:PreProcessing}

\colorbox{green!25}{\{}Numerical variables such as Age and BMI are first subjected to a Box-Cox transformation~\cite{box1964analysis} to stabilise variance and approximate a normal distribution. After transformation, these variables are standardised to ensure they have a mean of zero and a standard deviation of one. This standardisation is crucial for ensuring that the model’s parameters are not disproportionately influenced by variables on different scales.

Binary variables, including Sex and medical history are mapped to [0, 1]. This normalisation is necessary to ensure features can later be converted back into discrete binary values during postprocessing.\colorbox{green!25}{\}}

Following the descriptions given in the past two paragraphs (\textit{i.e.,} the material described in \colorbox{green!25}{\{} \colorbox{green!25}{\}}), all reconstruction targets now have values lying in [0, 1]; making it much simpler to be reconstructed using the \texttt{Softmax} function of the MLP in the second block of the MCM.

\begin{table}[h!]
    \small
    \centering
    \begin{tabular}{p{2cm}p{4cm}p{6cm}}
        \hline
        \textbf{Variable Type} & \textbf{Transformation Applied} & \textbf{Example Variables} \\
        \hline
        \hline
        \textbf{Numeric} & 
        Box-Cox Transformation,\newline Normalisation to [0, 1] & 
        Age,\newline Total cholesterol,\newline Serum creatinine,\newline eGFR,\newline Systolic blood pressure,\newline Diastolic blood pressure,\newline Body mass index,\newline Time to event \\
        \hline
        \textbf{Binary} & 
        Map to [0, 1] & 
        Sex,\newline Smoking history,\newline Obesity history,\newline Diabetes history,\newline Coronary heart disease,\newline Vascular disease,\newline Hypertension history,\newline Dyslipidaemia history,\newline Lipid-lowering medication,\newline Diabetes medication,\newline Blood pressure-lowering medication,\newline ACE inhibitors/angiotensin receptor blockers,\newline Event occurrence \\
        \hline
    \end{tabular}
    \caption{Preprocessing steps for the CKD EHR data.}
\end{table}

\subsection{Postprocessing}\label{App:PostProcessing}

We use postprocessing to revert the transformed variables to their original scale and format. The numerical variables undergo rescaling and inverse Box-Cox transformations to match the original data’s distribution and range. Binary variables are thresholded at 0.5 to convert the continuous outputs back into discrete binary values. This is applied to both the clinical covariates as well as the event outcome and time-to-event duration.

%###===>>>%###===>>>%###===>>>
%###===>>>%###===>>>%###===>>>
%###===>>>%###===>>>%###===>>>
\newpage
\section{Appendix: Implementation Details}

\begin{table}[h!]
    \small
    \centering
    \begin{tabular}{p{3cm}p{3cm}p{6cm}}
        \hline
        \textbf{Layer Name} & \textbf{Output Size} & \textbf{Layer Configuration} \\
        \hline
        \hline
        \textbf{input} & N $\times$ input\_dim & Input features\newline (\textit{e.g.,} clinical and demographic data) \\
        \textbf{attention1} & N $\times$ input\_dim & Attention weights applied to input features \\ 
        \textbf{mlp1\_hidden} & N $\times$ hidden\_dim & [\texttt{Linear(input\_dim, hidden\_dim)},\newline
        \texttt{ReLU},\newline
        \texttt{LayerNorm}] \\
        \textbf{mlp1\_output} & N $\times$ hidden\_dim & [\texttt{Linear(hidden\_dim, hidden\_dim)},\newline
        \texttt{ReLU},\newline
        \texttt{LayerNorm}] \\
        \textbf{attention2} & N $\times$ hidden\_dim & Attention weights applied to first MLP outputs \\ 
        \textbf{mlp2\_hidden} & N $\times$ hidden\_dim & [\texttt{Linear(hidden\_dim, hidden\_dim)},\newline
        \texttt{ReLU},\newline
        \texttt{LayerNorm}] \\
        \textbf{mlp2\_output} & N $\times$ input\_dim & [\texttt{Linear(hidden\_dim, input\_dim)},\newline
        \texttt{Sigmoid}] \\
        \hline
    \end{tabular}
    \caption{The Attention-MLP model architecture.}
    \label{tab:attention_mlp_architecture}
\end{table}

In this architecture, \textbf{N} denotes the batch size, \textbf{input\_dim} represents the dimensionality of the input features, and \textbf{hidden\_dim}, defaulted to 64, is a tunable hyperparameter determining the size of the hidden layers. The details of the attention layer can be found below.

\begin{table}[h!]
    \small
    \centering
    \begin{tabular}{p{3cm}p{3cm}p{6cm}}
        \hline
        \textbf{Layer Name} & \textbf{Output Size} & \textbf{Layer Configuration} \\
        \hline
        \hline
        \textbf{input} & $N \times \text{input\_dim}$ & Input features with optional masking to handle invalid positions. \\ 
        \textbf{attention\_weights} & $N \times \text{input\_dim}$ & \texttt{Linear(input\_dim, input\_dim,}\newline 
        \hspace*{12mm}\texttt{bias=False)} 
        \newline Produces raw attention scores for each feature. \\ 
        \textbf{masking} & $N \times \text{input\_dim}$ & Applies masking to attention scores using a binary mask. Invalid positions are set to $-\infty$ to exclude them during softmax computation. \\ 
        \textbf{softmax} & $N \times \text{input\_dim}$ & Normalises attention scores across the feature dimension to ensures the weights sum to 1. \\ 
        \textbf{weighted\_sum} & $N \times \text{input\_dim}$ & Element-wise multiplication of input features with attention weights.\\ 
        \hline
    \end{tabular}
    \caption{The Attention Layer architecture.}
    \label{tab:attention_layer_architecture}
\end{table}

The attention mechanism applied in this study is simple yet effective approach. Unlike the complex multi-head dot-product attention mechanisms employed in large language models (LLMs) such as Transformers~\cite{vaswani2017attention}, our attention mechanism is implemented through a straightforward linear layer.

Our model uses a two-block design, with the second block incorporating a residual connection and an all-ones mask to refine the output. The residual connection ensures retention of essential raw input information while enhancing feature expressiveness. Using an all-ones mask allowing the attention mechanism to focus on higher-order feature relationships.

%###===>>>%###===>>>%###===>>>
%###===>>>%###===>>>%###===>>>
%###===>>>%###===>>>%###===>>>
\newpage
\section{Appendix: Integrating MICE in Conditional Data Augmentation to\\
\hspace*{27mm}Further Enhanced Stratified Calibration}\label{App:McmMice}

This appendix provides a detailed account of the integration of Multiple Imputation by Chained Equations (MICE) into MCM and all alternative techniques for data simulation. This hybrid approach refines the imputation of critical survival-related variables, addressing limitations of standalone MCM and offering a robust alternative for generating datasets that align closely with observed distributions.

\subsection{Methodological Overview}

MCM remains the cornerstone of synthetic data generation (see Algorithm \ref{alg:StratifiedCalibration}). By masking a predefined proportion of features and reconstructing the masked values, MCM leverages its pre-trained understanding of the dataset to produce simulated data that aligns closely with the distributions of the original data. However, challenges arise when outcome variables, specifically event indicators and time-to-event durations, are excluded from masking and instead serve as independent variables for reconstructing other features.

First, at a masking ratio of 50\%, critical dependencies between clinical covariates can be simultaneously masked. For instance, if eGFR is closely dependent on age, and age is a key determinant of time-to-event durations, masking both variables concurrently means that eGFR cannot be reconstructed in a way that reliably reflects plausible durations with high precision. This decoupling of essential relationships undermines the fidelity of the simulated data.

Second, the dual importance of event indicators and time-to-event durations exacerbates this issue. When both outcomes, along with key risk factors, are masked, the reconstruction process relies solely on the remaining unmasked variables, which may lack sufficient predictive power. This limitation is less problematic for generating standalone clinical datasets intended for general access, where slight imprecisions may be acceptable. However, for stratified calibration -- a process highly sensitive to precision akin to controlling type I and type II errors -- such inaccuracies can significantly impair the quality of the calibration.

Due to these reasons, while MCM excels at reconstructing masked features, it is less tailored for directly imputing survival-related target variables such as time to event (\( T \)) and event occurrence (\( E \)). To address this, MICE is introduced after the MCM simulation phase to iteratively impute these critical variables. MICE estimates missing values based on multivariate relationships among all features, offering a more statistically rigorous imputation process that accounts for the dependencies between covariates and target variables.

\subsection{Implementation Details}

A comprehensive pseudocode can be found in Algorithm \ref{alg:MCM_MICE_Strategy}.

\subsubsection{Data Preparation with MCM}
The process begins with MCM pre-trained on the entire dataset. For each stratified subgroup defined by the stratification condition \( Z^\dagger \), MCM generates synthetic datasets dynamically during the 5x2 cross-validation process. At this stage, the synthetic dataset is complete but lacks values for the survival-related target variables \( T \) and \( E \).

\subsubsection*{B.2. Imputation with MICE}
To impute \( T \) and \( E \), the MCM-generated synthetic data is combined with the original dataset:

\begin{itemize}
    \item \textbf{Introducing Missing Placeholders}: In the synthetic dataset, placeholders (e.g., NaN values) are introduced for \( T \) and \( E \), ensuring these fields are treated as missing values.
    \item \textbf{Combining Data}: The synthetic dataset is concatenated with the observed data for the same stratified subgroup to benefit from the outcome characteristics in observed data.
    \item \textbf{Iterative Imputation}: Use MICE to impute the missing values for \( T \) and \( E \). This process iteratively models each variable as a function of all others, refining the imputations in successive steps until convergence.
\end{itemize}

\newpage
\begin{algorithm}[h]
\caption{5x2 Cross-Validation with MCM and MICE Integration\\
\hspace*{19mm}for Enhanced Stratified Risk Simulation}
\label{alg:MCM_MICE_Strategy}
\begin{algorithmic}[1]
\Require Clinical dataset \( \mathcal{D} \) with covariates \( \mathbf{X} \) and event information,\\
\hspace*{9mm}stratification target variable \( Z^\dagger \), masking ratio \( r \), and number of iterations \( I \),\\
\hspace*{9mm}and set of time points \( \mathcal{T} \)

\Procedure{5x2 Cross-Validation with Simulated Data Using MCM and MICE}{}
            \For{each iteration \( i = 1, \dots, I \)}
                \For{each fold \( f = 1, \dots, 5 \)}
                    \For{each split \( s = 0, 1 \)}
                        \State \textbf{Split Dataset}: Create training set \( \mathcal{D}_{\text{train}} \) and test set \( \mathcal{D}_{\text{test}} \) based on \( f \) and \( s \)
                        \State \textbf{Filter Training Data}: Select subset \( \mathcal{D}_{\text{strat}} \subseteq \mathcal{D}_{\text{train}} \) meeting \( Z^\dagger \)
                        \State \colorbox{yellow!25}{\textbf{Generate Simulated Data Using MCM}:}\\
                            \hspace*{27mm}\colorbox{yellow!25}{Apply masking on \( \mathcal{D}_{\text{strat}} \) based on ratio \( r \)}\\
                            \hspace*{27mm}\colorbox{yellow!25}{Reconstruct masked features using the MCM model to get $\mathcal{D}_{\text{sim}}$}\\
                            \hspace*{27mm}\colorbox{yellow!25}{Mask the outcome variables in $\mathcal{D}_{\text{sim}}$ with \texttt{NaN}}
                        
                        \State \colorbox{yellow!25}{\textbf{Integrate with MICE}:}\\ 
                        \hspace*{27mm}\colorbox{yellow!25}{Combine the current training \& simulated \( \mathcal{D}_{\text{train}} \gets \mathcal{D}_{\text{train}} \cup \mathcal{D}_{\text{sim}} \)}\\
                        \hspace*{27mm}\colorbox{yellow!25}{Use MICE to impute event and duration variables}
                        \State \textbf{Fit CoxPH Model on Augmented Training Data}
                        \State \textbf{Compute LPH for Test Set} for each \( t \in \mathcal{T} \)
                    \EndFor
                \EndFor
            \EndFor
\EndProcedure
\end{algorithmic}
\end{algorithm}

The pseudocode is colour-coded for clarity, with sections highlighted in \colorbox{yellow!25}{yellow} indicating the integration of MICE. This addition is specifically designed to enhance the precision of outcome variables in the MCM-simulated conditional data, optimising the process for stratified calibration. The remainder of the pseudocode closely follows the structure of Algorithm \ref{alg:StratifiedCalibration}, ensuring consistency while accommodating the added functionality. Additionally, the modular design allows line 11 to be easily replaced with alternative data simulation techniques, such as SMOTE or VAEs. This flexibility supports seamless integration of diverse methods, enabling a robust framework for fair and comprehensive comparisons.

%###===>>>%###===>>>%###===>>>
%###===>>>%###===>>>%###===>>>
%###===>>>%###===>>>%###===>>>
\newpage
\section{Appendix: More Results}
\subsection{Realism Check}\label{App:ExtraResults1}

\begin{figure}[h]
    \centering
    \includegraphics[width=0.72\linewidth]{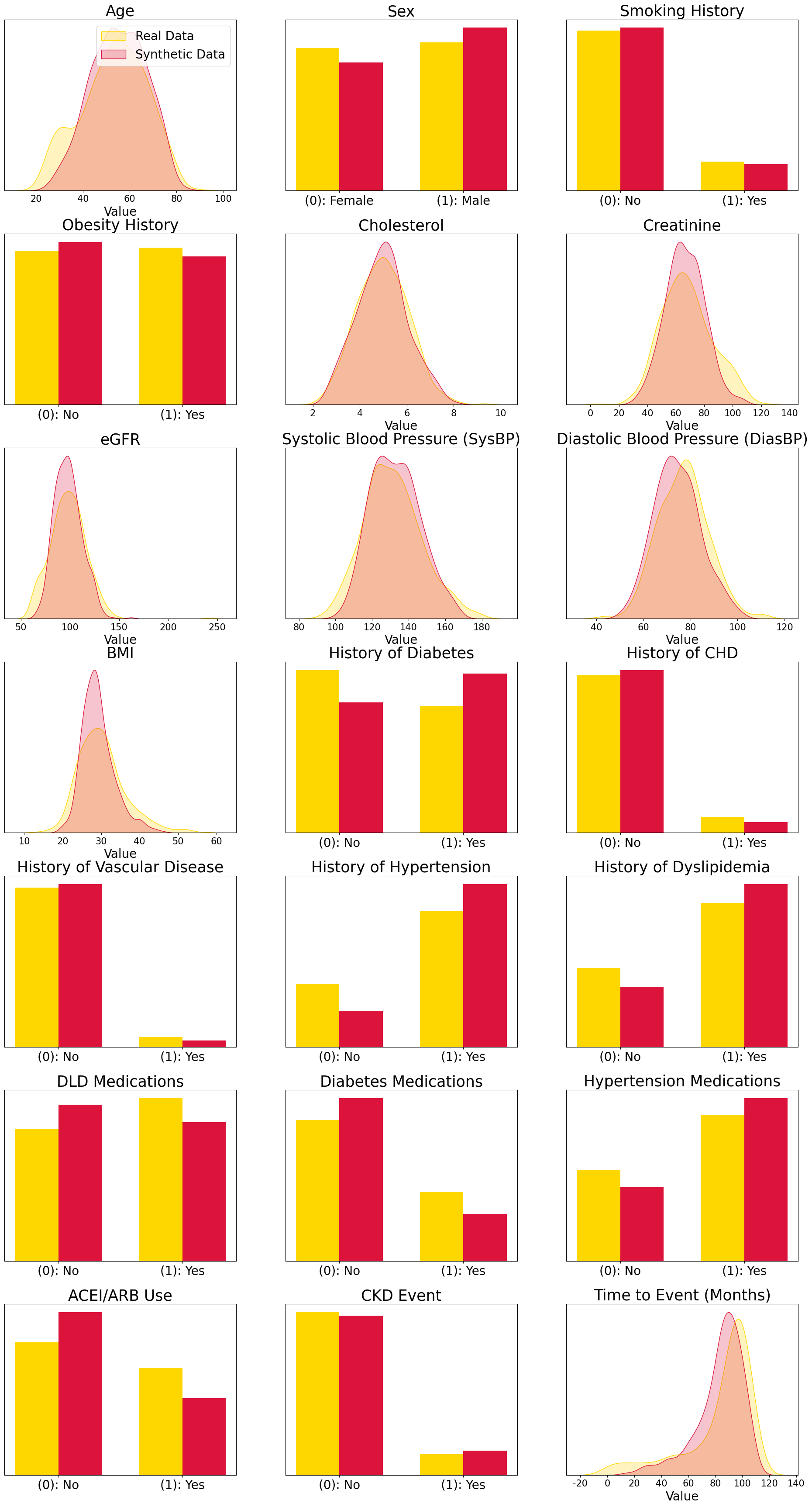}
    \caption{Comparison of real variables (gold) and synthetic counterparts (crimson) from the CKD EHR dataset generated using VAE.}
    \label{Fig:CkdDist_VAE}
\end{figure}

%###===>>>%###===>>>%###===>>>
\newpage
\begin{figure}[h]
    \centering
    \includegraphics[width=0.72\linewidth]{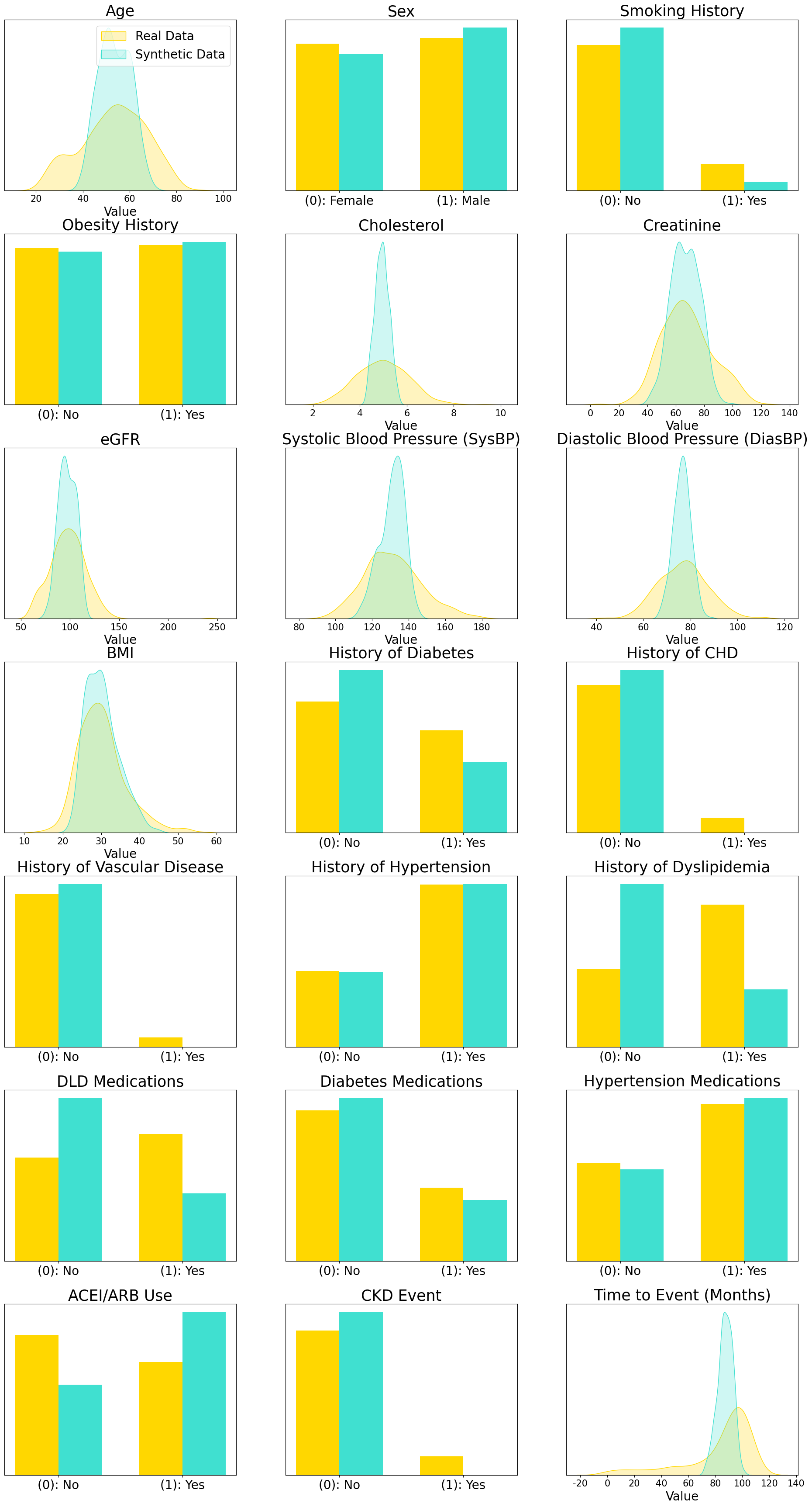}
    \caption{Comparison of real variables (gold) and synthetic counterparts (turquoise) from the CKD EHR dataset generated using WGAN.}
    \label{Fig:CkdDist_WGAN}
\end{figure}

%###===>>>%###===>>>%###===>>>
\newpage
\begin{figure}[h]
    \centering
    \includegraphics[width=0.72\linewidth]{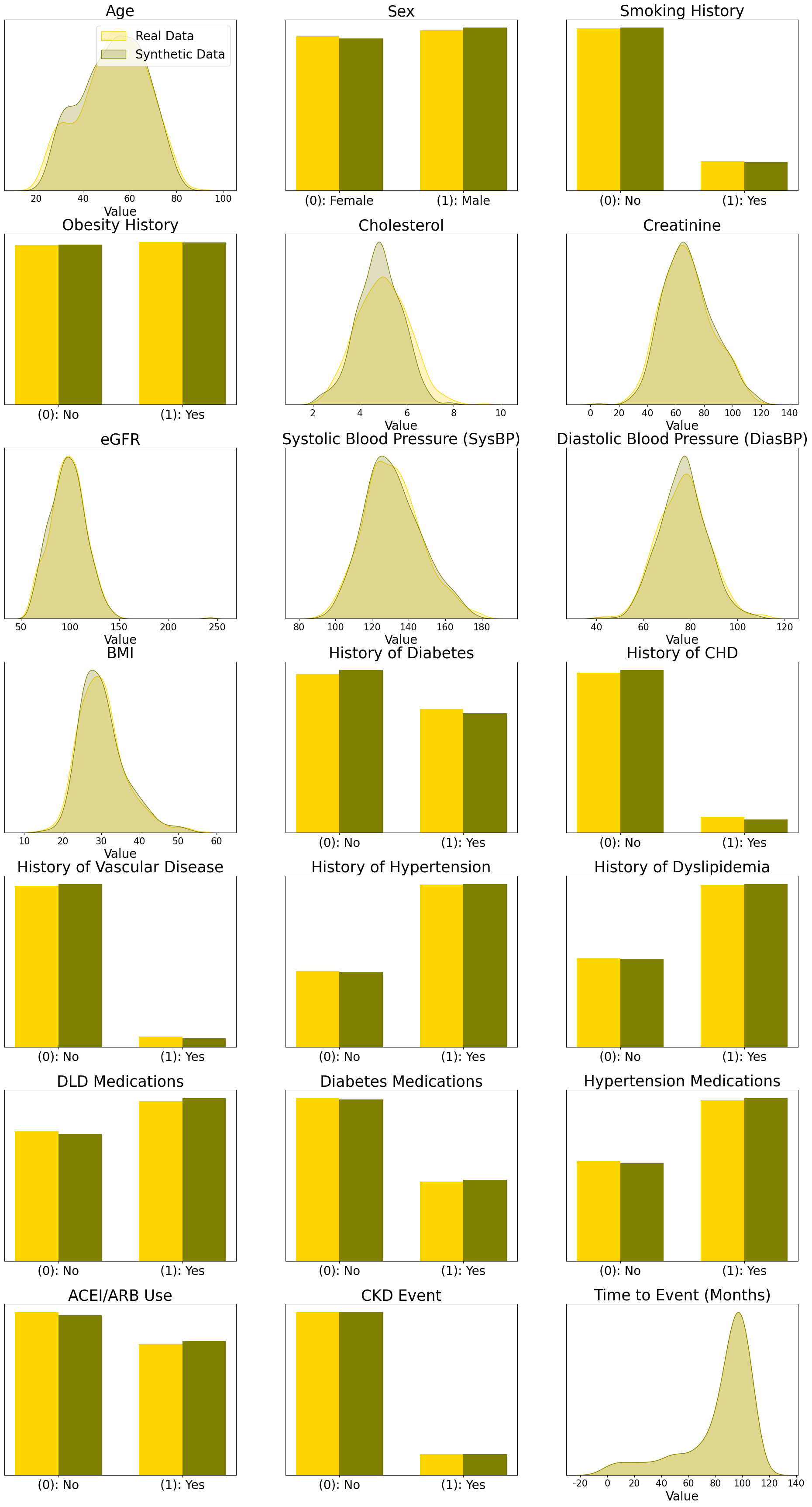}
    \caption{Comparison of real variables (gold) and synthetic counterparts (olive) from the CKD EHR dataset generated using CK4Gen.}
    \label{Fig:CkdDist_CK4Gen}
\end{figure}

%###===>>>%###===>>>%###===>>>
\newpage
\begin{figure}[h]
    \centering
    \includegraphics[width=0.88\linewidth]{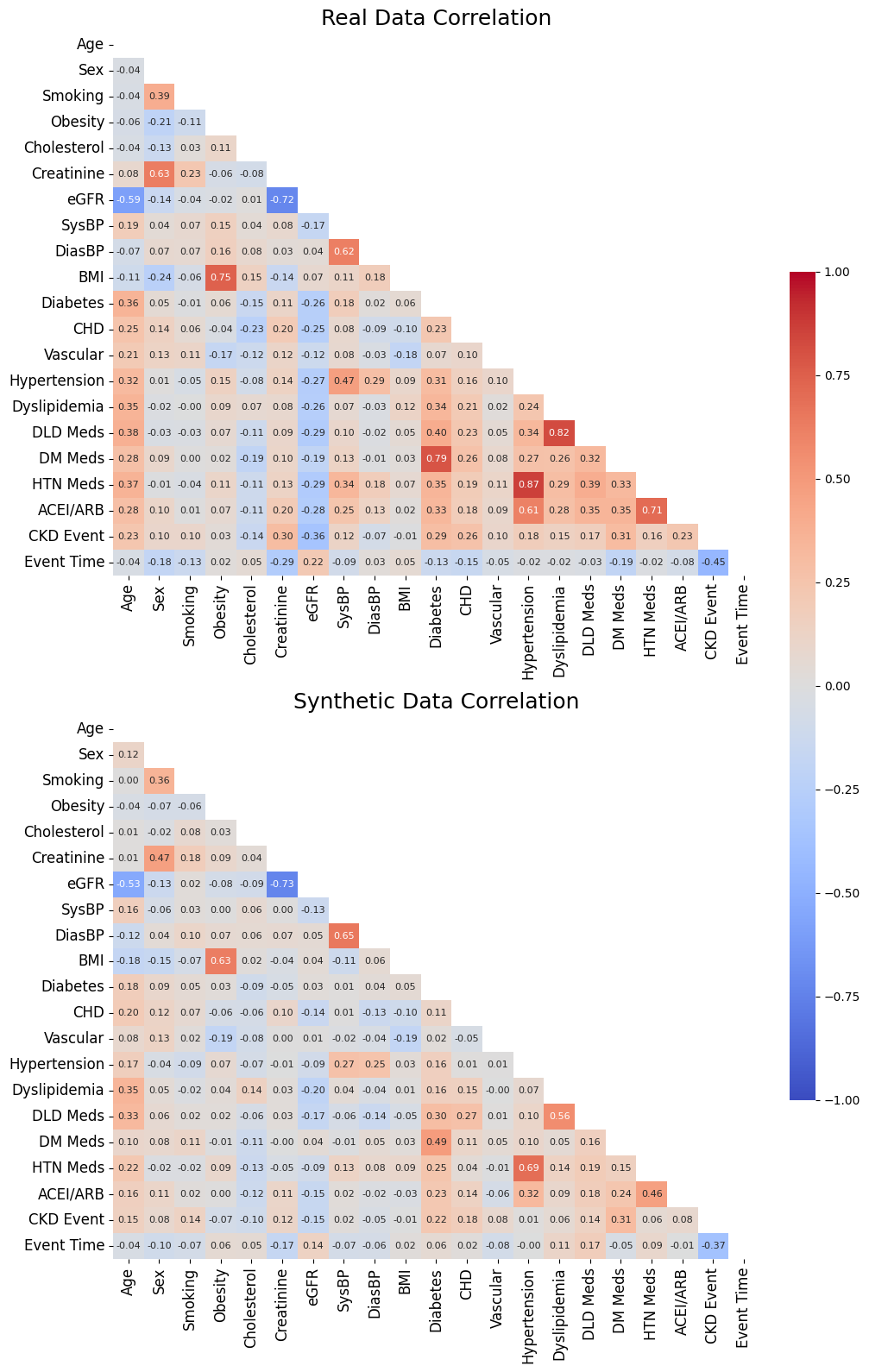}
    \caption{Complete Correlation, generated using VAE.}
    \label{Fig:CorrFull_VAE}
\end{figure}

%###===>>>%###===>>>%###===>>>
\newpage
\begin{figure}[h]
    \centering
    \includegraphics[width=0.88\linewidth]{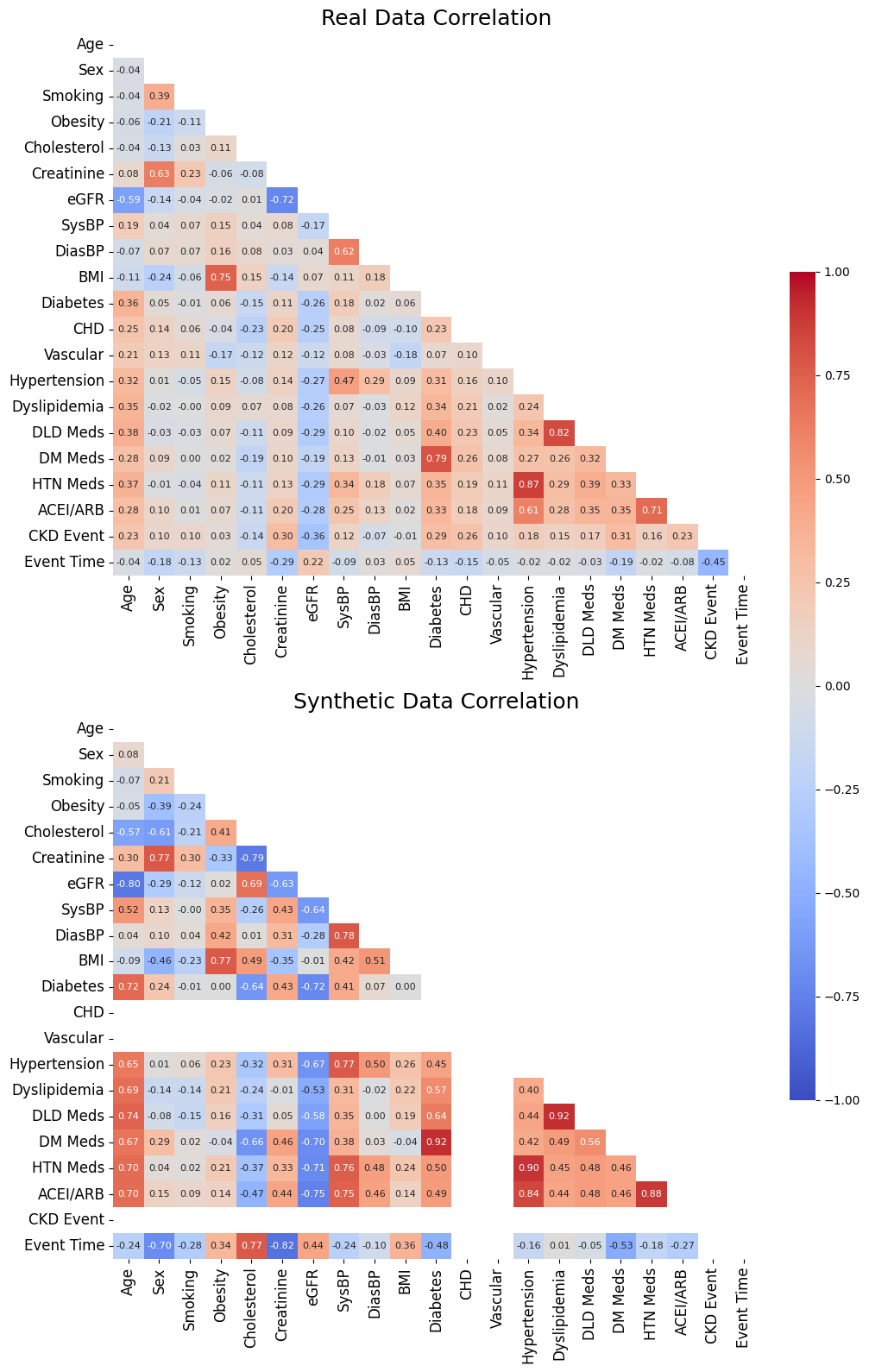}
    \caption{Complete Correlation, generated using WGAN.}
    \label{Fig:CorrFull_WGAN}
\end{figure}

%###===>>>%###===>>>%###===>>>
\newpage
\begin{figure}[h]
    \centering
    \includegraphics[width=0.88\linewidth]{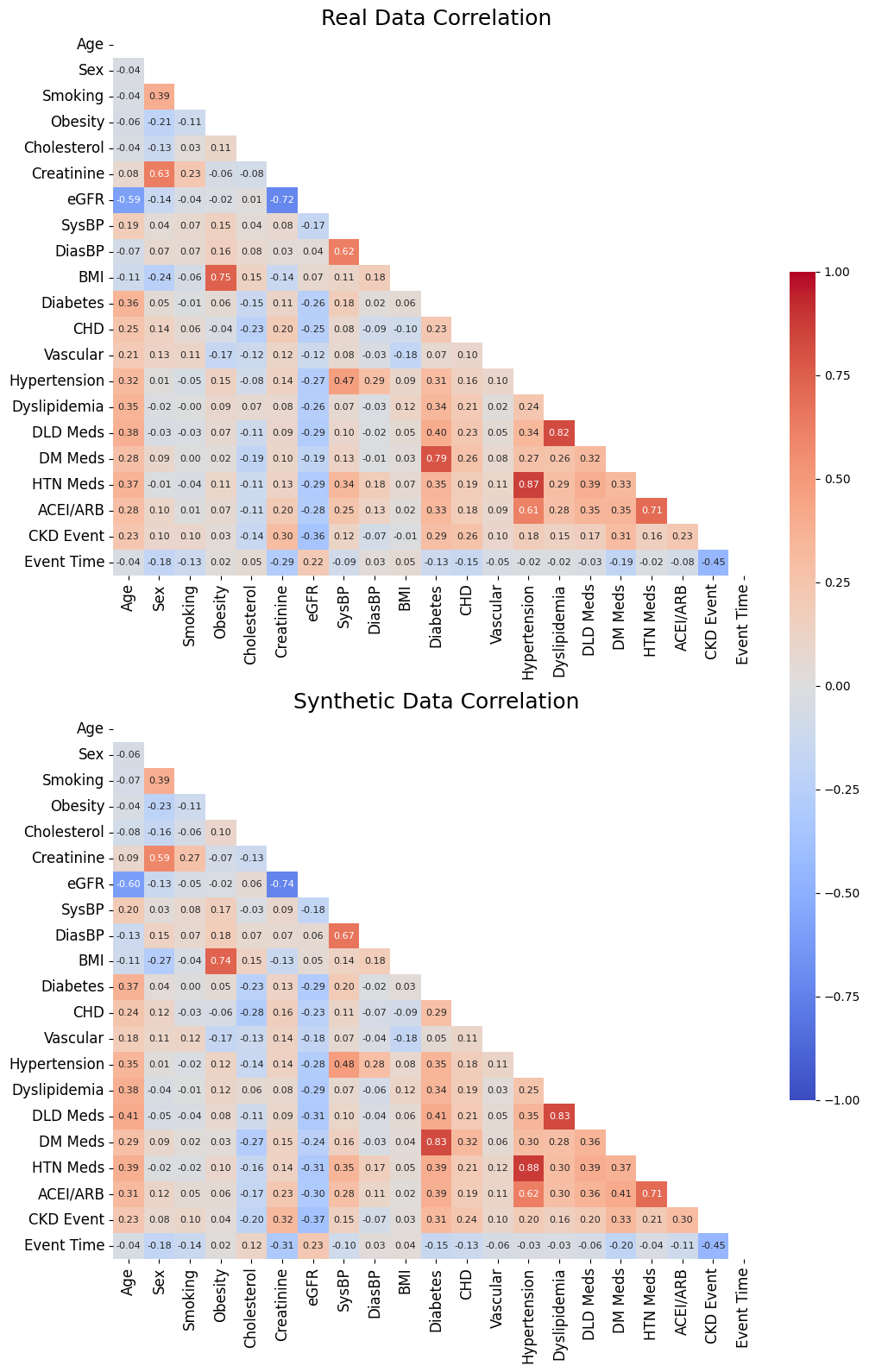}
    \caption{Complete Correlation, generated using CK4Gen.}
    \label{Fig:CorrFull_CK4Gen}
\end{figure}

%###===>>>%###===>>>%###===>>>
%###===>>>%###===>>>%###===>>>
%###===>>>%###===>>>%###===>>>
\newpage
\subsection{Utility Verification}\label{App:ExtraResults2}

%#-----------------------------------------------
\begin{figure}[h]
    \centering
    \includegraphics[width=\linewidth]{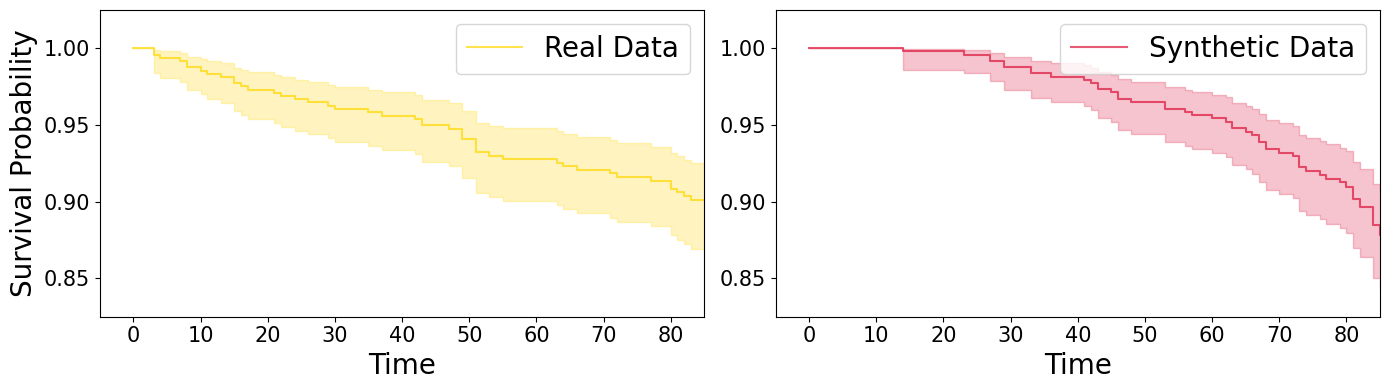}
    \caption{Comparison of KM curves between real and synthetic data generated using VAE.}
    \label{Fig:KmCurve_VAE}
\end{figure}

%#-----------------------------------------------
\begin{figure}[h]
    \centering
    \includegraphics[width=\linewidth]{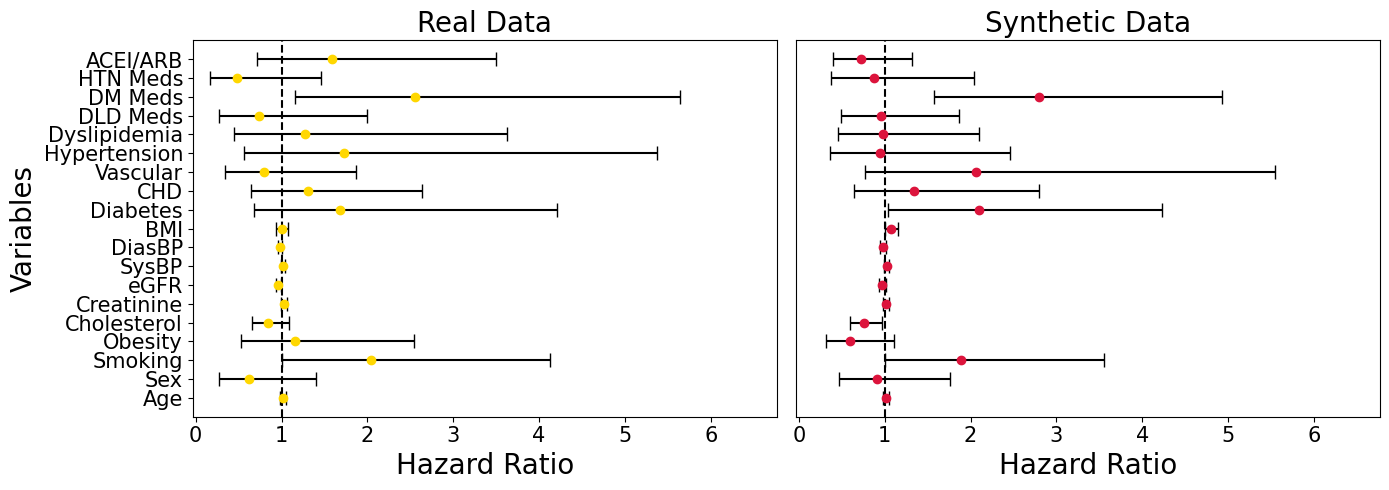}
    \caption{Comparison of HR consistencies between real and synthetic data generated using VAE.}
    \label{Fig:HrConsistency_VAE}
\end{figure}

%#-----------------------------------------------
\newpage
\begin{figure}[h]
    \centering
    \includegraphics[width=\linewidth]{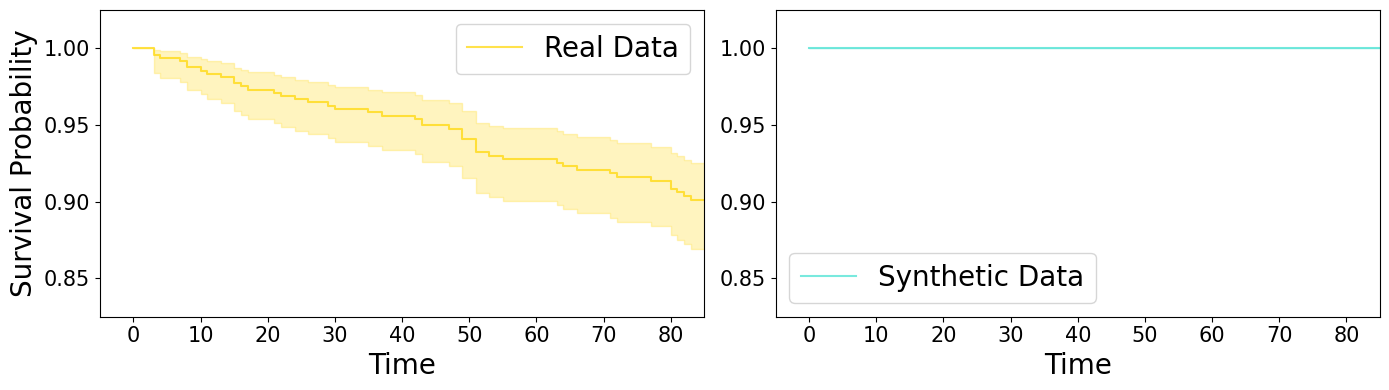}
    \caption{Comparison of KM curves between real and synthetic data generated using WGAN.}
    \label{Fig:KmCurve_WGAN}
\end{figure}

%#-----------------------------------------------
\begin{figure}[h]
    \centering
    \includegraphics[width=\linewidth]{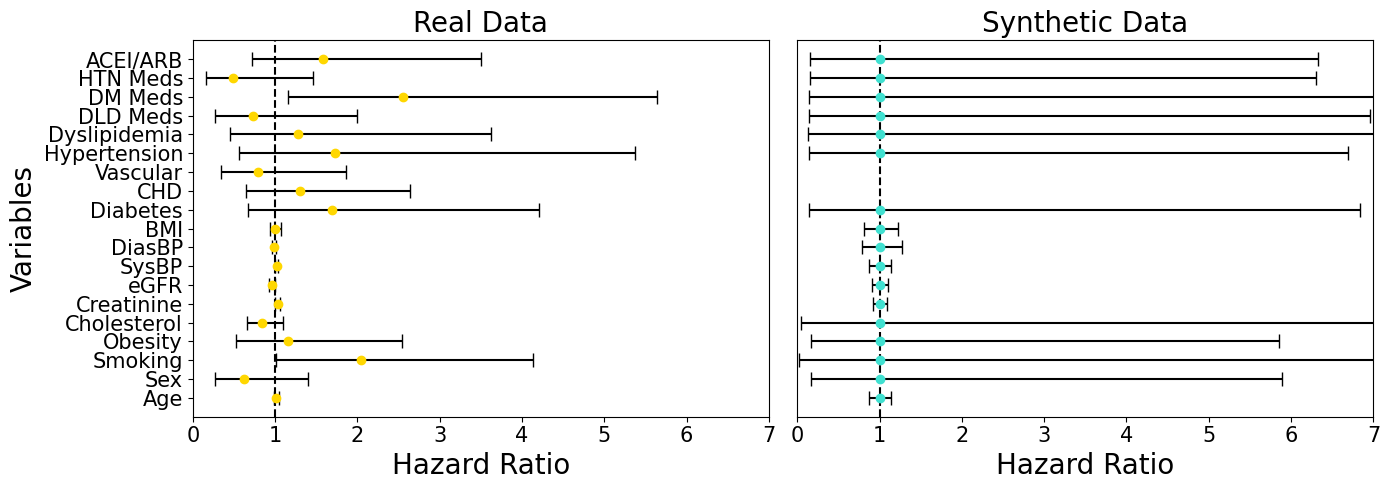}
    \caption{Comparison of HR consistencies between real and synthetic data generated using WGAN.}
    \label{Fig:HrConsistency_WGAN}
\end{figure}

%#-----------------------------------------------
\newpage
\begin{figure}[h]
    \centering
    \includegraphics[width=\linewidth]{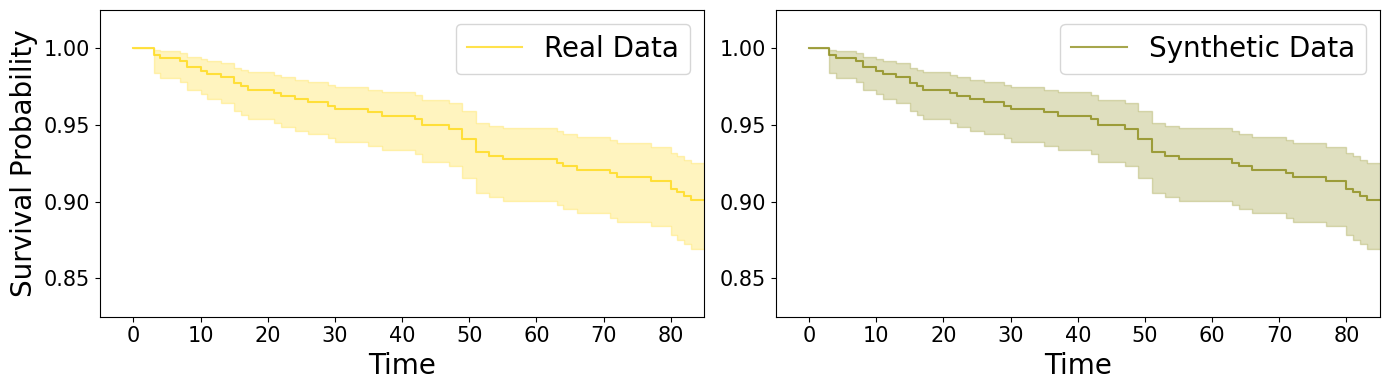}
    \caption{Comparison of KM curves between real and synthetic data generated using CK4Gen.}
    \label{Fig:KmCurve_CK4Gen}
\end{figure}

%#-----------------------------------------------
\begin{figure}[h]
    \centering
    \includegraphics[width=\linewidth]{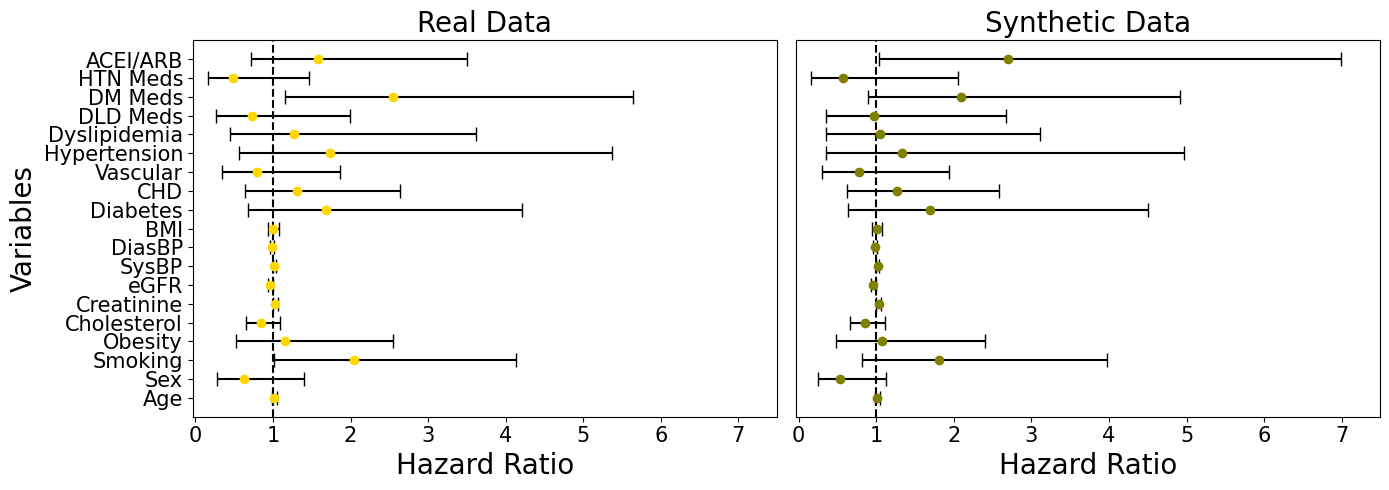}
    \caption{Comparison of HR consistencies between real and synthetic data generated using CK4Gen.}
    \label{Fig:HrConsistency_CK4Gen}
\end{figure}

%###===>>>%###===>>>%###===>>>
%###===>>>%###===>>>%###===>>>
%###===>>>%###===>>>%###===>>>
\newpage
\subsection{Practicality Analysis}\label{App:ExtraResults3}
\begin{table}[h]
\tiny
\centering
\caption{Calibration Target: eGFR}
\label{tab:calibration_loss_egfr_0_mice}
\begin{tabular}{|p{3.0cm}||p{1.75cm}|p{1.75cm}|p{1.75cm}|p{1.75cm}|p{1cm}|}

\multicolumn{6}{c}{\footnotesize{Stratification On: Normal Renal Function (eGFR $\geq$ 90 mL/min/1.73m²)}}\\
\hline
\textbf{Augmentation Method} & 
\textbf{Duration Target:\newline 25th Percentile} &
\textbf{Duration Target:\newline 50th Percentile} &
\textbf{Duration Target:\newline 75th Percentile} &
\textbf{Sum of\newline Calibration Loss} & 
\textbf{Rank} \\ 
\hline\hline

\cellcolor{green!20}\textbf{No augmentation}
& \cellcolor{green!20}0.3183
& \cellcolor{green!20}0.4603                
& \cellcolor{green!20}0.5086                      
& \cellcolor{green!20}1.2872              
& \cellcolor{green!20}11 \\ 
\hline\hline

\cellcolor{blue!20}\textbf{MCM + MICE}
& \cellcolor{blue!20}0.3079 (0.0017)
& \cellcolor{blue!20}0.0617 (0.0012)
& \cellcolor{blue!20}0.1853 (0.0009)
& \cellcolor{blue!20}0.5549
& \cellcolor{blue!20}2 \\ 
\hline\hline

\textbf{RandomOverSampler + MICE}   
& 0.1241 (0.0918)
& 0.2356 (0.0980)
& 0.3383 (0.0824)
& 0.6980
& 9 \\ 
\hline

\textbf{SMOTE + MICE}
& 0.2975 (0.1021)
& 0.0794 (0.0648)
& 0.1952 (0.0641)
& 0.5721
& 4 \\ 
\hline

\textbf{SMOTENC + MICE}             
& 0.2282 (0.0710)
& 0.1194 (0.0391)
& 0.2427 (0.0285)
& 0.5903
& 7 \\ 
\hline

\textbf{ADASYN + MICE}
& 0.2412 (0.1206)
& 0.1129 (0.0803)
& 0.2275 (0.0719)
& 0.5816
& 5 \\ 
\hline

\textbf{BorderlineSMOTE + MICE}     
& 0.1991 (0.0688)
& 0.1312 (0.0652)
& 0.2403 (0.0625)
& 0.5706
& 3 \\ 
\hline

\textbf{SVMSMOTE + MICE}
& 0.2227 (0.0649)
& 0.1223 (0.0447)
& 0.2369 (0.0386)
& 0.5819
& 6 \\ 
\hline

\textbf{VAE + MICE}
& 0.0522 (0.0246)
& 0.2643 (0.0303)
& 0.3533 (0.0217)
& 0.6698
& 8 \\ 
\hline

\textbf{WGAN + MICE}
& 0.2326 (0.0389)
& 0.4405 (0.0272)
& 0.5096 (0.0233)
& 1.1827
& 10 \\ 
\hline

\cellcolor{purple!20}\textbf{CK4Gen + MICE}
& \cellcolor{purple!20}0.3266 (0.0000)
& \cellcolor{purple!20}0.0489 (0.0000)
& \cellcolor{purple!20}0.1749 (0.0000)
& \cellcolor{purple!20}0.5504
& \cellcolor{purple!20}1 \\ 
\hline
%###===>>>%###===>>>%###===>>>
%###===>>>%###===>>>%###===>>>
%###===>>>%###===>>>%###===>>>
\multicolumn{6}{c}{}\\
\multicolumn{6}{c}{\footnotesize{Stratification On: Non-Ideal Renal Function (eGFR $<$ 90 mL/min/1.73m²)}}\\
\hline
\textbf{Augmentation Method} & 
\textbf{Duration Target:\newline 25th Percentile} &
\textbf{Duration Target:\newline 50th Percentile} &
\textbf{Duration Target:\newline 75th Percentile} &
\textbf{Sum of\newline Calibration Loss} & 
\textbf{Rank} \\ 
\hline\hline

\cellcolor{green!20}\textbf{No Augmentation}
& \cellcolor{green!20}0.6959
& \cellcolor{green!20}0.7375                
& \cellcolor{green!20}0.752                      
& \cellcolor{green!20}2.1854              
& \cellcolor{green!20}9 \\ 
\hline\hline

\cellcolor{blue!20}\textbf{MCM + MICE}
& \cellcolor{blue!20}0.6326 (0.0015)
& \cellcolor{blue!20}0.6987 (0.0010)
& \cellcolor{blue!20}0.7214 (0.0008)
& \cellcolor{blue!20}2.0527
& \cellcolor{blue!20}2 \\ 
\hline\hline

\textbf{RandomOverSampler + MICE}   
& 0.6372 (0.0316)
& 0.7035 (0.0194)
& 0.7265 (0.0157)
& 2.0672
& 4 \\ 
\hline

\textbf{SMOTE + MICE}
& 0.6517 (0.0231)
& 0.7096 (0.0135)
& 0.7296 (0.0106)
& 2.0909
& 6 \\ 
\hline

\textbf{SMOTENC + MICE}             
& 0.6349 (0.0321)
& 0.6998 (0.0183)
& 0.7225 (0.0135)
& 2.0572
& 3 \\ 
\hline

\textbf{ADASYN + MICE}
& 0.6539 (0.0224)
& 0.7117 (0.0134)
& 0.7317 (0.0106)
& 2.0973
& 8 \\ 
\hline

\textbf{BorderlineSMOTE + MICE}     
& 0.6492 (0.0234)
& 0.7082 (0.0136)
& 0.7285 (0.0105)
& 2.0859
& 5 \\ 
\hline

\textbf{SVMSMOTE + MICE}
& 0.6541 (0.0281)
& 0.7111 (0.0163)
& 0.7310 (0.0129)
& 2.0962
& 7 \\ 
\hline

\textbf{VAE + MICE}
& 0.6973 (0.0041)
& 0.7371 (0.0029)
& 0.7512 (0.0024)
& 2.1856
& 10 \\ 
\hline

\textbf{WGAN + MICE}
& 0.7118 (0.0016)
& 0.7444 (0.0013)
& 0.7554 (0.0011)
& 2.2116
& 11 \\ 
\hline

\cellcolor{purple!20}\textbf{CK4Gen + MICE}
& \cellcolor{purple!20}0.6275 (0.0000)
& \cellcolor{purple!20}0.6955 (0.0000)
& \cellcolor{purple!20}0.7188 (0.0000)
& \cellcolor{purple!20}2.0418
& \cellcolor{purple!20}1 \\ 
\hline
\end{tabular}
\end{table}

%###===>>>%###===>>>%###===>>>
%###===>>>%###===>>>%###===>>>
%###===>>>%###===>>>%###===>>>
\begin{table}[ht]
\tiny
\centering
\caption{Calibration Target: Diabetes Status}
\label{tab:calibration_loss_dia_fin_0_mice}
\begin{tabular}{|p{3.5cm}||p{1.75cm}|p{1.75cm}|p{1.75cm}|p{1.75cm}|p{1cm}|}
\multicolumn{6}{c}{\footnotesize{Stratification On: No Diabetes}}\\
\hline
\textbf{Augmentation Method} & 
\textbf{Duration Target:\newline 25th Percentile} &
\textbf{Duration Target:\newline 50th Percentile} &
\textbf{Duration Target:\newline 75th Percentile} &
\textbf{Sum of\newline Calibration Loss} & 
\textbf{Rank} \\ 
\hline\hline

\cellcolor{green!20}\textbf{No Augmentation}
& \cellcolor{green!20}0.0268
& \cellcolor{green!20}0.2583                
& \cellcolor{green!20}0.3536                      
& \cellcolor{green!20}0.6387              
& \cellcolor{green!20}1 \\ 
\hline\hline

\cellcolor{blue!20}\textbf{MCM + MICE}
& \cellcolor{blue!20}1.1645 (0.0840)
& \cellcolor{blue!20}0.5466 (0.0597)
& \cellcolor{blue!20}0.3434 (0.0517)
& \cellcolor{blue!20}2.0545
& \cellcolor{blue!20}9 \\ 
\hline

\textbf{RandomOverSampler + MICE}   
& 1.4702 (0.4508)
& 0.7725 (0.3242)
& 0.5391 (0.2844)
& 2.7818
& 11 \\ 
\hline

\textbf{SMOTE + MICE}
& 1.0470 (0.4404)
& 0.5042 (0.2263)
& 0.3583 (0.1290)
& 1.9095
& 7 \\ 
\hline

\textbf{SMOTENC + MICE}             
& 1.1795 (0.4257)
& 0.5589 (0.3059)
& 0.4053 (0.1779)
& 2.1437
& 10 \\ 
\hline

\textbf{ADASYN + MICE}
& 1.0744 (0.3850)
& 0.4716 (0.2832)
& 0.3209 (0.1883)
& 1.8669
& 6 \\ 
\hline

\textbf{BorderlineSMOTE + MICE}     
& 0.8040 (0.4317)
& 0.3243 (0.2634)
& 0.2443 (0.1614)
& 1.3726
& 4 \\ 
\hline

\textbf{SVMSMOTE + MICE}
& 1.0586 (0.1685)
& 0.4679 (0.1191)
& 0.2766 (0.1035)
& 1.8031
& 5 \\ 
\hline

\textbf{VAE + MICE}
& 0.1660 (0.0862)
& 0.3963 (0.0624)
& 0.4727 (0.0544)
& 1.035
& 2 \\ 
\hline

\textbf{WGAN + MICE}
& 0.2827 (0.0900)
& 0.4769 (0.0634)
& 0.5378 (0.0559)
& 1.2974
& 3 \\ 
\hline

\textbf{CK4Gen + MICE}
& 1.1469 (0.0000)
& 0.5344 (0.0000)
& 0.3331 (0.0000)
& 2.0144
& 8 \\ 
\hline
%###===>>>%###===>>>%###===>>>
%###===>>>%###===>>>%###===>>>
%###===>>>%###===>>>%###===>>>
\multicolumn{6}{c}{}\\
\multicolumn{6}{c}{\footnotesize{Stratification On: Diabetes}}\\
\hline
\textbf{Augmentation Method} & 
\textbf{Duration Target:\newline 25th Percentile} &
\textbf{Duration Target:\newline 50th Percentile} &
\textbf{Duration Target:\newline 75th Percentile} &
\textbf{Sum of\newline Calibration Loss} & 
\textbf{Rank} \\ 
\hline\hline

\cellcolor{green!20}\textbf{No Augmentation}
& \cellcolor{green!20}0.5898
& \cellcolor{green!20}0.6458                
& \cellcolor{green!20}0.6655                      
& \cellcolor{green!20}1.9011              
& \cellcolor{green!20}9 \\ 
\hline\hline

\cellcolor{blue!20}\textbf{MCM + MICE}
& \cellcolor{blue!20}0.5244 (0.0037)
& \cellcolor{blue!20}0.5992 (0.0030)
& \cellcolor{blue!20}0.6259 (0.0029)
& \cellcolor{blue!20}1.7495
& \cellcolor{blue!20}7 \\ 
\hline

\cellcolor{purple!20}\textbf{RandomOverSampler + MICE}   
& \cellcolor{purple!20}0.4808 (0.0460)
& \cellcolor{purple!20}0.5767 (0.0238)
& \cellcolor{purple!20}0.6107 (0.0167)
& \cellcolor{purple!20}1.6682
& \cellcolor{purple!20}1 \\ 
\hline

\textbf{SMOTE + MICE}
& 0.5247 (0.0301)
& 0.6022 (0.0153)
& 0.6294 (0.0103)
& 1.7563
& 8 \\ 
\hline

\textbf{SMOTENC + MICE}             
& 0.4911 (0.0335)
& 0.5808 (0.0166)
& 0.6135 (0.0111)
& 1.6854
& 2 \\ 
\hline

\textbf{ADASYN + MICE}
& 0.5197 (0.0237)
& 0.5983 (0.0113)
& 0.6261 (0.0073)
& 1.7441
& 5 \\ 
\hline

\textbf{BorderlineSMOTE + MICE}     
& 0.5214 (0.0312)
& 0.5985 (0.0155)
& 0.6257 (0.0103)
& 1.7456
& 6 \\ 
\hline

\textbf{SVMSMOTE + MICE}
& 0.5097 (0.0310)
& 0.5937 (0.0146)
& 0.6233 (0.0095)
& 1.7267
& 3 \\ 
\hline

\textbf{VAE + MICE}
& 0.5966 (0.0051)
& 0.6476 (0.0028)
& 0.6660 (0.0020)
& 1.9102
& 10 \\ 
\hline

\textbf{WGAN + MICE}
& 0.6020 (0.0053)
& 0.6467 (0.0032)
& 0.6621 (0.0025)
& 1.9108
& 11 \\ 
\hline

\textbf{CK4Gen + MICE}
& 0.5194 (0.0000)
& 0.5975 (0.0000)
& 0.6255 (0.0000)
& 1.7424
& 4 \\ 
\hline
\end{tabular}
\end{table}

%###===>>>%###===>>>%###===>>>
%###===>>>%###===>>>%###===>>>
%###===>>>%###===>>>%###===>>>
%\newpage
\begin{table}[ht]
\tiny
\centering
\caption{Calibration Target: Hypertension Status}
\label{tab:calibration_loss_hyper_fin_0_mice}
\begin{tabular}{|p{3.5cm}||p{1.75cm}|p{1.75cm}|p{1.75cm}|p{1.75cm}|p{1cm}|}
\multicolumn{6}{c}{\footnotesize{Stratification On: No Hypertension}}\\
\hline
\textbf{Augmentation Method} & 
\textbf{Duration Target:\newline 25th Percentile} &
\textbf{Duration Target:\newline 50th Percentile} &
\textbf{Duration Target:\newline 75th Percentile} &
\textbf{Sum of\newline Calibration Loss} & 
\textbf{Rank} \\ 
\hline\hline

\cellcolor{green!20}\textbf{No Augmentation}
& \cellcolor{green!20}0.359
& \cellcolor{green!20}0.5224                
& \cellcolor{green!20}0.5778                      
& \cellcolor{green!20}1.4592              
& \cellcolor{green!20}8 \\ 
\hline\hline

\cellcolor{blue!20}\textbf{MCM + MICE}
& \cellcolor{blue!20}0.4993 (0.0106)
& \cellcolor{blue!20}0.0824 (0.0075)
& \cellcolor{blue!20}0.0534 (0.0065)
& \cellcolor{blue!20}0.6351
& \cellcolor{blue!20}2 \\ 
\hline

\textbf{RandomOverSampler + MICE} 
& 0.9091 (0.3078)
& 0.4873 (0.2440)
& 0.3525 (0.2755)
& 1.7489
& 9 \\ 
\hline

\textbf{SMOTE + MICE}
& 0.6426 (0.2269)
& 0.1811 (0.1604)
& 0.1044 (0.0977)
& 0.9281
& 6 \\ 
\hline

\textbf{SMOTENC + MICE}             
& 1.0574 (0.3550)
& 0.4938 (0.2578)
& 0.3061 (0.2255)
& 1.8573
& 11 \\ 
\hline

\textbf{ADASYN + MICE}
& 0.5841 (0.2613)
& 0.1446 (0.1799)
& 0.1335 (0.0879)
& 0.8622
& 5 \\ 
\hline

\cellcolor{purple!20}\textbf{BorderlineSMOTE + MICE}     
& \cellcolor{purple!20}0.2656 (0.2034)
& \cellcolor{purple!20}0.1530 (0.0644)
& \cellcolor{purple!20}0.1977 (0.1244)
& \cellcolor{purple!20}0.6163
& \cellcolor{purple!20}1 \\ 
\hline

\textbf{SVMSMOTE + MICE}
& 0.4016 (0.2159)
& 0.1117 (0.1051)
& 0.1695 (0.0457)
& 0.6828
& 3 \\ 
\hline

\textbf{VAE + MICE}
& 0.3009 (0.0401)
& 0.4835 (0.0278)
& 0.5466 (0.0247)
& 1.331
& 7 \\ 
\hline

\textbf{WGAN + MICE}
& 0.4912 (0.0045)
& 0.6214 (0.0025)
& 0.6627 (0.0023)
& 1.7753
& 10 \\ 
\hline

\textbf{CK4Gen + MICE}
& 0.5489 (0.0000)
& 0.1167 (0.0000)
& 0.0241 (0.0000)
& 0.6897
& 4 \\ 
\hline
%###===>>>%###===>>>%###===>>>
%###===>>>%###===>>>%###===>>>
%###===>>>%###===>>>%###===>>>
\multicolumn{6}{c}{}\\
\multicolumn{6}{c}{\footnotesize{Stratification On: Hypertension}}\\
\hline
\textbf{Augmentation Method} & 
\textbf{Duration Target:\newline 25th Percentile} &
\textbf{Duration Target:\newline 50th Percentile} &
\textbf{Duration Target:\newline 75th Percentile} &
\textbf{Sum of\newline Calibration Loss} & 
\textbf{Rank} \\ 
\hline\hline

\cellcolor{green!20}\textbf{No Augmentation}
& \cellcolor{green!20}0.3503
& \cellcolor{green!20}0.4434                
& \cellcolor{green!20}0.4759                      
& \cellcolor{green!20}1.2696              
& \cellcolor{green!20}10 \\ 
\hline\hline

\cellcolor{blue!20}\textbf{MCM + MICE}
& \cellcolor{blue!20}0.1732 (0.0030)
& \cellcolor{blue!20}0.3374 (0.0020)
& \cellcolor{blue!20}0.3958 (0.0016)
& \cellcolor{blue!20}0.9064
& \cellcolor{blue!20}4 \\ 
\hline

\cellcolor{purple!20}\textbf{RandomOverSampler + MICE}   
& \cellcolor{purple!20}0.1405 (0.0373)
& \cellcolor{purple!20}0.3162 (0.0158)
& \cellcolor{purple!20}0.3787 (0.0108)
& \cellcolor{purple!20}0.8354
& \cellcolor{purple!20}1 \\ 
\hline

\textbf{SMOTE + MICE}
& 0.2421 (0.0162)
& 0.3749 (0.0197)
& 0.4219 (0.0208)
& 1.0389
& 5 \\ 
\hline

\textbf{SMOTENC + MICE}             
& 0.1609 (0.0325)
& 0.3230 (0.0187)
& 0.3821 (0.0163)
& 0.866
& 2 \\ 
\hline

\textbf{ADASYN + MICE}
& 0.2429 (0.0134)
& 0.3748 (0.0153)
& 0.4215 (0.0166)
& 1.0392
& 6 \\ 
\hline

\textbf{BorderlineSMOTE + MICE}
& 0.2549 (0.0137)
& 0.3838 (0.0123)
& 0.4292 (0.0129)
& 1.0679
& 8 \\ 
\hline

\textbf{SVMSMOTE + MICE}
& 0.2483 (0.0378)
& 0.3792 (0.0209)
& 0.4251 (0.0164)
& 1.0526
& 7 \\ 
\hline

\textbf{VAE + MICE}
& 0.3480 (0.0094)
& 0.4379 (0.0104)
& 0.4699 (0.0109)
& 1.2558
& 9 \\ 
\hline

\textbf{WGAN + MICE}
& 0.4082 (0.0127)
& 0.4693 (0.0098)
& 0.4899 (0.0088)
& 1.3674
& 11 \\ 
\hline

\textbf{CK4Gen + MICE}
& 0.1695 (0.0000)
& 0.3356 (0.0000)
& 0.3946 (0.0000)
& 0.8997
& 3 \\ 
\hline
\end{tabular}
\end{table}

\newpage
\begin{table}[h]
\tiny
\centering
\caption{Calibration Target: Age}
\begin{tabular}{|p{3.5cm}||p{1.75cm}|p{1.75cm}|p{1.75cm}|p{1.75cm}|p{1cm}|}
\multicolumn{6}{c}{\footnotesize{Stratification On: Younger (Age $<$ 65 years)}}\\
\hline
\textbf{Augmentation Method} & 
\textbf{Duration Target:\newline 25th Percentile} &
\textbf{Duration Target:\newline 50th Percentile} &
\textbf{Duration Target:\newline 75th Percentile} &
\textbf{Sum of\newline Calibration Loss} & 
\textbf{Rank} \\ 
\hline\hline

\cellcolor{green!20}\textbf{Original}
& \cellcolor{green!20}0.0885
& \cellcolor{green!20}0.2405
& \cellcolor{green!20}0.2932
& \cellcolor{green!20}0.6222
& \cellcolor{green!20}10 \\ 
\hline\hline

\cellcolor{blue!20}\textbf{MCM + MICE}
& \cellcolor{blue!20}0.0768 (0.0037)
& \cellcolor{blue!20}0.1041 (0.0030)
& \cellcolor{blue!20}0.1680 (0.0029)
& \cellcolor{blue!20}0.3489
& \cellcolor{blue!20}2 \\ 
\hline

\textbf{RandomOverSampler + MICE}
& 0.2718 (0.0832)
& 0.0534 (0.0316)
& 0.1246 (0.0492)
& 0.4498
& 8 \\ 
\hline

\textbf{SMOTE + MICE}
& 0.1814 (0.0709)
& 0.0606 (0.0454)
& 0.1415 (0.0382)
& 0.3835
& 5 \\ 
\hline

\textbf{SMOTENC + MICE}
& 0.2561 (0.0884)
& 0.0526 (0.0283)
& 0.1180 (0.0472)
& 0.4267
& 7 \\ 
\hline

\textbf{ADASYN + MICE}
& 0.1742 (0.0421)
& 0.0631 (0.0349)
& 0.1423 (0.0331)
& 0.3796
& 4 \\ 
\hline

\textbf{BorderlineSMOTE + MICE}
& 0.2116 (0.0384)
& 0.0407 (0.0351)
& 0.1250 (0.0343)
& 0.3773
& 3 \\ 
\hline

\textbf{SVMSMOTE + MICE}
& 0.1935 (0.0345)
& 0.0536 (0.0294)
& 0.1370 (0.0285)
& 0.3841
& 6 \\ 
\hline

\textbf{VAE + MICE}
& 0.0738 (0.0383)
& 0.2190 (0.0470)
& 0.2699 (0.0502)
& 0.5627
& 9 \\ 
\hline

\textbf{WGAN + MICE}
& 0.1368 (0.0087)
& 0.2621 (0.0072)
& 0.3088 (0.0069)
& 0.7077
& 11 \\ 
\hline

\cellcolor{purple!20}\textbf{CK4Gen + MICE}
& \cellcolor{purple!20}0.0804 (0.0000)
& \cellcolor{purple!20}0.1002 (0.0000)
& \cellcolor{purple!20}0.1640 (0.0000)
& \cellcolor{purple!20}0.3446
& \cellcolor{purple!20}1 \\ 
\hline
%###===>>>%###===>>>%###===>>>
\multicolumn{6}{c}{}\\
\multicolumn{6}{c}{\footnotesize{Stratification On: Older (Age $\geq$ 65 years)}}\\
\hline
\textbf{Augmentation Method} & 
\textbf{Duration Target:\newline 25th Percentile} &
\textbf{Duration Target:\newline 50th Percentile} &
\textbf{Duration Target:\newline 75th Percentile} &
\textbf{Sum of\newline Calibration Loss} & 
\textbf{Rank} \\ 
\hline\hline

\cellcolor{green!20}\textbf{Original}
& \cellcolor{green!20}0.8115
& \cellcolor{green!20}0.8387
& \cellcolor{green!20}0.8483
& \cellcolor{green!20}2.4985
& \cellcolor{green!20}10 \\ 
\hline\hline

\cellcolor{blue!20}\textbf{MCM + MICE}
& \cellcolor{blue!20}0.7271 (0.0006)
& \cellcolor{blue!20}0.7833 (0.0004)
& \cellcolor{blue!20}0.8030 (0.0003)
& \cellcolor{blue!20}2.3134
& \cellcolor{blue!20}2 \\ 
\hline

\textbf{RandomOverSampler + MICE}
& 0.7267 (0.0161)
& 0.7837 (0.0094)
& 0.8038 (0.0070)
& 2.3142
& 3 \\ 
\hline

\textbf{SMOTE + MICE}
& 0.7572 (0.0203)
& 0.8044 (0.0123)
& 0.8208 (0.0095)
& 2.3824
& 7 \\ 
\hline

\textbf{SMOTENC + MICE}
& 0.7383 (0.0226)
& 0.7914 (0.0153)
& 0.8106 (0.0124)
& 2.3403
& 4 \\ 
\hline

\textbf{ADASYN + MICE}
& 0.7571 (0.0180)
& 0.8043 (0.0111)
& 0.8208 (0.0086)
& 2.3822
& 6 \\ 
\hline

\textbf{BorderlineSMOTE + MICE}
& 0.7579 (0.0167)
& 0.8049 (0.0099)
& 0.8212 (0.0074)
& 2.384
& 8 \\ 
\hline

\textbf{SVMSMOTE + MICE}
& 0.7541 (0.0205)
& 0.8020 (0.0123)
& 0.8187 (0.0093)
& 2.3748
& 5 \\ 
\hline

\textbf{VAE + MICE}
& 0.8151 (0.0054)
& 0.8402 (0.0037)
& 0.8491 (0.0030)
& 2.5044
& 11 \\ 
\hline

\textbf{WGAN + MICE}
& 0.8087 (0.0136)
& 0.8377 (0.0061)
& 0.8474 (0.0039)
& 2.4938
& 9 \\ 
\hline

\cellcolor{purple!20}\textbf{CK4Gen + MICE}
& \cellcolor{purple!20}0.7225 (0.0000)
& \cellcolor{purple!20}0.7805 (0.0000)
& \cellcolor{purple!20}0.8007 (0.0000)
& \cellcolor{purple!20}2.3037
& \cellcolor{purple!20}1 \\ 
\hline
\end{tabular}
\end{table}

%###===###%###===###%###===###
%###===###%###===###%###===###
%###===###%###===###%###===###
\newpage
\begin{table}[h]
\tiny
\centering
\caption{Calibration Target: Cardiovascular Disease Status}
\label{tab:calibration_loss_cvd}
\begin{tabular}{|p{3.5cm}||p{1.75cm}|p{1.75cm}|p{1.75cm}|p{1.75cm}|p{1cm}|}

\multicolumn{6}{c}{\footnotesize{Stratification On: No CVD}}\\
\hline
\textbf{Augmentation Method} & 
\textbf{Duration Target:\newline 25th Percentile} &
\textbf{Duration Target:\newline 50th Percentile} &
\textbf{Duration Target:\newline 75th Percentile} &
\textbf{Sum of\newline Calibration Loss} & 
\textbf{Rank} \\ 
\hline\hline

\cellcolor{green!20}\textbf{Original}
& \cellcolor{green!20}0.0966
& \cellcolor{green!20}0.1259
& \cellcolor{green!20}0.2021
& \cellcolor{green!20}0.4246
& \cellcolor{green!20}2 \\ 
\hline\hline

\cellcolor{blue!20}\textbf{MCM + MICE}
& \cellcolor{blue!20}0.6011 (0.0087)
& \cellcolor{blue!20}0.2345 (0.0063)
& \cellcolor{blue!20}0.0925 (0.0053)
& \cellcolor{blue!20}0.9281
& \cellcolor{blue!20}4 \\ 
\hline

\textbf{RandomOverSampler + MICE}
& 1.2009 (0.6740)
& 0.6551 (0.4938)
& 0.4317 (0.4032)
& 2.2877
& 11 \\ 
\hline

\textbf{SMOTE + MICE}
& 0.6962 (0.1699)
& 0.2830 (0.1024)
& 0.1333 (0.0837)
& 1.1125
& 6 \\ 
\hline

\textbf{SMOTENC + MICE}
& 0.7375 (0.1335)
& 0.3090 (0.0907)
& 0.1375 (0.0693)
& 1.184
& 7 \\ 
\hline

\textbf{ADASYN + MICE}
& 0.7613 (0.3089)
& 0.3234 (0.1922)
& 0.1663 (0.1532)
& 1.251
& 8 \\ 
\hline

\textbf{BorderlineSMOTE + MICE}
& 0.7963 (0.2426)
& 0.3321 (0.1609)
& 0.1703 (0.1335)
& 1.2987
& 9 \\ 
\hline

\textbf{SVMSMOTE + MICE}
& 0.7815 (0.1360)
& 0.3574 (0.0913)
& 0.2097 (0.0742)
& 1.3486
& 10 \\ 
\hline

\cellcolor{purple!20}\textbf{VAE + MICE}
& \cellcolor{purple!20}0.1600 (0.0994)
& \cellcolor{purple!20}0.0886 (0.0542)
& \cellcolor{purple!20}0.1473 (0.0706)
& \cellcolor{purple!20}0.3959
& \cellcolor{purple!20}1 \\ 
\hline

\textbf{WGAN + MICE}
& 0.0861 (0.0136)
& 0.1469 (0.0088)
& 0.2279 (0.0073)
& 0.4609
& 3 \\ 
\hline

\textbf{CK4Gen + MICE}
& 0.6544 (0.0000)
& 0.2717 (0.0000)
& 0.1235 (0.0000)
& 1.0496
& 5 \\ 
\hline
%###===>>>%###===>>>%###===>>>
\multicolumn{6}{c}{}\\
\multicolumn{6}{c}{\footnotesize{Stratification On: CVD}}\\
\hline
\textbf{Augmentation Method} & 
\textbf{Duration Target:\newline 25th Percentile} &
\textbf{Duration Target:\newline 50th Percentile} &
\textbf{Duration Target:\newline 75th Percentile} &
\textbf{Sum of\newline Calibration Loss} & 
\textbf{Rank} \\ 
\hline\hline

\cellcolor{green!20}\textbf{Original}
& \cellcolor{green!20}0.8686
& \cellcolor{green!20}0.886
& \cellcolor{green!20}0.8921
& \cellcolor{green!20}2.6467
& \cellcolor{green!20}10 \\ 
\hline\hline

\cellcolor{blue!20}\textbf{MCM + MICE}
& \cellcolor{blue!20}0.8211 (0.0032)
& \cellcolor{blue!20}0.8517 (0.0020)
& \cellcolor{blue!20}0.8629 (0.0016)
& \cellcolor{blue!20}2.5357
& \cellcolor{blue!20}3 \\ 
\hline

\textbf{RandomOverSampler + MICE}
& 0.8195 (0.0324)
& 0.8530 (0.0215)
& 0.8648 (0.0181)
& 2.5373
& 4 \\ 
\hline

\textbf{SMOTE + MICE}
& 0.8273 (0.0183)
& 0.8581 (0.0119)
& 0.8688 (0.0100)
& 2.5542
& 8 \\ 
\hline

\cellcolor{purple!20}\textbf{SMOTENC + MICE}
& \cellcolor{purple!20}0.8151 (0.0296)
& \cellcolor{purple!20}0.8497 (0.0198)
& \cellcolor{purple!20}0.8616 (0.0168)
& \cellcolor{purple!20}2.5264
& \cellcolor{purple!20}1 \\ 
\hline

\textbf{ADASYN + MICE}
& 0.8251 (0.0209)
& 0.8569 (0.0140)
& 0.8679 (0.0119)
& 2.5499
& 6 \\ 
\hline

\textbf{BorderlineSMOTE + MICE}
& 0.8268 (0.0240)
& 0.8577 (0.0164)
& 0.8684 (0.0140)
& 2.5529
& 7 \\ 
\hline

\textbf{SVMSMOTE + MICE}
& 0.8241 (0.0230)
& 0.8560 (0.0156)
& 0.8671 (0.0133)
& 2.5472
& 5 \\ 
\hline

\textbf{VAE + MICE}
& 0.8502 (0.0056)
& 0.8726 (0.0043)
& 0.8807 (0.0038)
& 2.6035
& 9 \\ 
\hline

\textbf{WGAN + MICE}
& 0.8731 (0.0025)
& 0.8888 (0.0020)
& 0.943 (0.0017)
& 2.7049
& 11 \\ 
\hline

\textbf{CK4Gen + MICE}
& 0.8171 (0.0000)
& 0.8492 (0.0000)
& 0.8610 (0.0000)
& 2.5273
& 2 \\ 
\hline
\end{tabular}
\end{table}

\end{document}